\documentclass[letterpaper,10pt,conference]{ieeeconf}

\IEEEoverridecommandlockouts
\usepackage{amsmath}
\usepackage{amssymb}
\usepackage{bbm}
\usepackage{mathtools}
\usepackage{booktabs}
\usepackage{graphicx}
\usepackage{algorithm}
\usepackage{algpseudocode}
\usepackage{array}
\usepackage{tabularx}
\usepackage{multirow}
\usepackage{makecell}
\usepackage{url}
\usepackage{float}
\usepackage{placeins}
\usepackage{balance}
\usepackage{xspace}
\usepackage{xcolor}
\usepackage[table]{xcolor}
\makeatletter\let\NAT@parse\undefined\makeatother
\usepackage[numbers,sort&compress]{natbib}
\usepackage[hidelinks]{hyperref}

\usepackage{xcolor}
\usepackage{pifont}
\usepackage{anyfontsize}

\hypersetup{hypertexnames=false}

\newcommand{\cmark}{$\checkmark$}
\newcommand{\xmark}{$\times$}

\newcolumntype{Y}{>{\centering\arraybackslash}X}

\newcommand{\RALkeywords}[1]{%
  \vspace{0.35em}\noindent\textbf{Index Terms---}#1\par\vspace{0.35em}}
\title{\LARGE \bf
SUN: Agentic Robot Policy Learning with Persistent Task Programs
}

\author{%
  \normalsize
  Weiqi Wang$^{1,*}$, Zhi Li$^{1,*}$, Yudong Lei$^{2}$, David Martinez$^{1}$, Xiaofeng Gao$^{3, \ddagger}$, Yuxin Jiang$^{1}$,\\
  Chenfanfu Jiang$^{1}$, Yingnian Wu$^{1,\dagger}$, Demetri Terzopoulos$^{1,\dagger}$, Ran Gong$^{4,\dagger}$\\[3pt]
  \footnotesize
  $^{1}$University of California, Los Angeles \quad
  $^{2}$Carnegie Mellon University \quad
  $^{3}$Amazon \quad
  $^{4}$Robotics and AI Institute\\[3pt]
  Project website: \href{https://kuafu-robo.github.io/}{\textcolor{blue}{\texttt{https://kuafu-robo.github.io/}}}
  \thanks{$^{*}$Equal contribution.}%
  \thanks{$^{\dagger}$Corresponding author.}%
  \thanks{$^{\ddagger}$This work does not relate to the author's position at Amazon.}%
}

\begin{document}
\bstctlcite{RAL:BSTcontrol}
\maketitle
\thispagestyle{empty}
\pagestyle{empty}
\renewcommand{\baselinestretch}{0.99}
\begin{abstract}
Model-based control can directly execute specified objectives, while learning can amortize such behaviors into reactive policies, making their combination a natural solution to multi-stage manipulation. We introduce Semantically UNified (SUN) Programs, typed executables that compile grounded relations into aligned optimal control objectives, satisfaction predicates, and learning rewards. Our harness, Kuafu, equips a foundation model as a task-level agent to orchestrate scene preparation, verification, residual RL, and data production. The agent uses program feedback to repair candidate programs and training diagnostics to calibrate relative reward weights, retaining accepted task semantics across tool calls. Across nine multi-stage manipulation tasks, Kuafu achieves 82.03\% average success rate, significantly outperforming all learned baselines. Its learned controllers generate demonstrations at 10.57$\times$ the human-teleoperation rate, yielding data that improve visual-policy success by 23.6 percentage points over the strongest baseline. The policies transfer zero-shot to physical Franka and Kinova robots, demonstrating sim-to-real generalization.
\end{abstract}

\RALkeywords{AI-Enabled Robotics, Integrated Planning and Learning, Reinforcement Learning}

\section{Introduction}
\label{sec:introduction}

Automatic multi-stage skill acquisition remains a central challenge in robot manipulation. Recent approaches leverage the code-writing capabilities of large language models to decompose tasks into motion primitives or synthesize reward functions for policy learning~\cite{liang2022codeaspolicies, wang2024robogen,singh2023progprompt}. However, they face distinct limitations: motion-planning compositions often struggle with contact-rich dynamics, while reinforcement learning scales poorly to multi-stage tasks due to sparse supervision and semantic ambiguity.

Recent advances in agentic systems enable multi-step digital workflows, including web navigation~\citep{he2024webvoyager}, research and article writing~\citep{shao2024storm}, and task execution across desktop applications~\citep{zhang2025ufo}. These capabilities motivate a robotics-specific execution and learning harness, where an agent can test physical task specifications and acquire deployable skills for contact-rich, multi-stage manipulation through robotics interfaces. Analogous to how a code interpreter lets agents verify software logic, a robotics harness allows them to test task descriptions under simulated physics, identify failure modes, and derive policies from verified successes. 

\begin{figure}[t]
    \centering
    \includegraphics[width=1.0\columnwidth]{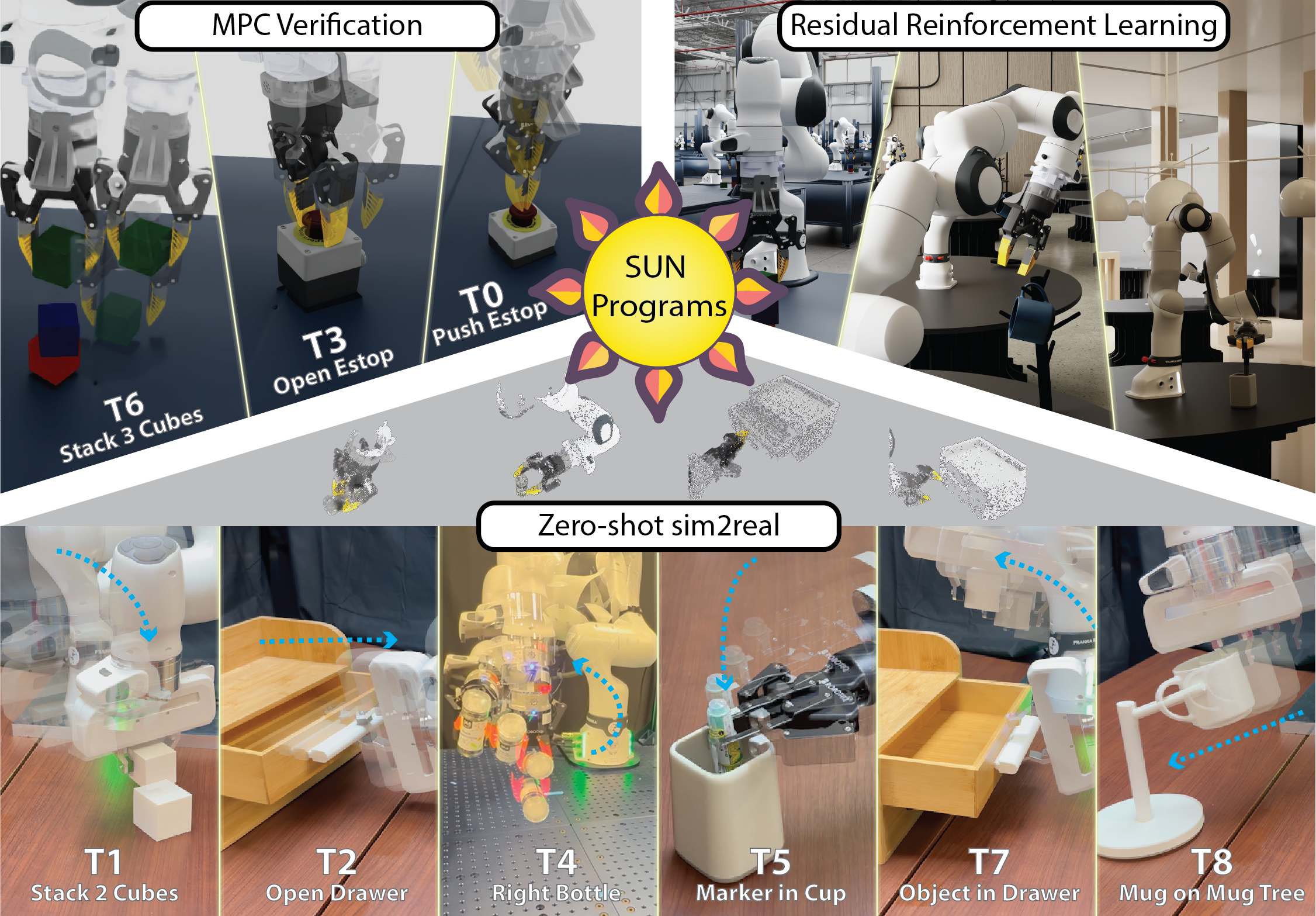}
    \caption{Coordinated by a persistent SUN Program, the Kuafu harness equips the agent with physics-based verification, policy learning, and data-production capabilities. The program supervises the simulation controller and its demonstrations; downstream visual policies learn from recorded observations and actions alone. We evaluate nine tasks and transfer visual policies for six of them to Franka FR3 and Kinova Gen3 robots.}
    \label{fig:kuafu-achievement}
\end{figure}

Realizing such a harness requires mechanisms that support both rapid physics-based verification and the acquisition of deployable behaviors. Control and learning provide complementary mechanisms for these objectives: MPC enables rapid feasibility verification via closed-loop simulation~\citep{toussaint2015logic,mayne2000constrained}, whereas RL produces deployable reactive behaviors. An agent thus benefits from validating semantics through low-cost control verification prior to expensive policy learning. 

\begin{figure*}[t]
    \centering
    \includegraphics[width=0.99\textwidth]{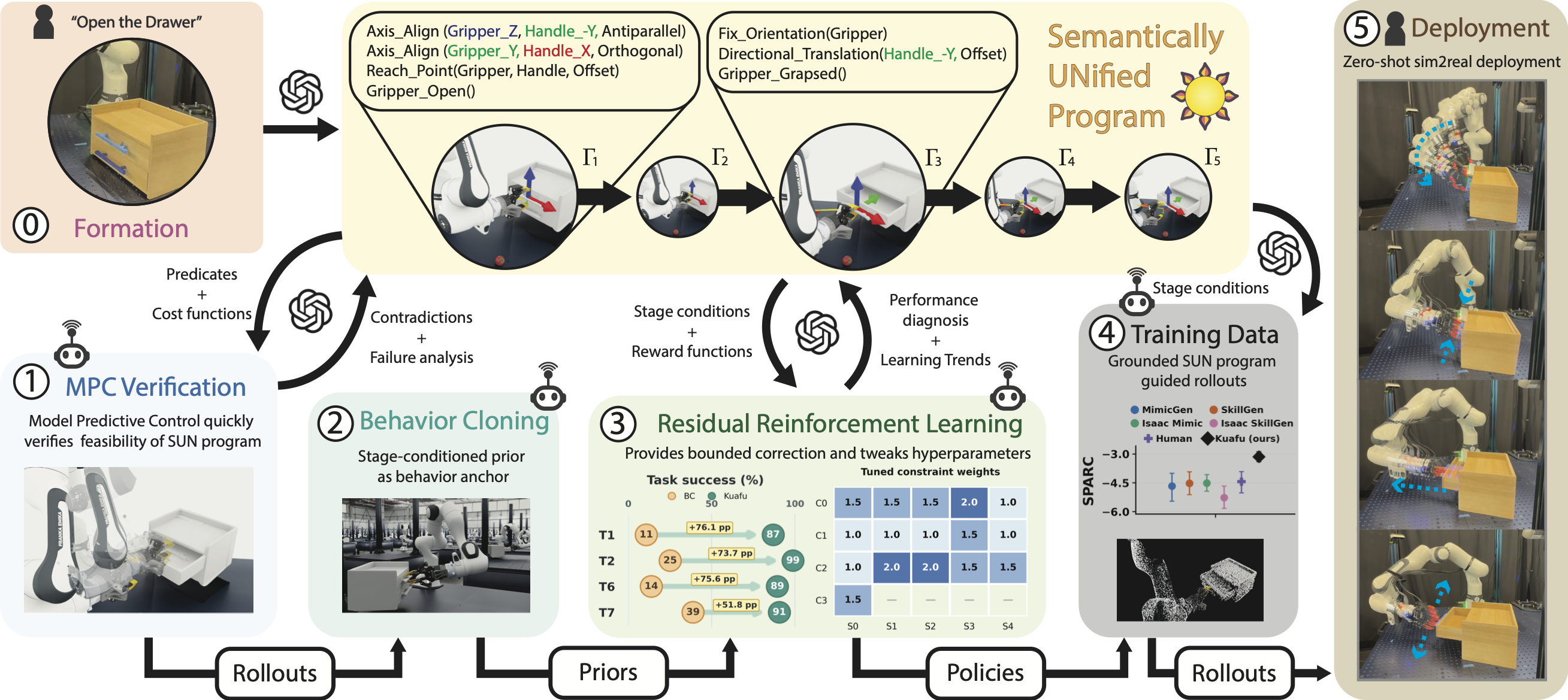}
    \caption{The Kuafu harness for agentic robot policy learning. After video-based scene preparation, the agent forms and repairs a SUN Program (0), invokes MPC verification (1), and uses returned physical diagnostics to revise failed specifications. An accepted program supplies aligned control, monitoring, and learning interfaces for demonstration generation, Stage-BC (2), bounded residual RL (3), and evaluation/data production (4). Training feedback supports scalar calibration while accepted task semantics remain fixed. Forward arrows carry programs, policies, or data; return arrows carry diagnostics to the agent. The final handoff exports recorded observations and actions alone to a visual policy (5). Controllers execute the individual control steps.}
    \label{fig:pipeline}
\end{figure*}

To this end, we introduce the \emph{Semantically UNified (SUN) Program}: a typed, executable task specification that acts as a shared semantic contract between the control, verification, and learning tools (\autoref{fig:kuafu-achievement}). Our harness, \emph{Kuafu}, makes this semantic contract operational through a coordinated set of robot execution and learning capabilities that the agent can invoke (\autoref{fig:pipeline}). These capabilities enable the agent to develop deployable policies from real-world videos and natural-language instructions using simulated physical and training feedback, while the SUN program preserves the verified task definition across tool calls.

Our contributions are thus twofold:
\begin{itemize}
    \item \textbf{SUN Programs: persistent, executable task semantics.} A typed program compiles grounded relations into aligned control, monitoring, and learning interfaces, preserving the task knowledge tested by MPC across the control-to-learning handoff.
    \item \textbf{Kuafu: a harness for robot policy learning.} It equips the agent with scene preparation, physical verification, demonstration generation, Stage-BC, bounded residual RL, simulation evaluation, and downstream data production around that same program, with diagnostic repair and constrained training calibration.
\end{itemize}
Across nine multi-stage tasks, we demonstrate that the harness and its ablations establish the complementary roles of program-derived supervision, the behavioral prior, and bounded correction.
Our generated data support visual policies and vision-language-action (VLA) models, with zero-shot DP3 deployment on physical Franka and Kinova robots.

\section{Related Work}
\label{sec:related}

\noindent \textbf{Executable Task Representations.}
Multi-stage manipulation requires representations that bridge discrete task progression with continuous motion.
Symbolic planning couples task structure to geometric feasibility~\cite{toussaint2015logic}.
Language-grounded approaches express tasks as spatial objectives, including composable 3D value maps, reward parameters evaluated by MPC~\cite{huang2023voxposer,yu2023language}, and grounded keypoint constraints coupled with closed-loop optimization~\cite{huang2025rekep}.
Formal methods encode temporal progress via finite-state rewards or signal-temporal-logic constraints for learning and predictive control~\cite{icarte2018reward,meng2023stlnpc}.
Unlike these works, SUN Programs do not reconstruct task semantics at each tool interface. Instead, they retain a physically tested program across control, monitoring, and learning modules, thereby preserving semantic consistency throughout the pipeline.

\noindent \textbf{Control-to-Policy Transfer.}
Task structure enters policy learning through generated objectives or controller-derived supervision.
Prior work synthesizes executable rewards and domain-randomization configurations directly from task descriptions~\cite{xie2023text2reward,ma2024eureka,ma2024dreureka}.
However, iteratively refined objectives risk overfitting observed behavior or misstating the underlying task, meaning high return need not imply success on the intended physical outcome~\cite{booth2023perils}.
Alternatively, controller-derived supervision facilitates transfer via optimized trajectory targets~\cite{levine2013guided}, MPC teachers adapted to the learner's state distribution~\cite{kahn2016plato}, and Hamiltonian losses encoding control constraints~\cite{carius2020mpcnet}, while
residual learning preserves nominal behavior by acquiring corrective actions~\cite{johannink2019residual}.
Unlike these disjoint approaches, the Kuafu harness integrates these capabilities around a persistent SUN Program. Control verifies the candidate specification, demonstrations initialize the behavioral prior, and residual learning refines execution using the compiled interface---with the agent calibrating permitted scalars while the accepted task semantics remain fixed.

\noindent \textbf{Agentic Robotics.}
LLM and VLM agents coordinate robotic planning, execution, and learning through generated programs and environmental feedback.
Existing approaches ground action programs in robot capabilities and scene objects~\cite{singh2023progprompt}, incorporate environmental feedback into planning and execution~\cite{huang2023inner,pan2025omnomanip}, and leverage generative agents to automate task construction and produce supervision for skill learning~\cite{wang2024robogen}.
More recently, ASPIRE~\cite{lu2026aspire} distills execution-validated program repairs into reusable skills, while ENPIRE~\cite{xiao2026enpire} supports real-world policy improvement through automated reset, verification, and agent-driven experimentation.
Kuafu handles control-to-learning transfer by retaining the accepted SUN Program across all tools. This persistence allows policy learning to inherit physically tested task semantics directly, circumventing the need for the agent to repeatedly reconstruct objectives and completion conditions.

\noindent \textbf{Autonomous Robot Data Generation.}
Autonomous generation integrates multimodal task creation with planning and RL solvers to synthesize demonstrations for downstream learning~\cite{hua2025gensim2}.
Other methods expand existing demonstrations via object-relative segment retargeting or planned connections between reusable skills~\cite{mandlekar2023mimicgen,garrett2025skillmimicgen}.
Unlike these approaches, Kuafu generates high-fidelity data by first verifying task specifications with MPC and then scaling production via a learned controller. The retained SUN monitor supplies stage context, while an independent task-success criterion filters demonstrations, ensuring downstream policies learn only from physically valid trajectories.

\section{Methods}
\label{sec:method}

Kuafu is an agent harness that converts language-described manipulation tasks into learned multi-stage robot policies.
Given a natural-language instruction $\mathcal I$ and videos of the task objects, Kuafu produces a privileged simulation policy for scalable data generation and a visual policy for physical deployment.
As summarized in \autoref{fig:pipeline}, the agent operates at task-level decision points: it forms and repairs SUN Programs using feedback from MPC verification, initiates behavioral-prior learning from successful rollouts, refines residual learning through multimodal reflection, and leverages the accepted program to guide data-generation rollouts.
Crucially, the same accepted SUN Program persists from verification through policy learning and data production, ensuring the task is never reinterpreted at each tool handoff.
\autoref{alg:kuafu} formalizes this feedback-driven control flow, with superscripted step numbers (e.g., ``$^{(2)}$Tool'') corresponding to the workflow stages in \autoref{fig:pipeline}.

\subsection{Scene Preparation and Agent Interface}
\label{sec:agent_interface}

Kuafu provides the agent with a robot model, simulation configuration, and a typed operator library $\mathcal L$.
From the input videos, the agent invokes reconstruction tools to generate simulation-ready object assets, which are imported into the scene and grounded with semantic frames $\mathcal F$.
These frames expose task-relevant object poses, axes, and reference points for binding to operators in $\mathcal L$.
Together, $\mathcal I$, $\mathcal F$, and $\mathcal L$ constitute the language instruction, scene grounding, and executable vocabulary required to construct a SUN Program.
Crucially, Kuafu instantiates an independent rule-based task-success criterion from the instruction prior to SUN Program synthesis; this evaluator verifies outcomes during program validation, policy learning, and data collection, ensuring ground-truth assessment remains decoupled from the generated programs.

We use ``open the drawer'' as a running example throughout this section.
Here, the scene frames $\mathcal F$ identify the drawer and handle, defining their semantic axes (e.g., the handle-bar $x$-axis and drawer $y$-axis, with $+y$ defined as the closing direction).
The operator library $\mathcal L$ provides typed relations and motion primitives such as \texttt{align\_axis}, \texttt{direction\_move}, and \texttt{fix\_orientation}.
The agent binds these operators to the relevant frames in $\mathcal F$ and sequences them into the ordered stages of a SUN Program.

\subsection{SUN Programs as a Shared Task Representation}
\label{sec:problem}

A SUN Program $\Gamma=(\Gamma_1,\ldots,\Gamma_K)$ serves as the executable task representation shared across Kuafu's control and learning tools.
The agent constructs $\Gamma$ by grounding operators from $\mathcal L$ in the semantic frames $\mathcal F$ and organizing the resulting relations into $K$ ordered stages. For the ``open the drawer'' example, these relations specify conditions such as aligning the gripper with the handle, reaching the target, maintaining orientation, and translating the drawer along its opening axis. Each relation retains its grounded arguments, satisfaction boundary, and stage membership throughout the entire workflow. For every relation $j$ in stage $k \in \{1,\ldots,K\}$, the SUN compiler exposes three executable views:
\begin{equation}
\operatorname{Compile}(\Gamma_k)
=
\left\{
\left(
\phi^{\mathrm{mpc}}_{k,j},
P_{k,j},
c^{\mathrm{rl}}_{k,j}
\right)
\right\}_j ,
\label{eq:sun_compile}
\end{equation}
where $\phi^{\mathrm{mpc}}_{k,j}$ is an MPC objective term, $P_{k,j}$ is a satisfaction predicate, and $c^{\mathrm{rl}}_{k,j}$ is an RL learning cost.
These views are operator-specific and need not share the same numerical form;
for instance, a point-reaching relation may employ a quadratic distance penalty for MPC, a tolerance test for satisfaction, and a bounded distance cost for learning.
What remains invariant across these views is the relation itself: its grounding, satisfaction boundary, and position in the task sequence.
We refer to this correspondence across control, monitoring, and learning as \emph{Semantic Unification}.

The SUN Program also governs task progression.
A runtime monitor maintains the active stage $z_t$ at time $t$ and stores an entry-state snapshot $\xi_{z_t}$ containing the reference poses captured at the start of the current stage.
After each action, it evaluates the stage's satisfaction predicates on the updated state, advancing only when all conditions are met.
This identical stage logic is applied uniformly across MPC verification, Stage-BC, residual learning, and data production.
Thus, downstream tools do not reconstruct task progression from raw trajectories or independently redefine completion conditions; they operate on the same verified task semantics established at the outset.

\begin{algorithm}[t] \small
\caption{Kuafu agent harness with the SUN program.}
\label{alg:kuafu}
\begin{algorithmic}[1]
\Statex \textbf{Input:} Instruction $\mathcal I$, object videos, operator library $\mathcal L$
\State \textbf{$^{(0)}$Tool:} reconstruct scene and expose frames $\mathcal F$
\State \textbf{$^{(0)}$Agent:} form SUN Program $\Gamma$ from $(\mathcal I,\mathcal F,\mathcal L)$
\While{verification budget remains}
    \State \textbf{$^{(1)}$Tool:} compile $\Gamma$; MPC verify on initial states
  \State \textbf{if accepted then} $\Gamma^\star\leftarrow\Gamma$; \textbf{break}
    \State \textbf{$^{(1)}$Agent:} repair $\Gamma$ from feedback
\EndWhile
\State \textbf{if no $\Gamma^\star$ then return} formation not accepted
\State \textbf{$^{(2)}$Tool:} MPC rollouts $\rightarrow$ frozen Stage-BC $\pi_\theta$
\State \textbf{$^{(3)}$Tool:} initialize residual learning using $\pi_\theta$ and $\Gamma^\star$
\Repeat\label{alg:calibration_start}
    \State \textbf{$^{(3)}$Agent:} propose weights $w$, scales $\alpha$
    \State \textbf{$^{(3)}$Tool:} train/evaluate; \textbf{stop early if} $\mathrm{SR}_t < \beta\, \mathrm{SR}_{\max}$
\Until{training completes without early stopping}\label{alg:calibration_end}
\State \textbf{$^{(4)}$Tool:} roll out privileged policy $\pi_{\mathrm{pri}}$: $\pi_{\mathrm{pri}}$ outputs actions; $\Gamma^\star$ advances and labels stages; retain successful trajectories
\State \textbf{$^{(5)}$Tool:} train and evaluate visual policy $\pi_{\mathrm{vis}}$ from the retained sensor--action trajectories
\State \Return Privileged policy $\pi_{\mathrm{pri}}$, visual policy $\pi_{\mathrm{vis}}$
\end{algorithmic}
\end{algorithm}

\subsection{Verification and Program Repair}
\label{sec:accept_amortize}

Kuafu verifies a candidate SUN Program via closed-loop MPC prior to policy learning (Step~1 in \autoref{fig:pipeline}).
For each active stage, the verification tool optimizes the compiled MPC objectives, executes the resulting action, observes the new state, and re-evaluates the SUN satisfaction predicates before re-planning.
In parallel, an independent rule-based task-success criterion evaluates whether the intended physical outcome has been achieved.
Crucially, this check is generated independently of the candidate SUN Program and frozen before synthesis begins.
For the ``open the drawer'' example, it declares success only when the drawer reaches the required opening displacement.

The resulting rollouts test two critical aspects: whether the grounded objectives are physically executable and whether the program's completion logic aligns with the intended task outcome.
The tool returns detailed diagnostics identifying unsatisfied relations, measured constraint violations, stage failures, and \emph{false completions}---cases where the SUN Program declares success, but the independent task-success check rejects the outcome.
The agent uses this feedback to repair the earliest diagnosed failure while preserving all previously validated requirements, then resubmits the revised program for verification.

A candidate program is accepted if at least one rollout achieves the independently verified task outcome and no trajectory in the verification batch produces a \emph{false completion}.
Upon acceptance, the grounded relations, satisfaction boundaries, predicates, operator definitions, and stage order are frozen in $\Gamma^\star$.
The agent ceases to revise the task semantics; subsequent interactions focus exclusively on learning how to execute the accepted program more robustly.

\subsection{Policy Learning and Agent Calibration}
\label{sec:bc}
\label{sec:residual_rl}

Once a SUN Program $\Gamma^\star$ is accepted, Kuafu leverages its successful MPC rollouts to learn a stage-conditioned behavioral prior $\pi_\theta(o_t,z_t)$ (Step~2 in \autoref{fig:pipeline}).
Here the observation $o_t$ comprises privileged robot and object observations, while the active stage $z_t$ is provided by the verified SUN monitor. The action $u_t$ corresponds to a task-space end-effector target paired with a binary gripper command.
Crucially, this prior learns how to execute the demonstrated behavior without needing to reconstruct task progression from raw trajectories.
At execution time, Stage-BC replaces the computationally expensive online MPC optimization with fast, learned action prediction, while $\Gamma^\star$ retains sole authority over determining when to advance stages~\cite{byravan2022evaluating}.
For the ``open the drawer'' example, $o_t$ includes the end-effector, drawer, and handle poses, along with the gripper state.

Kuafu then freezes $\pi_\theta$ and learns a bounded residual policy $\pi_\psi$ to refine execution (Step~3 in \autoref{fig:pipeline}).
Following residual RL~\citep{johannink2019residual}, the privileged whole-task policy is
\begin{equation}
\begin{aligned}
\pi_{\mathrm{pri}}(o_t,z_t)
&=
\operatorname{clip}\bigl(
\pi_\theta(o_t,z_t)
+
\alpha\pi_\psi(o_t,z_t),
u_{\min},u_{\max}
\bigr),
\\
u_t&=\pi_{\mathrm{pri}}(o_t,z_t),
\qquad
\left\|\pi_\psi(o_t,z_t)\right\|_\infty
\leq \rho_{\max}.
\end{aligned}
\label{eq:residual_policy}
\end{equation}
Here, $\alpha$ scales the correction magnitude, $\rho_{\max}$ bounds each residual component, and $u_{\min}$, $u_{\max}$ denote environment action limits.
Note that this learned action residual modifies control commands; it is distinct from the geometric residuals within SUN operators, which measure task satisfaction rather than modifying actions.

Residual learning inherits the learning costs compiled from $\Gamma^\star$, \textbf{avoiding the need to synthesize new task rewards}, ensuring stable behavior.
For relation $j$ in stage $k$, these costs $c^{\mathrm{rl}}_{k,j}(t)$ and the relative weights $w_{k,j}$ define the normalized progress score
\begin{equation}
S_k(t)
=
1-
\frac{
\sum_j w_{k,j}c^{\mathrm{rl}}_{k,j}(t)
}{
\sum_j w_{k,j}
},
\label{eq:residual_state_score}
\end{equation}
where costs are evaluated against the current state and the stage-specific reference information stored by the SUN monitor.
Higher values of $S_k(t)$ indicate greater progress toward the accepted stage requirements.
The reward function then structures learning around incremental improvement beyond the best score previously attained in the active stage:
\begin{equation}
\begin{aligned}
r_t={}&
\lambda_{\mathrm{prog}}
\left[
S_{z_t}(t+1)
-
S^{\mathrm{best}}_{z_t}(t)
\right]_+
+
\lambda_{\mathrm{stage}}
\mathbf{1}[z_{t+1}>z_t]
\\
&+
\lambda_{\mathrm{true}}
\mathbf{1}[\mathrm{TS}_t]
-
\lambda_{\mathrm{false}}
\mathbf{1}[\mathrm{FC}_t]
-
\lambda_{\mathrm{act}}
\|u_t\|_2^2 .
\end{aligned}
\label{eq:residual_reward}
\end{equation}
Here, $\mathrm{TS}_t$ and $\mathrm{FC}_t$ denote task-success and false-completion events, respectively.
The term $S^{\mathrm{best}}$ initializes to zero upon stage entry, tracks the largest score achieved within that stage, and resets when the SUN monitor advances.
Crucially, the progress term is evaluated prior to the stage transition, ensuring that both dense progress signals and stage advancement are measured against the same verified stage definition.
Thus, the SUN Program not only dictates task progression but also defines the metric by which residual learning measures progress within each stage.

During the learning phase (lines~\ref{alg:calibration_start}--\ref{alg:calibration_end} in \autoref{alg:kuafu}), the agent uses learning trends and stage-wise constraint satisfaction to calibrate the relative term weights $w_{k,j}$, residual strength $\alpha$, and bound $\rho_{\max}$ (Step~3 in \autoref{fig:pipeline}).
Each residual-RL iteration yields a training success rate $\mathrm{SR}_t$.
The process terminates early if $\mathrm{SR}_t$ drops below a fraction $\beta$ of the peak success rate achieved so far; in this case, the agent updates the hyperparameters and re-trains.
These updates are strictly limited to adjusting reward emphasis and the permitted deviation from the frozen Stage-BC prior. The agent is prohibited from altering the task semantics verified by the MPC mentor.

We define the sequence from an accepted SUN Program through its MPC mentor rollouts, Stage-BC prior, residual refinement, and final privileged policy as a \emph{lineage}.
Consequently, independent lineages jointly vary the program formation process, mentor trajectory data, and subsequent policy learning parameters, ensuring that each lineage represents a distinct, end-to-end execution of the full pipeline.

\subsection{SUN-Guided Data Production and Deployment}
\label{sec:sim2real}

After learning, Kuafu freezes the final privileged policy $\pi_{\mathrm{pri}}$ and validates it in simulation.
The frozen controller then drives scalable data production (Step~4 in \autoref{fig:pipeline}).
During each rollout, $\pi_{\mathrm{pri}}$ generates robot actions while the accepted SUN Program $\Gamma^\star$ remains active as the task monitor.
It supplies the current stage $z_t$ to the Stage-BC and residual policies, advances the task using the predicates verified by MPC, and records corresponding stage labels.
Only trajectories satisfying the independent task-success check are retained.
Thus, the learned policy executes the motion, while SUN organizes the task logic and annotates the resulting demonstrations.
The retained rollouts provide the observation-action pairs used to train the downstream visual policy $\pi_{\mathrm{vis}}$ (Step~5 in \autoref{fig:pipeline}), which is subsequently deployed on the physical robot.

\section{Experiments}
\label{sec:experiments}

\begin{table}[!tb]
\vspace{1mm}
\centering
\vspace{5pt}
\caption{Formation, verification, and agent interaction cost.}
\label{tab:formation_reliability}
\begingroup
\fontsize{6.5}{7.2}\selectfont
\setlength{\tabcolsep}{0.7pt}
\renewcommand{\arraystretch}{0.92}
\setlength{\aboverulesep}{0.25ex}
\setlength{\belowrulesep}{0.25ex}

\begin{tabular*}{\columnwidth}{
@{\extracolsep{\fill}}l*{9}{r}@{\hspace{4.5pt}}r@{}
}
\toprule
Metric / method
& T0 & T1 & T2 & T3 & T4 & T5 & T6 & T7 & T8 & Mean\\
\midrule

\rowcolor{black!7}
\multicolumn{11}{@{}l}{
\textit{A. SUN Program formation and refinement: accepted programs and formation cost}}\\
\midrule
Stages $K$
& 3 & 8 & 6 & 6 & 7 & 9 & 15 & 14 & 10 & 8.67\\
Constraints
& 12 & 25 & 19 & 19 & 27 & 31 & 53 & 44 & 31 & 29.00\\
Formation iter.
& 1 & 1 & 1 & 1 & 2 & 2 & 2 & 1 & 1 & 1.33\\
Refinement iter.
& 0 & 0 & 0 & 0 & 1 & 1 & 0 & 0 & 2 & 0.44\\
Agent calls/task
& 11.2 & 19.8 & 15.2 & 16.2 & 24.6
& 29.2 & 37.0 & 32.8 & 28.8 & 23.87\\

\midrule
\rowcolor{black!7}
\multicolumn{11}{@{}l}{
\textit{B. Five repeated formation-and-MPC-verification runs per task}}\\
\midrule
Success w/o rec.
& 5/5 & 5/5 & 5/5 & 4/5 & 0/5
& 1/5 & 5/5 & 2/5 & 2/5 & 29/45\\
Recovery entry
& 0/5 & 0/5 & 0/5 & 1/5 & 5/5
& 4/5 & 0/5 & 3/5 & 3/5 & 16/45\\
\textbf{Final success}
& \textbf{5/5} & \textbf{5/5} & \textbf{5/5}
& \textbf{5/5} & \textbf{5/5} & \textbf{5/5}
& \textbf{5/5} & \textbf{3/5} & \textbf{5/5}
& \textbf{43/45}\\
Recovery cycles
& 0 & 0 & 0 & 2 & 5 & 9 & 0 & 11 & 3 & 30\\

\midrule
\rowcolor{black!7}
\multicolumn{11}{@{}l}{
\textit{C. Comparator agent interaction cost (tokens in thousands)}}\\
\midrule
\textbf{Kuafu} tokens/task
& 37.6 & 84.0 & 57.1 & 62.6 & 110.1
& 144.1 & 183.7 & 152.5 & 163.1 & 110.50\\
Eureka tokens/task
& 224.3 & 246.2 & 241.1 & 227.2 & 208.7
& 243.0 & 208.9 & 248.8 & 261.0 & 234.40\\
VoxPoser calls/run
& 12.9 & 13.9 & 13.0 & 13.0 & 16.1
& 21.8 & 18.8 & 19.9 & 23.5 & 16.99\\
VoxPoser tokens/run
& 25.1 & 24.6 & 24.9 & 23.6 & 27.2
& 38.0 & 31.4 & 39.7 & 49.8 & 31.60\\
\bottomrule
\end{tabular*}
\endgroup
\end{table}

\begin{table*}[!t]
\centering
\vspace{-5pt}
\caption{Control-to-learning handoff and baselines
}
\label{tab:control_learning_ablation}
\begingroup
\scriptsize
\setlength{\tabcolsep}{2.2pt}
\renewcommand{\arraystretch}{1.0}
\setlength{\aboverulesep}{0.4ex}
\setlength{\belowrulesep}{0.4ex}

\begin{tabular*}{\textwidth}{
@{\extracolsep{\fill}}l*{10}{r}@{}
}
\toprule
Variant / method
& T0 & T1 & T2 & T3 & T4 & T5 & T6 & T7 & T8
& Mean \\
\midrule

\rowcolor{black!7}
\multicolumn{11}{@{}l}{
\textit{A. \textbf{Kuafu} component-ablation success (\%): partial handoffs and alternative state-policy learners}
}\\
\midrule
MPC&100.00&96.15&100.00&89.29&62.50&58.14&94.34&70.92&82.50&83.76\\ 
\midrule
Pure RL, naive reward&0.04&0.00&0.13&0.00&0.43&0.00&0.00&0.00&0.00&0.07\\ 
Pure RL, SUN reward &0.00&0.00&0.00&0.00&0.00&0.00&0.00&0.00&0.00&0.00\\
Stage-free-BC &52.20&12.16&3.38&11.17&5.29&0.00&0.00&0.00&0.77&9.44\\ 
Stage-BC&36.01&10.79&25.11&29.80&22.47&31.82&13.81&38.80&14.16&24.75\\ 
Stage-free-BC + RRL&1.20&67.30&0.03&0.00&\textbf{60.31}&17.49&3.83&0.16&16.69&18.56\\ 
Bounded RRL + Sparse Rewards &58.67&87.20&49.56&58.67&58.96&0.00&0.00&7.95&0.00&35.67\\ 
TD-MPC2~\citep{hansen2024tdmpc2}, naive reward&79.64&27.00&0.00&5.99&1.07&0.00&0.00&0.00&0.00&12.63\\ 
TD-MPC2 + prior, naive reward &77.58&13.00&97.47&\textbf{99.27}&0.20&0.00&0.00&0.00&0.00&31.95\\
\textbf{Kuafu (Primary)}
&\textbf{98.48}&\textbf{95.78}&\textbf{98.01}&84.62&58.80&\textbf{69.70}&\textbf{87.25}&\textbf{95.21}&\textbf{50.43}&\textbf{82.03}\\
\textbf{Kuafu} (3 Runs)  &98.37$\pm$0.81 & 94.75$\pm$0.62 & 97.58$\pm$1.04 & 82.09$\pm$0.65 & 56.24$\pm$1.81 & 62.04$\pm$5.34 & 84.72$\pm$1.32 & 94.06$\pm$1.06 & 44.99$\pm$1.84& 79.43\\
\midrule
\rowcolor{black!7}
\multicolumn{11}{@{}l}{
\textit{B. Baseline performance success (\%): iterative reward search (Eureka) and online planning (VoxPoser)}
}\\
\midrule 
Eureka: Best Train $\leq5$&60.53&0.02&78.96&18.17&6.47&0.01&0.00&0.01&0.01&18.24\\ 
Eureka: Final Eval.&25.85&0.01&71.75&15.16&2.83&0.01&0.00&0.00&0.01&12.85\\ 
VoxPoser: Direct&50.00&10.00&0.00&20.00&0.00&0.00&10.00&0.00&0.00&10.00\\
\bottomrule
\end{tabular*}
\endgroup
\end{table*}

We address four central questions in our experiments: (1) \emph{Formation Reliability:} How reliably can SUN Programs be synthesized from language and scene semantics? (2) \emph{Performance Elevation:} Does the persistent support of SUN Programs significantly elevate task success compared to non-persistent baselines? (3) \emph{Data Efficiency:} Can Kuafu outperform existing data-generation methods in terms of trajectory quality and sample efficiency? (4) \emph{Visual Policy:} Does the dataset produced by Kuafu enable visual policies and robust zero-shot transfer to physical robots?
As shown in \autoref{fig:kuafu-achievement}, we designed nine multi-stage tasks around constrained motion, precise placements, orientation-sensitive grasping, and high-precision manipulation to evaluate physical capabilities that recur throughout everyday manipulation.

\emph{E-stop interactions} (T0: push, T3: twist), \emph{stacking} (T1: two cubes, T6: three cubes), \emph{drawer manipulation} (T2: pull open, T7: open and place object), \emph{object reorientation} (T4: right a bottle), \emph{insertion} (T5: marker into cup), and \emph{dexterous hanging} (T8: hang cup by handle). 
The tasks isolate different forms of compositional scalability for objects and interactions.
T0/T3 vary the complexity of contact-constrained motion on the same mechanism. T6 extends T1 by repeating its pick-and-place sequence. T7 combines T2’s drawer motion with T1's pick-and-place skill. T4/T5 both require object reorientation, but T5 requires an additional inter-object geometric constraint. Lastly, T5 and T8 compare high-precision manipulation across distinct object morphologies and contact geometries.

We report overall performance as task-macro success: the unweighted mean of the success rate across all tasks.
The agents in the primary Kuafu, Eureka, and VoxPoser experiments use the same underlying model, GPT-5.5, for all task-level reasoning calls. All methods share the same task-specific initial-condition sampler.

\subsection{Formation Reliability}
\label{sec:exp_program_formation}

Panels~A and B of \autoref{tab:formation_reliability} demonstrate results for program formation and refinement.
Of 45 independent runs, 29 passed initially; diagnostic repair recovered 14 of the remaining 16, raising the yield from 64.4\% to 95.6\%.
Recovery was most critical for T4 and T5, improving acceptance from 10\% to 100\%.
T7 was the only task with 2 unrecovered runs, and it accounted for 11 of the 30 total recovery cycles.
Since the larger 15-stage T6 program succeeded in all five runs without recovery, raw program size alone does not explain formation difficulty. 
The most common errors in parsed programs involve axis-reference signs (i.e., mistaking the direction of operation) and axis relations (i.e., confusing orthogonal and parallel relations), suggesting that the agent struggles with spatial reasoning, especially for orientation-heavy tasks. Since T7 combines constrained movement with a tight orientation requirement and a separate pick-and-place action, the unrecovered runs suggest that combining heterogeneous manipulation phases poses a greater challenge.
These results establish diagnostic-guided repair as an effective mechanism for recovering initially invalid programs.

Panel~C reports costs of the agentic harness.
Across the 9 tasks, Kuafu consistently used fewer tokens than Eureka, which rewrites the reward functions in each iteration.
At the reported mean rates, VoxPoser's cumulative costs will exceed Kuafu's after just two executions (calls) and four rollouts (tokens). 
These results demonstrate the benefits of forming a persistent program, which converts a recurring agent interaction cost into a task-level cost.

\subsection{Performance and Ablation Analysis}
\label{sec:exp_control_learning}
Panel~A of \autoref{tab:control_learning_ablation} examines the control-to-learning transfer through component ablations and alternative learners, using MPC's 83.76\% task-macro success as the reference.
Pure RL is effectively unsuccessful with either the naive or SUN reward.
With the SUN reward, removing the accepted Stage-BC prior leaves learning trapped in unsuccessful early-stage behaviors.
When constraints are active, the policy struggles to balance geometric terms and often prioritizes their orientation constraints while moving away from positional targets, eventually reaching an equilibrium in which exploration stalls.
Although each reward term is numerically bounded and verified, individual terms—or groups of simultaneously active terms—can still dominate the learned behavior. 
Thus, program-derived rewards alone do not enable the learner to discover useful behavior.

Demonstration supervision provides a starting point, but its effectiveness depends on how task progress is represented.
Stage-free BC leaves several tasks unsolved, whereas Stage-BC achieves nonzero success throughout the suite.
This suggests that explicit progress context helps distinguish the control decisions at different stages.
Nevertheless, the remaining gap from MPC shows that imitation alone does not recover the mentor's performance.
Residual learning does not consistently repair an incomplete handoff.
Starting from stage-free BC, residual learning improves selected tasks but severely degrades T0, T2, and T3. 
Retaining the stage-conditioned prior improves aggregate performance further under sparse rewards, yet T5, T6, and T8 become entirely unsuccessful.
These regressions suggest that sparse feedback provides insufficient guidance to preserve and refine intermediate behavior.
A useful initialization and action corrections therefore do not, by themselves, ensure successful refinement, especially for longer tasks. The alternative learner exhibits a similarly selective benefit from demonstrations. TD-MPC2 with the naive reward performs best on T0; adding a behavioral prior produces strong results on T2 and T3 but leaves T5--T8 unsolved.
Its strongest performance remains concentrated on the shorter tasks rather than extending throughout the suite.
Under this reward, a behavioral prior helps the learner acquire selected tasks but is insufficient to reproduce the coverage of the complete handoff.

Combining the stage-conditioned prior, program-derived supervision, and bounded correction, Kuafu reaches 82.03\% in the primary experiment, the highest task-macro success among the evaluated learned variants in Panel~A.
Unlike the partial handoffs, it maintains success throughout the task suite.
The comparison with sparse-reward refinement is particularly informative: with the Stage-BC prior and residual bound unchanged, SUN supervision recovers successful completion on every task left unsolved by that variant.
Together with the pure-RL failures, this supports complementary roles for the prior and program-derived supervision: the prior supplies useful behavior that reward-driven exploration fails to discover, while the program guides its refinement.
The three-run results for Kuafu retain broad task coverage when the accepted program, MPC mentor set, and residual training vary jointly.
This supports the repeatability of the complete pipeline beyond a single lineage. The paired tasks show Kuafu's robustness to longer stage sequences. Across the stacking and drawer pairs (T1/T6, T2/T7), Kuafu has much smaller success-rate gaps between the shorter and longer variants than the sparse-reward RRL ablation. More broadly, no non-Kuafu learned handoff or baseline reaches 50\% on either T6 or T7, with taskwise maxima of only 13.81\% and 38.80\%, whereas Kuafu reaches 87.25\% and 95.21\%. This shows that the retained SUN Program preserves competence as stage sequences lengthen rather than merely improving isolated tasks.

\noindent \textbf{Baseline Analysis.}
Conversely, baselines that rely on iterative reward search or online planning without persistent task semantics struggle to achieve robust, multi-stage success (Panel~B). Eureka exhibits larger gaps between training and evaluation,
indicating reward overfitting. 
Its failed controllers exhibit the same early fixation on selected geometric features observed in the SUN-reward runs without a prior.
This shared failure suggests that revising rewards alone cannot navigate through multiple complex objectives.
VoxPoser achieves only 10.00\% success, completely failing on five tasks. 
Kuafu's gains stem from integration of a verified prior, program-derived supervision, and bounded correction, which keep the task specification aligned across control and learning. 

\autoref{tab:end_to_end_compute} reports controller-development compute on the prepared simulation scenes, covering program formation, MPC verification, mentor trajectory collection, 
Stage-BC, and bounded residual learning. Kuafu's entries include unsuccessful loops during formation and calibrations. Summation is across all 9 tasks, under single-GPU worker-hour accounting.
Eureka consumes 3.91 times Kuafu's compute. 
Kuafu's cost is dominated by MPC mentor-trajectory collection and residual RL, whereas SUN Program formation and validation require only 0.172 hours per task. 
This small overhead demonstrates the computational benefit of an executable task representation: candidate SUN programs can be evaluated directly through control without training a policy for every candidate.
Thus, Kuafu's superior performance is achieved at a fraction of the computational cost, ruling out compute budget as a confounding factor.

\newcommand{\sumrange}[2]{%
  \shortstack{#1\\[-0.6ex]{\tiny #2}}%
}

\begin{table}[!tb]
\centering
\caption{Controller-development compute (GPU-hrs)}
\label{tab:end_to_end_compute}
\scriptsize
\setlength{\tabcolsep}{1.0pt}
\renewcommand{\arraystretch}{0.98}
\begin{tabular*}{\columnwidth}{
@{\extracolsep{\fill}}lrrrrrrrr@{}
}
\toprule
& \multicolumn{5}{c}{\textbf{Kuafu}}
& \multicolumn{3}{c}{Eureka}\\
\cmidrule(lr){2-6}\cmidrule(l){7-9}
& Form. & MPC & BC & RRL & Total
& Search & Final & Total\\
\midrule
Sum
& 1.55
& 69.03
& 3.21
& 48.86
& \textbf{122.67}
& 412.24
& 67.36
& \textbf{479.60}\\
\bottomrule
\end{tabular*}
\par\raggedright
\end{table}

\vspace{-3pt}
\subsection{Data Generation Efficiency and Data Quality}
\label{sec:exp_data_generation}

During data collection, the learned controller replaces online MPC, while the retained SUN Program continues to supply stage annotations.
Although the MPC mentor leads in overall success rates (as shown in Panel~A of \autoref{tab:control_learning_ablation}), its online optimization loop and single-environment execution limit demonstration throughput severely.
We therefore exclude MPC from the data-collection throughput comparison. 
\autoref{tab:data_engine_main} shows the evaluation results of the data engine through production throughput and controller amortization, MPC-relative trajectory diagnostics, and downstream visual-policy learning under a matched successful-data budget.

We first assess production efficiency, accounting for both collection throughput and the initial time and cost needed to prepare the controller (Panels~A and B).
Human baselines were collected via simulation teleoperation using an AgileX PIKA gripper, with measured poses defining the simulated end-effector targets. 
Once the final state controller is ready, Kuafu produces 10.57$\times$ as much successful trajectory time per active production hour as human teleoperation.
To contextualize, MimicGen's data-generation experiments use 1,000 successful demonstrations per task variant~\citep{mandlekar2023mimicgen}.
Including controller preparation, Kuafu catches up with human teleoperation in cumulative successful demonstration count before reaching 1000 demonstrations on every task.
Pricing human teleoperation at the U.S. federal minimum wage of \$7.25/h and A6000 rental for both controller preparation and data production at \$0.51/GPU h, Kuafu reaches cost parity after 35 successful demonstrations per task on average. 

\begin{table}[!tb]

\centering
\caption{Kuafu data generation
}
\label{tab:data_engine_main}
\renewcommand{\arraystretch}{0.94}
\par\vspace{2pt}
\setlength{\tabcolsep}{0.8pt}
\begin{tabular*}{\linewidth}{@{\extracolsep{\fill}}lrrrrrr@{}}
\toprule
\rowcolor{black!7}
\multicolumn{7}{@{}l}{
\textit{A. Successful trajectory-time throughput (min/active production hr, \(\uparrow\))}
}
\\
\midrule
Task
& Human
& MimicGen
& SkillGen
& \makecell{Isaac Lab\\Mimic}
& \makecell{Isaac Skill\\Mimic}
& \textbf{Kuafu}
\\
\midrule
\textbf{Task macro}
& \textbf{23.34}
& \textbf{29.36}
& \textbf{26.48}
& \textbf{32.19}
& \textbf{24.79}
& \textbf{246.79}
\\
\makecell[l]{vs.\ Human}
& 1.00
& 1.26
& 1.13
& 1.38
& 1.06
& \textbf{10.57}
\\
\end{tabular*}

\par\vspace{2pt}
\setlength{\tabcolsep}{2.15pt}
\begin{tabular*}{\linewidth}{@{\extracolsep{\fill}}lrrr@{}}
\toprule
\rowcolor{black!7}
\multicolumn{4}{@{}l}{
\textit{B. Controller readiness and break-even}}
\\
\midrule
& \makecell{Controller ready}
& \makecell{Time break-even}
& \makecell{Cost break-even}
\\
\midrule
\textbf{Task macro}
& \makecell{\textbf{13.63 GPU h}\\\textbf{\$6.95}}
& \makecell{\textbf{513 demos}\\\textbf{29.10 min}}
& \makecell{\textbf{35.22 demos}\\\textbf{1.99 min}}
\\
\makecell[l]{Across-task\\range}
& \makecell{4.84--25.52 GPU h\\\$2.47--\$13.02}
& \makecell{225--854 demos\\4.35--76.19 min}
& \makecell{16--57 demos\\0.30--5.09 min}
\\
\end{tabular*}

\par\vspace{2pt}
\setlength{\tabcolsep}{2.15pt}
\begin{tabular*}{\linewidth}{@{\extracolsep{\fill}}lrrrrr@{}}
\toprule
\rowcolor{black!7}
\multicolumn{6}{@{}l}{
\textit{C. MPC-relative Q95 yield and successful-trajectory diagnostics}
}
\\
\midrule
Method
& \makecell{Q95\\(\%$\uparrow$)}
& \makecell{Time/\\$\text{MPC}_{50}$ $\downarrow$}
& \makecell{EEF path/\\$\text{MPC}_{50}$ $\downarrow$}
& \makecell{Action path/\\$\text{MPC}_{50}$ $\downarrow$}
& \makecell{Action $\Delta^2$/\\$\text{MPC}_{50}$ $\downarrow$}
\\
\midrule
MimicGen
& 20.4
& 1.11
& 1.14
& 1.94
& 1.27
\\
SkillGen
& 6.9
& 1.66
& 1.15
& 3.01
& 1.28
\\
Isaac Lab Mimic
& 12.1
& 1.12
& 1.38
& 1.96
& 1.25
\\
Isaac SkillGen
& 6.2
& 1.61
& 1.46
& 2.89
& 1.21
\\
\textbf{Kuafu (ours)}
& \textbf{57.3}
& \textbf{1.04}
& \textbf{0.67}
& \textbf{0.52}
& \textbf{0.28}
\\
\midrule
Human reference
& 0.0
& 3.06
& 1.59
& 2.81
& 1.25 \\
\end{tabular*}

\par\vspace{2pt}
{\footnotesize
\setlength{\tabcolsep}{0.5pt}
\begin{tabularx}{\linewidth}{@{}l*{10}{>{\raggedleft\arraybackslash}X}@{}}
\toprule
\rowcolor{black!7}
\multicolumn{11}{@{}l}{\textit{D. Downstream DP3 success (\%) (500 successful trajectories per task)}}\\
\midrule
Method
& T0
& T1
& T2
& T3
& T4
& T5
& T6
& T7
& T8
& Mean\\
\midrule
MimicGen
& 94.52
& 4.95
& 36.12
& 33.33
& 8.27
& 0.00
& 0.00
& 0.00
& 0.00
& 19.69\\

SkillGen
& \textbf{97.07}
& 3.91
& 30.65
& 62.00
& 8.05
& 0.00
& 0.00
& 0.00
& 0.26
& 22.44\\

IL Mimic
& 90.78
& 14.21
& 40.92
& 36.54
& \textbf{9.35}
& 0.00
& 0.00
& 0.00
& 0.00
& 21.31\\

IS Mimic
& 72.82
& 1.04
& 22.54
& 15.25
& 5.47
& 0.00
& 0.00
& 0.00
& 0.26
& 13.04\\

\textbf{Kuafu}
& 95.70
& \textbf{58.70}
& \textbf{55.50}
& \textbf{85.30}
& 0.78
& \textbf{15.80}
& \textbf{12.70}
& \textbf{69.00}
& \textbf{20.70}
& \textbf{46.02}\\
\bottomrule
\end{tabularx}%
}
\end{table}

We next assess trajectory efficiency and smoothness relative to the accepted MPC mentor (Panel~C).
Q95 yield is the fraction of evaluated rollouts that succeed and whose completion time, end-effector path length, action path length, and action second difference are each no greater than the corresponding task-specific 95th percentile among successful MPC trajectories.
The four diagnostics are normalized by task-specific medians from successful MPC trajectories ($\mathrm{MPC}_{50}$).
Kuafu achieves the highest Q95 yield, while its successful rollouts combine near-MPC completion times with shorter end-effector paths and lower action variation than competing generators.
Additionally, we measure motion smoothness using the spectral arc length (SPARC) of measured end-effector translational speed at Step~4 of \autoref{fig:pipeline}, where higher values indicate smoother motion~\cite{balasubramanian2015smoothness}.
Across all nine tasks, Kuafu achieves a task-balanced SPARC of $-3.14$, compared with $-4.44$ for human demonstrations, $-4.52$ for SkillGen, and $-4.67$ for MimicGen.
Results show that smoothness improvement can be seen in both manipulator's motion and action sequences.

Efficiency and smoothness alone do not show whether the data support a visual learner, so we test utility directly in downstream policy learning (Panel~D).
With a fixed 3D Diffusion Policy (DP3) learner and a matched budget of 500 successful trajectories per task, Kuafu yields data that improve visual-policy success by 23.6 percentage points over SkillGen, the strongest alternative.
The improvement also can be seen from task coverage: Kuafu alone yields nonzero downstream success on all nine tasks, including T5--T7, where every competing source yields none.
Kuafu therefore not only generates demonstrations faster than competing methods, but also produces more effective training data.

\begin{table}[!tb]

\centering
\caption{Visual policy performance and zero-shot transfer}
\label{tab:visual_policy_summary_main}
\scriptsize
\setlength{\tabcolsep}{1.15pt}
\renewcommand{\arraystretch}{1.05}

\begin{tabular*}{\linewidth}{@{\extracolsep{\fill}}lcccc@{}}
\toprule
\rowcolor{black!7}
\multicolumn{5}{@{}l}{%
\textit{A. Single-task visual-policy success (\%; same Kuafu data)}}\\
\midrule
\textbf{RGB}
& \makecell[c]{ACT~\citep{zhao2023act}\\35.40}
& \makecell[c]{Diffusion Policy\\\textbf{38.52}}
& \makecell[c]{smolVLA~\cite{shukor2025smolvla}\\26.73}
& \makecell[c]{ManiFlow-RGB~\cite{yan2025maniflow}\\34.09}\\
\cmidrule(lr){2-5}
\textbf{Point cloud}
& \makecell[c]{DP3\\46.02}
& \makecell[c]{FlowPolicy~\cite{zhang2025flowpolicy}\\21.01}
& \makecell[c]{ManiFlow-PC~\cite{yan2025maniflow}\\\textbf{46.46}}
& \makecell[c]{MP1~\cite{sheng2026mp1}\\12.94}\\
\end{tabular*}

\par\vspace{2pt}
\setlength{\tabcolsep}{0.5pt}
\begin{tabular*}{\linewidth}{
@{\extracolsep{\fill}}lrrrrrrrrrr@{}
}
\toprule
\rowcolor{black!7}
\multicolumn{11}{@{}l}{
\textit{B. Per-task maxima and state-controller success (\%)}
}\\
\midrule
& T0 & T1 & T2 & T3 & T4 & T5 & T6 & T7 & T8 & Mean\\
\midrule
RGB max
& 84.53
& 30.54
& 73.98
& 60.70
& \textbf{25.50}
& 10.10
& 12.10
& \textbf{73.00}
& \textbf{23.00}
& 43.72\\
PC max
& \textbf{98.38}
& \textbf{65.53}
& \textbf{86.28}
& \textbf{85.30}
& 10.05
& \textbf{15.80}
& \textbf{12.70}
& 69.00
& 20.70
& \textbf{51.53}\\
\midrule
State
& 98.48
& 95.78
& 98.01
& 84.62
& 58.80
& 69.70
& 87.25
& 95.21
& 50.43
& 82.03\\
\end{tabular*}

\par\vspace{2pt}
\setlength{\tabcolsep}{0.7pt}
\begin{tabular*}{\linewidth}{
@{\extracolsep{\fill}}lrrrrrrrrrr@{}
}
\toprule
\rowcolor{black!7}
\multicolumn{11}{@{}l}{
\textit{C. Multi-task VLA success (\%; one policy)}
}\\
\midrule
Policy & T0 & T1 & T2 & T3 & T4 & T5 & T6 & T7 & T8 & Mean\\
\midrule
SmolVLA~\cite{shukor2025smolvla}
& 5.87
& 0.26
& 12.94
& 2.08
& 0.00
& 0.00
& 0.00
& 0.00
& 0.00
& 2.35\\
WALL-X~\cite{zhai2025walloss}
& 22.22
& \textbf{1.56}
& 60.07
& \textbf{11.78}
& 0.00
& 0.00
& 0.00
& 10.51
& 0.00
& 11.79\\
$\pi_{0.5}$~\citep{black2025pi05}
& \textbf{26.61}
& 0.00
& \textbf{81.87}
& 0.00
& 4.17
& 4.17
& 0.00
& \textbf{69.17}
& 0.00
& \textbf{20.67}\\
\bottomrule
\end{tabular*}
\end{table}

\vspace{-3pt}
\subsection{Visual Learning and Zero-Shot Transfer}
\label{sec:exp_visual_transfer}

\autoref{tab:visual_policy_summary_main} presents the utility of Kuafu's trajectories after removing privileged state and stage inputs. 
For single-task policies (Panel~A), ManiFlow-PC preserves fine-grained point features and uses adaptive cross-attention for flow-based action generation, while DP3 conditions diffusion-based action sequences on a compact 3D representation.
Their gains over their RGB counterparts and other point-cloud methods suggest that explicitly encoding spatial relationships helps reproduce precise manipulation behaviors without recovering those relationships from images and point clouds alone.

Panel~B reports per-task maxima within each modality family; these represent optimal architectures for each task, not a single deployable oracle.
While maxima nearly match the state controller on T0 and exceed it on T3, the 15-stage T6 shows the largest gap of 74.55 points. These gaps, along with the multi-task results, reflect losses from observation conditioning, partial observability, and architectural constraints rather than state-based execution alone. 
With one policy across all nine tasks, $\pi_{0.5}$ reaches 20.67\% task-macro success, much higher than WALL-X and SmolVLA (Panel~C). SmolVLA prioritizes a compact, efficient architecture, WALL-X emphasizes tightly coupled language–action learning, and $\pi_{0.5}$ emphasizes transfer from heterogeneous robot and semantic pretraining.
With identical demonstrations, SmolVLA struggles even on tasks that the other two models learn successfully.
Although $\pi_{0.5}$ performs best overall, its advantage over WALL-X is concentrated in T2 and T7.
This pattern is consistent with differences in model capacity and pretrained representations affecting how effectively each learner uses the same supervision.
The shared failures further indicate that these learner-dependent gains improve selected behaviors, while the policies still struggle with longer tasks.

For physical deployment, we train DP3 on Kuafu demonstrations. We use 1,000 trajectories per task and deploy the resulting policies without real-world fine-tuning.
Across six tasks and three robot--gripper configurations, DP3 achieves 34.0\% success over 106 trials, with successful executions on every task. T4 also achieves a higher success rate in deployment than in simulation.
These experiments demonstrate that Kuafu-generated demonstrations provide usable supervision for visual policies deployed on physical hardware, closing the loop from SUN Program supervision through simulated demonstration generation to real-robot execution.

\section{Conclusions}
\label{sec:conclusions}

We presented SUN Programs---persistent executable task artifacts that inject MPC-verified semantics into policy learning---and Kuafu, an agentic harness that builds robot policies around these artifacts.
Kuafu achieves 82.03\% task-macro success on the primary evaluation, compared with 35.67\% for the matched sparse-reward baseline.
Ablations confirm the complementary roles of program-derived supervision, the verified behavioral prior, and bounded residual correction.
Furthermore, Kuafu-generated data yield 46.0\% success in fixed-volume DP3 training, support 8 single-task visual policies and 3 multi-task VLAs, and enable zero-shot transfer across 106 physical trials.
These results demonstrate how control-verified task knowledge can underpin both robust learned execution and observation-only visual policies.

\noindent \textbf{Scope and limitations.}
The current harness relies on rich object semantics and a finite typed-operator library, with a monotone stage monitor that advances through ordered sequences of stages without recovery.
Extending the operator vocabulary and supporting stage recovery are natural next steps.
Additionally, our physical evaluation assumes well-calibrated real-world sensors matching simulation conditions.  Currently, Kuafu does not handle moving obstacles or external disturbances during execution.

\noindent \textbf{Ethics Statement.} This work did not involve human-subject research. Human involvement was limited to teleoperation for generating a robot-control baseline in simulation. No information about the operators was collected; only robot trajectories and task-performance measurements were used.

\balance
\scriptsize
\bibliographystyle{IEEEtranN}
\bibliography{references}

\clearpage
\normalsize
\appendices
\clearpage
\onecolumn
\section{Positioning Relative to Related Work}
\label{app:sun_positioning}

Table~\ref{tab:related_positioning} compares task representations and the
capabilities they support in online execution, policy learning, and automated
task-to-policy workflows. Its ten criteria cover explicit task specifications,
stage structure, grounded geometry, control, verification, learning, and
execution diagnostics. Each entry indicates demonstrated support in the
specified role.

\begin{table}[H]
\centering
\begingroup

\definecolor{checkgreen}{HTML}{218739}
\definecolor{crossred}{HTML}{C62828}
\providecommand{\cmark}{}
\providecommand{\xmark}{}

\renewcommand{\cmark}{%
  {\fontsize{7.5}{8}\selectfont\color{checkgreen}\ding{51}}%
}
\renewcommand{\xmark}{%
  {\fontsize{7.5}{8}\selectfont\color{crossred}\ding{55}}%
}
\newcommand{\criterion}[2]{%
    {\fontsize{6.5}{8.0}\selectfont
     \bfseries
     \makecell[c]{#1\\#2}}%
}

\newcommand{\policycell}[1]{%
    {\fontsize{6.5}{8.0}\selectfont
     \makecell[c]{#1}}%
}
\fontsize{8.0}{10.0}\selectfont
\setlength{\tabcolsep}{1.75pt}
\renewcommand{\arraystretch}{1.12}

\begin{tabularx}{0.997\textwidth}{
@{}
>{\raggedright\arraybackslash}X
*{10}{>{\centering\arraybackslash}m{0.062\textwidth}}
@{}
}
\toprule

\makecell[l]{\textbf{Representative systems}}
& \criterion{Task}{spec.}
& \criterion{Stage}{aware}
& \criterion{Stage}{guards}
& \criterion{Structured}{geometry}
& \criterion{Direct}{control}
& \criterion{Pre-policy}{verification}
& \criterion{Policy}{learning}
& \criterion{Whole-task}{policy}
& \criterion{Spec. at}{execution}
& \criterion{Execution}{diag.} \\

\midrule

\multicolumn{11}{@{}l}{%
    \scriptsize\itshape
    A. Language-generated task representations used for online execution
}\\[-0.18em]

SayCan~\citep{ichter2023language}
& \cmark & \cmark & \cmark & \xmark
& \cmark & \xmark & \xmark
& \xmark & \cmark & \xmark \\

Code as Policies~\citep{liang2022codeaspolicies}
& \cmark & \cmark & \cmark & \xmark
& \cmark & \xmark & \xmark
& \xmark & \xmark & \xmark \\

ProgPrompt~\citep{singh2023progprompt}
& \cmark & \cmark & \cmark & \xmark
& \cmark & \xmark & \xmark
& \xmark & \xmark & \xmark \\

Language to Rewards~\citep{yu2023language}
& \cmark & \xmark & \xmark & \cmark
& \cmark & \xmark & \xmark
& \xmark & \xmark & \xmark \\

VoxPoser~\citep{huang2023voxposer}
& \cmark & \cmark & \xmark & \cmark
& \cmark & \xmark & \xmark
& \xmark & \xmark & \xmark \\

ReKep~\citep{huang2025rekep}
& \cmark & \cmark & \cmark & \cmark
& \cmark & \xmark & \xmark
& \xmark & \xmark & \xmark \\

OmniManip~\citep{pan2025omnomanip}
& \cmark & \cmark & \xmark & \cmark
& \cmark & \xmark & \cmark
& \cmark & \xmark & \xmark \\

CoRAL~\citep{coral2026}
& \cmark & \cmark & \cmark & \cmark
& \cmark & \xmark & \xmark
& \xmark & \xmark & \xmark \\

STeP~\citep{torshizi2026step}
& \cmark & \cmark & \cmark & \cmark
& \cmark & \xmark & \xmark
& \xmark & \cmark & \cmark \\

\addlinespace[1.4pt]
\multicolumn{11}{@{}l}{%
    \scriptsize\itshape
    B. Generated or learned task supervision used for policy learning
}\\[-0.18em]

Text2Reward~\citep{xie2023text2reward}
& \cmark & \cmark & \cmark & \xmark
& \xmark & \xmark & \cmark
& \xmark & \xmark & \xmark \\

Eureka / DrEureka~\citep{ma2024eureka,ma2024dreureka}
& \cmark & \xmark & \xmark & \xmark
& \xmark & \xmark & \cmark
& \xmark & \xmark & \xmark \\

DrS~\citep{Mu2024-eo}
& \xmark & \cmark & \cmark & \xmark
& \xmark & \xmark & \cmark
& \cmark & \xmark & \xmark \\

DEMO3~\citep{Escoriza2025-dj}
& \xmark & \cmark & \cmark & \xmark
& \cmark & \xmark & \cmark
& \cmark & \cmark & \xmark \\

IKER~\citep{patel2025iker}
& \cmark & \cmark & \cmark & \cmark
& \xmark & \xmark & \cmark
& \xmark & \cmark & \xmark \\

ReWiND~\citep{zhang2025rewind}
& \xmark & \xmark & \xmark & \xmark
& \xmark & \xmark & \cmark
& \xmark & \xmark & \xmark \\

SARM2 / SPIRAL~\citep{chen2026sarm2}
& \xmark & \cmark & \cmark & \xmark
& \xmark & \xmark & \cmark
& \cmark & \xmark & \xmark \\

\addlinespace[1.4pt]
\multicolumn{11}{@{}l}{%
    \scriptsize\itshape
    C. Automated task-to-policy workflows
}\\[-0.18em]

AnyTask~\citep{gong2025anytask}
& \cmark & \cmark & \xmark & \xmark
& \cmark & \cmark & \cmark
& \cmark & \xmark & \xmark \\

\midrule

\textbf{SUN Program / Kuafu (ours)}
& \cmark & \cmark & \cmark & \cmark
& \cmark & \cmark & \cmark
& \cmark & \cmark & \cmark \\

\bottomrule
\end{tabularx}
\endgroup

\caption{\textbf{Task representations and supported capabilities.}
\cmark{} denotes explicit and experimentally demonstrated support;
\xmark{} denotes support not demonstrated in the specified role.
\textbf{Task spec.}: an explicit, separately represented, inspectable task specification;
\textbf{Stage aware}: decomposes, sequences, labels, or conditions supervision on distinct task stages;
\textbf{Stage guards}: state-dependent conditions that govern stage transitions;
\textbf{Structured geometry}: task-relevant geometry represented through
residuals, robot or object frames and axes, 3D keypoint constraints, spatial
value maps, contact points, normals or regions, clearances, or formal metric predicates;
\textbf{Direct control}: the task representation is consumed directly by a
planner or controller;
\textbf{Pre-policy verification}: a generated task specification or reward is verified before policy learning;
\textbf{Policy learning}: the task representation directly supplies learning supervision;
\textbf{Whole-task policy}: one learned controller for the entire multi-stage task;
\textbf{Spec. at execution}: the task representation remains consulted during execution; and
\textbf{Execution diag.}: diagnostics localize term-level failures during execution.}
\label{tab:related_positioning}
\end{table}

The SUN Program serves as the shared task contract in the Kuafu harness:
physically verified relations remain available to guide policy learning and
monitor execution. This continuity preserves accepted task semantics across
the agent's tool calls.
\clearpage

\section{Additional SUN Implementation Details}
\label{app:sun_implementation}

This appendix specifies the grounding conventions, compiled operator
semantics, and policy interfaces used by the Kuafu harness. Each SUN relation
provides the MPC objective term, satisfaction predicate, and RL learning cost
introduced in Section~\ref{sec:problem}. The following sections describe how
these interfaces share grounded arguments and stage-entry references, how
the receiving application assigns tolerances, and how the learned policies
represent actions.

We use the main text's notation: $k\in\{1,\ldots,K\}$ indexes stages,
$j\in\{1,\ldots,m_k\}$ indexes the $m_k$ relations in stage $k$, and $z_t$
is the active stage at time $t$, with entry snapshot $\xi_{z_t}$.
After verification, policy learning and data production use the accepted
program $\Gamma^\star$.

\subsection{Grounding passes and active operator vocabulary}
\label{app:sun_grounding_vocabulary}

SUN Program formation uses three structured transformations. First, the instruction is
decomposed into \texttt{[Approach]}, \texttt{[Manipulate]}, and
\texttt{[Release]} subtasks. Each subtask contains one controller-stage-sized
change---one spatial change or one gripper-state change---together
with simultaneous pose or grasp requirements. Second, an atomic-grounding pass
resolves the current subtask into frame-, axis-, distance-, and
feedback-specific language. Third, the constraint-writing pass selects and
binds operators from $\mathcal L$ to produce one flat stage $\Gamma_k$ for
compilation. Appendix~\ref{app:sun_prompts} provides the prompt contracts for
these transformations and their diagnostic repair calls.

Grounding binds operators to names registered in $\mathcal F$. The harness
uses tool $+z$ as the approach axis and tool $+y$ as the finger axis. A top-down
approach makes tool $+z$ antiparallel to world $+z$; a level approach makes
those axes orthogonal while pointing tool $+z$ at the target; and an
axis-specific approach makes tool $+z$ antiparallel to the selected signed
object axis. For an elongated grasp target, tool $+y$ is orthogonal to its long
axis; for a box, it is orthogonal to the selected clamping axis. Rotational
symmetry is otherwise left unconstrained.

The grounding prompt distinguishes clearance from contact. For moving- and
reference-object extents $l_c$ and $l_r$ and a non-contact margin $m$, the
requested center separation is
\begin{equation}
d_{\mathrm{sep}}=\frac{l_c}{2}+\frac{l_r}{2}+m,
\qquad m\geq0.10\ \mathrm{m},
\label{app:eq:sun_prompt_clearance}
\end{equation}
with $m\geq0.15\,$m when the instruction requests a ``large'' clearance,
unless it provides a tighter value. Contact and grasp setup target the named
frame origin or object center and specify the intended contact feedback.
These separations define target geometry. The receiving application separately
assigns the numerical tolerances used to test satisfaction.

Table~\ref{app:tab:sun_term_api} lists the ten operations exposed by the active
constraint-writing prompt. The implementation names
\texttt{axis\_align}, \texttt{directional\_move}, and
\texttt{freeze\_orientation} correspond to \texttt{align\_axis},
\texttt{direction\_move}, and \texttt{fix\_orientation} in
Section~\ref{sec:agent_interface}, respectively. The table retains the
implementation names used by the prompt API. The
\texttt{controlled\_entity} is either a registered robot frame or a gripped
object propagated through a fixed grasp transform. A reference basis supplies
coordinates and axes, while a reference origin supplies the corresponding
pose. Distances are in metres and angles are in radians. For all seven
geometric operators, the receiving application supplies
$\epsilon_p$ or $\epsilon_\theta$ according to the operator and profile.
These are the positional and angular instances of the relation tolerance
$\epsilon_{k,j}$ and enter the compiled predicates through receiver-owned fields, which
are omitted from the prompt's output schema. Per-term weights are assigned
through a separate scalar-selection interface. Compilation initializes
$w_{k,j}=1$ unless that interface supplies a value; later
calibration may change this weight without changing the accepted relation or
its satisfaction boundary.

\begin{table}[H]
\centering
\caption{Prompt-visible SUN operators, their grounded arguments, and their
implemented semantics. The receiving application adds numerical predicate
tolerances before compilation; these fields are absent from the agent's
output schema.}
\label{app:tab:sun_term_api}
\small
\begin{tabularx}{\textwidth}{@{}p{0.20\textwidth}p{0.30\textwidth}X@{}}
\toprule
Operation & Principal parameters & Implemented meaning\\
\midrule
\texttt{reach\_point} & controlled entity/target, reference basis/origin,
offset & Reach one frame-relative point; the intervening path is
unconstrained.\\
\texttt{directional\_move} & controlled entity/target, reference axis, $D>0$
& Translate by $D$ along one signed axis while suppressing lateral
drift.\\
\texttt{axis\_align} & controlled and reference signed axes, relation
& Enforce codirectional, antiparallel, bipolar, or orthogonal
axes.\\
\texttt{point\_toward} & controlled signed axis, target point
& Aim an axis at a frame-relative point.\\
\texttt{directional\_rotate} & reference signed axis, angle $\vartheta$
& Rotate from the stage-entry orientation by a specified
amount.\\
\texttt{freeze\_position} & controlled entity/target & Preserve
the position captured on stage entry.\\
\texttt{freeze\_orientation} & controlled entity/target &
Preserve the orientation captured on stage entry.\\
\texttt{open\_gripper} & none & Require the released/open mechanical state.\\
\texttt{close\_gripper} & none & Require mechanical finger closure without
claiming a grasp.\\
\texttt{object\_grasped} & none & Require physical holding and retain it during
later motion.\\
\bottomrule
\end{tabularx}
\end{table}

\subsection{Receiver-owned tolerances and prompt filtering}
\label{app:sun_receiver_policy}

A new or corrected agent response specifies operators, grounded targets,
frames, distances, offsets, rotation amounts, and contact conditions. Before
compilation, the receiving application assigns $\epsilon_{k,j}$ from the
operator, stage semantics, effort conditions, and scene dimensions. This
deterministic assignment establishes numerical satisfaction boundaries and
records the selected profile and per-component decisions. The response schema
excludes \texttt{tolerance\_m} and \texttt{tolerance\_rad}; any unsolicited
values for these fields are ignored.

The implementation provides two deterministic initialization profiles.
Fresh MPC formulation caps
positional tolerances at $0.01\,$m, or $0.02\,$m for a nonzero-offset reach
target, and sets the nominal angular field to $0.20\,$rad. Learning preparation
instead derives positional fields in $[0.01,0.05]\,$m and angular fields in
$[0.30,0.50]\,$rad according to geometry and interaction role. Saved MPC
programs retain their recorded values on replay. Learning values are set before
weight selection and remain fixed during later scalar tuning. These profiles
use different numerical boundaries for MPC and learning preparation. In
contrast, the reported experiments follow the fixed-boundary protocol in
Algorithm~\ref{alg:kuafu}, which retains the accepted program's satisfaction
boundaries throughout policy learning.

Before each agent call, an outgoing-context filter removes receiver-owned
setting fields and recognized tolerance annotations from prompt copies,
including embedded JSON, correction diagnostics, and text history. The filter
preserves geometric targets, numerical geometry, and unrecognized prose;
the receiving application enforces tolerance assignment after the response.
Stored programs and image attachments remain unchanged. Initial and
BC-informed weight calls return \texttt{Diagnosis}, \texttt{Reasoning}, and
\texttt{Weights}; an unsolicited \texttt{Parameters} field is ignored. Saved
parameter artifacts retain the tolerance-assignment policy version and
per-stage decisions for traceability.

\subsection{Common frame and compiled-cost conventions}
\label{app:sun_compiled_semantics}

Let ${}^{W}p_A$ and ${}^{W}q_A$ be the world position and scalar-first unit
quaternion of frame $A$, let $R(q)\in SO(3)$ be its rotation matrix, and let a
superscript $0$ denote a value stored in the current stage-entry snapshot
$\xi_k$. The world superscript and time argument are omitted when unambiguous;
thus $p_c$ and $q_c$ denote the controlled entity's world pose.
A hat denotes unit normalization, $\otimes$ quaternion multiplication, and
$q^*$ quaternion conjugation. Rotation angles use $\vartheta$, reserving
$\theta$ for the Stage-BC parameters as in the main text.
For local offset $\delta$, reference origin $O$, and reference robot
frame $F$, \texttt{reach\_point} constructs
\begin{equation}
{}^{W}p^\star=
\begin{cases}
{}^{W}p_O(t)+\delta, & \text{world basis},\\
{}^{W}p_O(t)+R({}^{W}q_O(t))\delta, & \text{object basis},\\
{}^{W}p_F^0+R({}^{W}q_F^0)\delta, & \text{robot-frame basis}.
\end{cases}
\label{app:eq:sun_target_point}
\end{equation}
In the world branch, $\delta$ is a world-coordinate offset from the selected
origin; it is an absolute world point only when $O$ is the reserved world
frame.

When the controlled entity is an object held by end effector $E$, the entry
snapshot defines the fixed grasp transform
\begin{equation}
q_{O/E}=({}^{W}q_E^0)^*\otimes{}^{W}q_O^0,
\qquad
r_{O/E}=R({}^{W}q_E^0)^\top
({}^{W}p_O^0-{}^{W}p_E^0),
\label{app:eq:sun_grasp_transform}
\end{equation}
and the optimized end-effector state implies
\begin{equation}
{}^{W}q_O={}^Wq_E\otimes q_{O/E},
\qquad
{}^{W}p_O={}^Wp_E+R({}^{W}q_E)r_{O/E}.
\label{app:eq:sun_gripped_object_pose}
\end{equation}
The same position and orientation operations can therefore control the end
effector or a rigidly attached object without adding object pose to the MPC
decision state.

The implementation uses the quadratic MPC map and bounded learning map
\begin{equation}
Q(v)=\frac{1}{2}v^\top v,
\qquad
B_a(r)=\frac{r^2}{r^2+a^2},
\quad 0\leq B_a(r)<1.
\label{app:eq:sun_cost_maps}
\end{equation}
Here $a>0$ is the map's scale parameter. The rational map $B_a$ bounds the learning cost, with lower
$c^{\mathrm{rl}}$ indicating better satisfaction. The following equations
instantiate the three compiled interfaces in Eq.~\eqref{eq:sun_compile} for
each operator. Operator subscripts replace $(k,j)$ for readability:
for example, $\phi^{\mathrm{mpc}}_{\mathrm{reach}}$,
$P_{\mathrm{reach}}$, and $c^{\mathrm{rl}}_{\mathrm{reach}}$ denote
$\phi^{\mathrm{mpc}}_{k,j}$, $P_{k,j}$, and $c^{\mathrm{rl}}_{k,j}$ for a
point-reaching relation. Predicates use the same indicator notation
$\mathbf{1}[\cdot]$ as the main text.

\subsubsection{Translation operators}

For \texttt{reach\_point}, let $p_c$ be the controlled robot-frame position or
the composed object position from Eq.~\eqref{app:eq:sun_gripped_object_pose}.
Then
\begin{equation}
\begin{aligned}
e_{\mathrm{reach}}&=p_c-{}^{W}p^\star,
&d_{\mathrm{reach}}&=\lVert e_{\mathrm{reach}}\rVert_2,\\
\phi^{\mathrm{mpc}}_{\mathrm{reach}}&=Q(e_{\mathrm{reach}}),
&P_{\mathrm{reach}}&=\mathbf{1}[d_{\mathrm{reach}}\leq\epsilon_p],\\
c^{\mathrm{rl}}_{\mathrm{reach}}&=B_{0.5}(d_{\mathrm{reach}}).&&
\end{aligned}
\label{app:eq:sun_reach_point}
\end{equation}
The tolerance $\epsilon_p$ sets the satisfaction boundary; the MPC objective
and RL learning cost retain the forms above.

For \texttt{directional\_move}, resolve the signed basis axis as world unit
vector $\widehat a$, let $D>0$ be the requested travel, and measure from the
controlled entity's entry position $p_c^0\in\xi_k$:
\begin{align}
\Delta p&=p_c-p_c^0,
&d_\parallel&=\widehat a^\top\Delta p,
&e_\parallel&=d_\parallel-D,\nonumber\\
d_\perp&=\Delta p-\widehat a d_\parallel,
&m_{\mathrm{move}}&=
\sqrt{e_\parallel^2+\tfrac12\lVert d_\perp\rVert_2^2}.
\label{app:eq:sun_directional_move_error}
\end{align}
Its compiled views are
\begin{equation}
\begin{aligned}
\phi^{\mathrm{mpc}}_{\mathrm{move}}
&=\tfrac12(e_\parallel^2+\lVert d_\perp\rVert_2^2),\\
P_{\mathrm{move}}&=\mathbf{1}[m_{\mathrm{move}}\leq\epsilon_p],
\qquad
c^{\mathrm{rl}}_{\mathrm{move}}=B_{D/2}(m_{\mathrm{move}}).
\end{aligned}
\label{app:eq:sun_directional_move}
\end{equation}
All three views use the same entry position, travel direction, and requested
distance. Their numerical penalties differ: the predicate and RL learning
cost weight squared lateral deviation by one half within
$m_{\mathrm{move}}^2$, while the MPC objective uses its full value.

\subsubsection{Orientation operators}

For selected signed canonical axes $\widehat e_i$ and $\widehat e_\ell$, define
\begin{equation}
\widehat a_c=\widehat{R({}^{W}q_c)\widehat e_i},
\qquad
\widehat a_r=\widehat{R({}^{W}q_r)\widehat e_\ell},
\qquad
s=\widehat a_c^\top\widehat a_r,
\label{app:eq:sun_axis_vectors}
\end{equation}
where the reference axis may instead be a fixed world direction. The
\texttt{axis\_align} relation chooses
\begin{equation}
\begin{array}{c|c|c}
\text{relation}&\eta(s)&\bar\epsilon(\epsilon_\theta)\\ \hline
\text{codirectional}&1-s&1-\cos\epsilon_\theta\\
\text{antiparallel}&1+s&1-\cos\epsilon_\theta\\
\text{bipolar}&1-|s|&1-\cos\epsilon_\theta\\
\text{orthogonal}&\sqrt{s^2+10^{-18}}&\sin\epsilon_\theta
\end{array}
\label{app:eq:sun_axis_alignment_errors}
\end{equation}
and compiles it as
\begin{equation}
\begin{aligned}
\phi^{\mathrm{mpc}}_{\mathrm{align}}&=Q(\eta),
&P_{\mathrm{align}}&=\mathbf{1}[\eta\leq\bar\epsilon],\\
c^{\mathrm{rl}}_{\mathrm{align}}&=
\begin{cases}
B_{0.25}(\eta/2),&\text{co-, anti-, or bipolar},\\
B_{0.25}(|\eta|),&\text{orthogonal}.
\end{cases}
\end{aligned}
\label{app:eq:sun_axis_alignment}
\end{equation}

For \texttt{point\_toward}, the selected axis aims from the end effector toward
the target point:
\begin{equation}
\widehat d=\widehat{{}^{W}p^\star-{}^{W}p_E},
\qquad
\eta_{\mathrm{point}}=1-\widehat a_c^\top\widehat d,
\label{app:eq:sun_point_toward_error}
\end{equation}
\begin{equation}
\begin{aligned}
\phi^{\mathrm{mpc}}_{\mathrm{point}}&=Q(\eta_{\mathrm{point}}),\\
P_{\mathrm{point}}&=
\mathbf{1}[\eta_{\mathrm{point}}\leq1-\cos\epsilon_\theta],
\qquad
c^{\mathrm{rl}}_{\mathrm{point}}=B_{0.25}(\eta_{\mathrm{point}}/2).
\end{aligned}
\label{app:eq:sun_point_toward}
\end{equation}
The MPC interface binds the controlled axis to a robot frame and constructs
the target as ${}^{W}p_O(t)+\delta$, treating the offset as world coordinates
regardless of the supplied basis. The batched predicate additionally supports
a gripped object's orientation for $\widehat a_c$. In both cases,
$\widehat d$ originates at the configured end effector.

For \texttt{directional\_rotate}, let $\vartheta>0$ be the requested angle,
$\widehat a$ the signed axis in the selected basis, and $q_F^0$ the world
orientation of its reference frame stored in $\xi_k$. Then
\begin{equation}
q_\Delta=\left(\cos\frac{\vartheta}{2},
\widehat a\sin\frac{\vartheta}{2}\right),
\qquad
{}^{W}q_\Delta=
\begin{cases}
q_F^0\otimes q_\Delta\otimes(q_F^0)^*,
&\text{object or robot-frame basis},\\
q_\Delta,&\text{world basis},
\end{cases}
\label{app:eq:sun_directional_rotation_target}
\end{equation}
and, for controlled entry orientation $q_c^0$,
\begin{equation}
\begin{aligned}
q^\star&={}^{W}q_\Delta\otimes q_c^0,
&e_q&=q_c-q^\star,\\
\phi^{\mathrm{mpc}}_{\mathrm{rotate}}&=Q(e_q),
&P_{\mathrm{rotate}}&=\mathbf{1}[\lVert e_q\rVert_2<\epsilon_\theta],\\
c^{\mathrm{rl}}_{\mathrm{rotate}}&=B_{0.25}(\lVert e_q\rVert_2).&&
\end{aligned}
\label{app:eq:sun_directional_rotation}
\end{equation}
This predicate applies its nominal angular tolerance to quaternion-coordinate
distance, rather than to a geodesic angle; the distance is sensitive to the
equivalent quaternion signs $q$ and $-q$. The formation prompt places
\texttt{axis\_align} relations involving the rotating frame in an earlier
approach stage and the finite rotation in the subsequent manipulation stage.
These relations therefore occupy separate stages.

\subsubsection{Entry invariants and gripper operators}

The freeze operators preserve poses captured in the stage-entry snapshot.
Position freezing uses
\begin{equation}
\begin{aligned}
e_f^p&=p_c(t)-p_c^0,
&\phi^{\mathrm{mpc}}_{f,p}&=Q(e_f^p),\\
P_{f,p}&=\mathbf{1}[\lVert e_f^p\rVert_2\leq\epsilon_p],
&c^{\mathrm{rl}}_{f,p}&=B_{0.25}(\lVert e_f^p\rVert_2).
\end{aligned}
\label{app:eq:sun_freeze_position}
\end{equation}
For orientation, the predicate and learning path use the sign-invariant
geodesic distance
\begin{equation}
\begin{aligned}
\vartheta_f&=2\arccos\!\left(
\operatorname{clip}(|\widehat q_c(t)^\top\widehat q_c^0|,0,1)\right),\\
P_{f,q}&=\mathbf{1}[\vartheta_f\leq\epsilon_\theta],
\qquad
c^{\mathrm{rl}}_{f,q}=B_{0.25}(\vartheta_f).
\end{aligned}
\label{app:eq:sun_freeze_orientation_predicate}
\end{equation}
The MPC objective uses
\begin{equation}
\phi^{\mathrm{mpc}}_{f,q}=Q(q_E-q_E^0)
\label{app:eq:sun_freeze_orientation_mpc}
\end{equation}
and evaluates orientation changes at the configured end-effector state frame,
independently of the operation's controlled-entity and target arguments.

For normalized optimizer gripper state $g=(g_L,g_R)$,
\begin{equation}
g_{\mathrm{closed}}=(0,0),\quad g_{\mathrm{open}}=(1,1),\quad
\phi_\sigma^{\mathrm{mpc}}=Q(g-g_\sigma),\quad
c_\sigma^{\mathrm{rl}}=B_1(\lVert g-g_\sigma\rVert_2),
\label{app:eq:sun_gripper_costs}
\end{equation}
for gripper mode $\sigma\in\{\mathrm{closed},\mathrm{open}\}$.
Measured scalar openness $o_{\mathrm{grip}}$ is normalized from the $n_g$
finger-joint positions $q_\ell$ and their calibrated limits:
\begin{equation}
o_{\mathrm{grip}}=\frac{1}{n_g}\sum_{\ell=1}^{n_g}
\frac{q_\ell-q_\ell^{\mathrm{closed}}}
{q_\ell^{\mathrm{open}}-q_\ell^{\mathrm{closed}}},
\quad
P_{\mathrm{close}}=\mathbf{1}[o_{\mathrm{grip}}\leq0.1],
\quad
P_{\mathrm{open}}=\mathbf{1}[o_{\mathrm{grip}}\geq0.9].
\label{app:eq:sun_gripper_predicates}
\end{equation}

Physical holding is checked using finger contact and openness in addition to
the close-gripper command. Let $F_L,F_R$ be the
filtered finger-contact magnitudes, $F_{\max}$ the configured maximum
finger-force scale, and $o_{\max}$ the maximum grasp openness. Define
\begin{align}
o_c&=\operatorname{clip}(o_{\mathrm{grip}},0.03,o_{\max}),
&\widetilde o_{\mathrm{grip}}&=\frac{o_c-0.03}{o_{\max}-0.03},\nonumber\\
F_{\mathrm{req}}(o_{\mathrm{grip}})&=F_{\max}(0.1+0.6\widetilde o_{\mathrm{grip}}),
&P_{\mathrm{width}}&=\mathbf{1}[0.1<o_{\mathrm{grip}}\leq o_{\max}],\nonumber\\
P_{\mathrm{both}}&=\mathbf{1}[F_L\geq F_{\mathrm{req}}]\land
\mathbf{1}[F_R\geq F_{\mathrm{req}}],\nonumber\\
P_{\mathrm{one}}&=(\mathbf{1}[F_L\geq10]\oplus\mathbf{1}[F_R\geq10])
\land\mathbf{1}[o_{\mathrm{grip}}<0.3]\land P_{\mathrm{width}}.
\label{app:eq:sun_grasp_components}
\end{align}
The robot-specific predicate is
\begin{equation}
P_{\mathrm{grasp}}=
\begin{cases}
(P_{\mathrm{both}}\land P_{\mathrm{width}})\lor P_{\mathrm{one}},&\text{Boston Franka or Kinova Robotiq},\\
\mathbf{1}[F_L\geq5]\land\mathbf{1}[F_R\geq5]\land P_{\mathrm{width}},
&\text{Slichter Franka Panda}.
\end{cases}
\label{app:eq:sun_object_grasped}
\end{equation}
The Boston Franka and Kinova Robotiq predicates include the one-finger
exception $P_{\mathrm{one}}$, whereas the Slichter Franka Panda predicate requires contact
at both fingers. The MPC objective and RL learning cost for
\texttt{object\_grasped} reuse the close-gripper costs in
Eq.~\eqref{app:eq:sun_gripper_costs}; its satisfaction predicate additionally
checks physical holding. By default, a stage containing this operation must
remain satisfied for five consecutive frames before advancing. Later
pulling, carrying, and reorientation stages retain the operation to monitor
continued holding.

\subsection{Condition labels and valid term composition}
\label{app:sun_term_composition}

Condition labels attach satisfaction or feedback semantics to an operation.
\texttt{geometry} selects its geometric predicate;
\texttt{gripper\_opened} and \texttt{gripper\_closed} describe finger state;
and \texttt{manipulator\_high\_effort} and
\texttt{manipulator\_low\_effort} describe arm/TCP interaction rather than
finger squeeze. For exponentially filtered arm effort and a configured
high-effort threshold $E_{\mathrm{high}}$,
\begin{equation}
\bar\tau_t=0.2\tau_t+0.8\bar\tau_{t-1},
\qquad
E_t=\sum_{\ell=1}^{7}|\bar\tau_{t,\ell}|,
\qquad
P_{\mathrm{low}}=\mathbf{1}[E_t\leq10],
\label{app:eq:sun_effort_low}
\end{equation}
and sustained high effort is
\begin{equation}
h_t^E=
\begin{cases}
\min(h_{t-1}^E+1,5),&E_t>E_{\mathrm{high}},\\
0,&E_t\leq E_{\mathrm{high}},
\end{cases}
\qquad
P_{\mathrm{high}}=\mathbf{1}[h_t^E\geq5].
\label{app:eq:sun_effort_high}
\end{equation}
Effort labels contribute to satisfaction checks through local stopping
conditions within a required term, rather than through standalone MPC
objectives or RL learning costs. For a relation that permits high effort as a
local stopping alternative, its compiled predicate and the stage-completion
condition are
\begin{equation}
P_{k,j}(t)=P^{\mathrm{geometry}}_{k,j}(t)\lor P_{\mathrm{high}}(t),
\qquad
D_k(t)=\bigwedge_{j=1}^{m_k}P_{k,j}(t).
\label{app:eq:sun_local_condition_or}
\end{equation}
Here $D_k$ is the conjunction of the relation predicates. As in the main text,
the monitor evaluates $D_{z_t}(t+1)$ after applying $u_t$ and before updating
the active stage. A stage contains at most one of
\texttt{reach\_point}, \texttt{directional\_move}, and
\texttt{freeze\_position}; a freeze operation cannot substitute for a new
pose goal; and \texttt{close\_gripper} and \texttt{object\_grasped} cannot
co-occur. \texttt{directional\_move} is limited to manipulation or release
stages and requires an explicit direction and positive distance. When a
required clause falls outside $\mathcal L$, the agent records that clause as
unsupported and omits it from the executable terms. The runtime also supports
legacy \texttt{reach\_distance} and \texttt{free\_move} operators for replaying
older saved programs; the active prompt uses the vocabulary in
Table~\ref{app:tab:sun_term_api}.

At run time, diagnostics retain the source stage, bound frames, axis,
tolerance, measured residual, and satisfaction margin. After executing one
action, the monitor observes the resulting state, evaluates the current-stage
predicates and independent external outcome, computes learning and diagnostic
signals, and then updates the active stage. Each new stage captures its
reference poses after the guarded transition, preserving the evaluation
order used during verification, learning, and data production.

\subsection{Policy and action implementation}
\label{app:sun_policy_implementation}

Both learned state-policy components use feed-forward multilayer
perceptrons. The Stage-BC prior $\pi_\theta(o_t,z_t)$ is a deterministic MLP,
and the residual policy $\pi_\psi(o_t,z_t)$ is trained with PPO and generalized
advantage estimation using a feed-forward actor--critic. Here $o_t$ is the
privileged observation, distinct from scalar gripper openness
$o_{\mathrm{grip}}$. The absolute action $u_t$ in Eq.~\eqref{eq:residual_policy}
has the implementation representation
\begin{equation}
u_t=
\begin{bmatrix}
p_t^\top & (R_t^{6\mathrm D})^\top & g_t^\top
\end{bmatrix}^{\!\top}
\in\mathbb R^{11},
\label{app:eq:sun_absolute_action}
\end{equation}
where $p_t\in\mathbb R^3$ is the Cartesian position target,
$R_t^{6\mathrm D}\in\mathbb R^6$ is the continuous six-dimensional rotation
representation of the orientation target, and $g_t\in\mathbb R^2$ contains
the two finger targets used to encode the gripper command.

Stage-BC predicts chunks of ten actions and is trained with normalized
mean-squared error:
\begin{equation}
\mathcal L_{\mathrm{BC}}(\theta)
=\frac{1}{N}\sum_{n=1}^{N}
\left\|D_u^{-1}(\widehat U_n(\theta)-U_n)\right\|_2^2,
\label{app:eq:sun_stage_bc_loss}
\end{equation}
where $N$ is the number of training chunks, $U_n$ stacks the target actions,
and $\widehat U_n(\theta)$ is the chunk predicted from the privileged
observation and active stage. The diagonal matrix $D_u$ contains the
action-normalization scales repeated across the chunk. Chunks never cross a
stage boundary; when fewer than ten actions remain, the last action in that
stage is repeated as padding. The first predicted action is
$\pi_\theta(o_t,z_t)$; its bounded composition with the residual gives
$u_t=\pi_{\mathrm{pri}}(o_t,z_t)$ in Eq.~\eqref{eq:residual_policy}, after which
the SUN monitor is reevaluated.

Appendix~\ref{app:sun_bc_capacity} reports the architectures and temporal
targets evaluated in the task-specific BC sweep. Reproducing a final Kuafu
controller also requires its selected architecture, optimizer and epoch
schedule, normalization constants, PPO/GAE hyperparameters, residual bounds,
reward weights, MPC horizon and control rate, solver tolerances, and
domain-randomization ranges. Complete per-task values for those final
controller settings are not reported here; the sweep configurations describe
the supplemental BC comparisons.

\clearpage
\section{BC-Prior Capacity and Temporal-Target Sweep}
\label{app:sun_bc_capacity}

Table~\ref{app:tab:capacity_chunk_stride_sweep} reports the supplemental
BC-prior sweep on T6 (three-cube stacking) and T8 (cup hanging). It varies the
feed-forward policy architecture, supervised action-chunk length $C$, and
spacing $s_a$ between actions in the training target. This evaluates BC alone,
not the complete residual-RL controller. Model names retain the task-specific
sweep identifiers: their hidden-layer widths are stated explicitly, and the
suffix \texttt{3} denotes a three-hidden-layer variant, not an action stride.

\begin{table}[H]
\centering
\caption{BC policy-capacity, action-chunk, and stride sweep. Hidden-layer
widths are listed in input-to-output order. Rollout success rates (\%) are
grouped by supervised action-chunk length $C$ and sampling stride $s_a$.
Dashes retain unreported configurations, not zero success; bold marks the
best reported entry for each task.}
\label{app:tab:capacity_chunk_stride_sweep}
\begingroup
\small
\setlength{\tabcolsep}{3pt}
\renewcommand{\arraystretch}{1.12}
\begin{tabular*}{\linewidth}{@{\extracolsep{\fill}}ll*{6}{r}@{}}
\toprule
\multirow{2}{*}{Model} & \multirow{2}{*}{Hidden-layer widths}
& \multicolumn{2}{c}{$C=10$}
& \multicolumn{2}{c}{$C=15$}
& \multicolumn{2}{c@{}}{$C=30$}\\
\cmidrule(lr){3-4}\cmidrule(lr){5-6}\cmidrule(l){7-8}
& & $s_a=1$ & $s_a=3$ & $s_a=1$ & $s_a=3$ & $s_a=1$ & $s_a=3$\\
\midrule
\rowcolor{black!6}[0pt][0pt]
\multicolumn{8}{@{}l}{\textit{T6: Stack 3 Cubes}}\\
\texttt{ff} & 2048, 1024, 512, 1024, 2048 & 0.6 & 2.7 & 1.5 & 3.4 & 3.5 & 3.0\\
\texttt{fff} & 4096, 2048, 1024, 2048, 4096 & 7.3 & 7.4 & 10.1 & 14.2 & 10.2 & 11.3\\
\texttt{fff3} & 2048, 1024, 2048 & 2.0 & 5.0 & 3.1 & 4.9 & 4.1 & 0.5\\
\texttt{ffff} & 8192, 4096, 2048, 4096, 8192 & 16.1 & 23.6 & \textbf{30.0} & 27.8 & 22.5 & 18.4\\
\texttt{ffff3} & 4096, 2048, 4096 & 4.7 & 6.8 & 6.3 & 6.4 & 6.9 & 3.2\\
\texttt{fffff} & 16384, 8192, 4096, 8192, 16384 & 7.4 & 17.3 & 16.9 & 25.1 & 26.5 & 0.1\\
\texttt{fffff3} & 8192, 4096, 8192 & 6.1 & 10.3 & 3.0 & 11.6 & 5.8 & 8.8\\
\texttt{ffffff} & 32768, 16384, 8192, 16384, 32768 & 0.1 & 0.3 & 0.1 & 6.7 & 11.3 & 4.5\\
\texttt{ffffff3} & 16384, 8192, 16384 & 3.8 & 1.1 & 2.2 & 6.3 & 3.2 & 3.5\\
\midrule
\rowcolor{black!6}[0pt][0pt]
\multicolumn{8}{@{}l}{\textit{T8: Hang Cup}}\\
\texttt{ff} & 2048, 1024, 512, 1024, 2048 & 3.2 & 2.0 & 4.9 & 1.3 & 1.6 & 0.0\\
\texttt{fff} & 4096, 2048, 1024, 2048, 4096 & 17.7 & 3.4 & 9.1 & 3.3 & 5.4 & 2.1\\
\texttt{fff3} & 2048, 1024, 2048 & 12.5 & 8.6 & 13.5 & 6.2 & 8.8 & 4.3\\
\texttt{ffff} & 8192, 4096, 2048, 4096, 8192 & \textbf{22.5} & -- & 17.9 & -- & -- & --\\
\texttt{ffff3} & 4096, 2048, 4096 & 19.2 & -- & 17.8 & -- & -- & --\\
\texttt{fffff} & 16384, 8192, 4096, 8192, 16384 & 20.2 & -- & 12.7 & -- & 3.7 & --\\
\texttt{ffffff} & 32768, 16384, 8192, 16384, 32768 & 14.8 & -- & 16.6 & -- & 7.6 & --\\
\bottomrule
\end{tabular*}
\endgroup
\end{table}

The sweep reveals task-dependent effects of capacity and temporal targets.
Among the reported configurations, the intermediate-sized \texttt{ffff}
architecture achieves the highest success on both tasks: 30.0\% on T6 with
$C=15$, $s_a=1$, and 22.5\% on T8 with $C=10$, $s_a=1$. Increasing width
beyond this architecture reduces the best observed success on each task,
while the different optimal chunk lengths indicate that temporal targets
should be selected for the task. The stride comparison further distinguishes
the two tasks: increasing $s_a$ from 1 to 3 improves 19 of the 27 matched T6
configurations, but lowers success in all nine reported matched T8
configurations. The highest-success configuration for each task nevertheless
uses $s_a=1$.

These results support task-specific selection of the BC prior's architecture
and temporal targets. Even the strongest tested BC configurations retain
substantial execution failures on these two tasks. The sweep therefore
characterizes the starting behavior available to subsequent residual
refinement, complementing the complete-system ablations in
Section~\ref{sec:exp_control_learning}.

\clearpage
\section{Taskwise Data-Generation Throughput and Amortization}
\label{app:sun_data_economics}

\begin{figure}[H]
  \centering
  \includegraphics[width=0.96\textwidth]{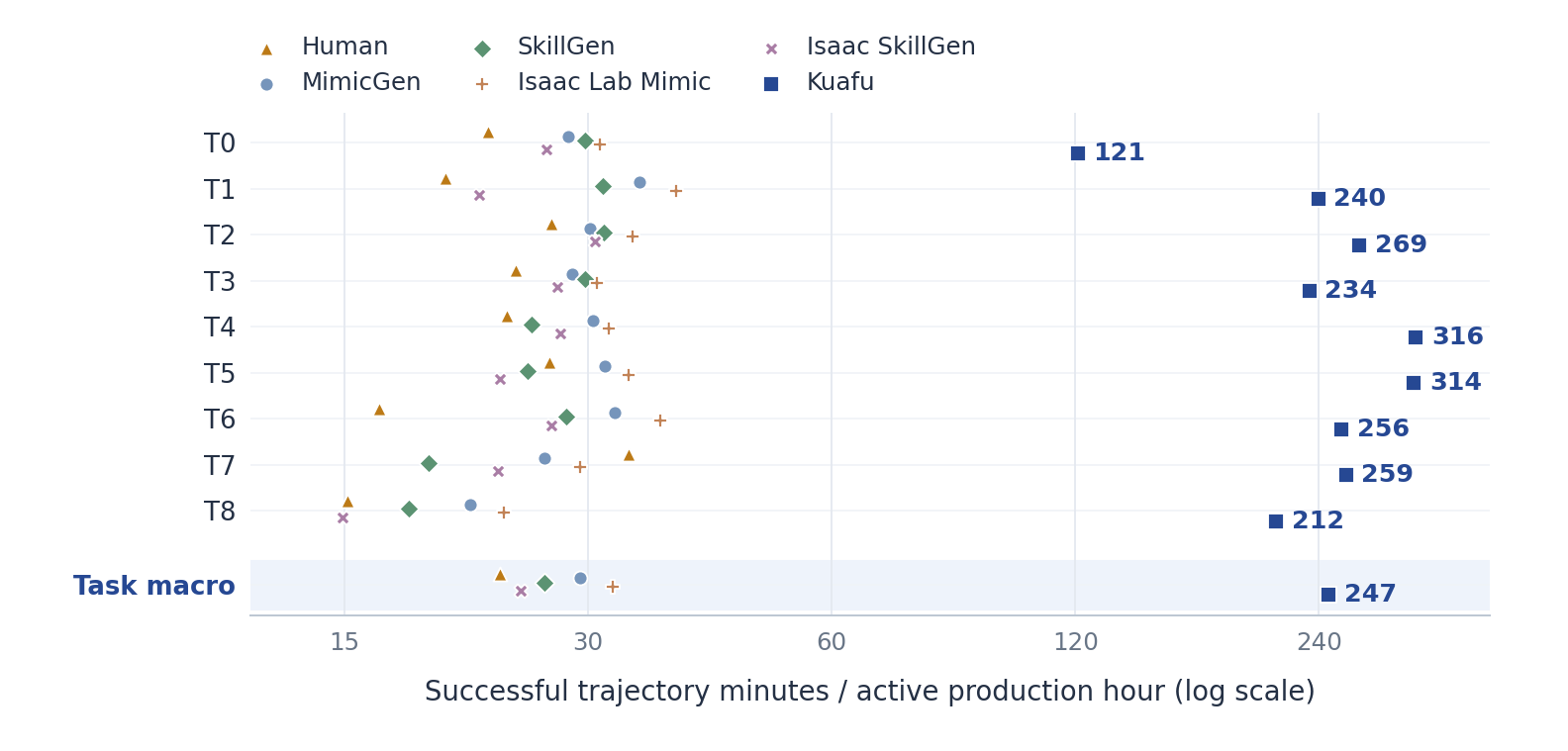}
  \par\smallskip
  \includegraphics[width=0.96\textwidth]{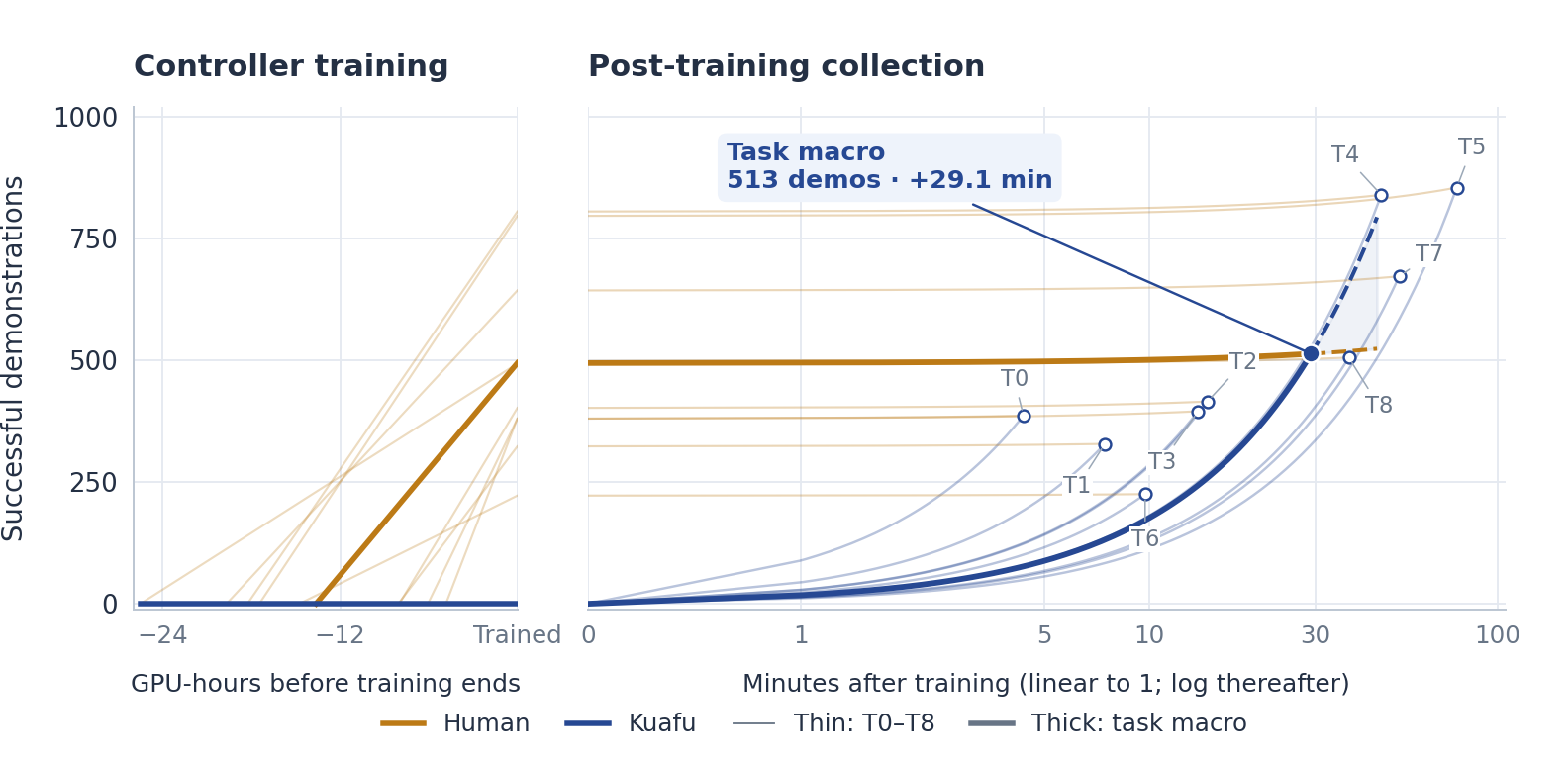}
  \caption{Taskwise data-generation throughput and startup break-even.
  \textbf{Top:} successful trajectory minutes per active production hour for
  tasks T0--T8 and their unweighted task-macro average; the horizontal axis is
  logarithmic. Production hours refer to active human collection for the
  human reference and GPU production for the simulated generators.
  \textbf{Bottom:} cumulative successful demonstrations reconstructed from
  the reported controller-training times and time-break-even values. Faint curves show
  individual tasks; thick curves show the reported task-macro summary, and
  dashed extensions project continued collection at the same rates.
  The horizontal axis uses GPU-hours before training ends and minutes afterward;
  the post-training axis is linear up to one minute and logarithmic
  thereafter. The macro break-even is 513 demonstrations after 29.1 minutes
  of post-training collection.}
  \label{app:fig:sun_data_economics}
\end{figure}

Figures~\ref{app:fig:sun_data_economics} and~\ref{app:fig:sun_cost_parity}
expand the throughput and break-even summaries in
Table~\ref{tab:data_engine_main}, panels A and B, with taskwise results from
the same evaluation. They distinguish steady-state production from the
upfront time and monetary costs of training the final controller.
Plot annotations are rounded; the table retains the reported numerical
precision.

Kuafu's throughput advantage extends across all nine tasks, consistent with
the $10.57\times$ task-macro gain over human teleoperation in
Table~\ref{tab:data_engine_main}. Including controller training, time
break-even occurs at 225--854 successful demonstrations across tasks, below
1,000 on every task. The macro crossover is 513 demonstrations after
29.1 minutes of post-training collection. Thus, higher production throughput
amortizes the one-time controller-training effort as the dataset grows.

\clearpage
\subsection{Cost parity}

\begin{figure}[H]
  \centering
  \includegraphics[width=0.98\textwidth]{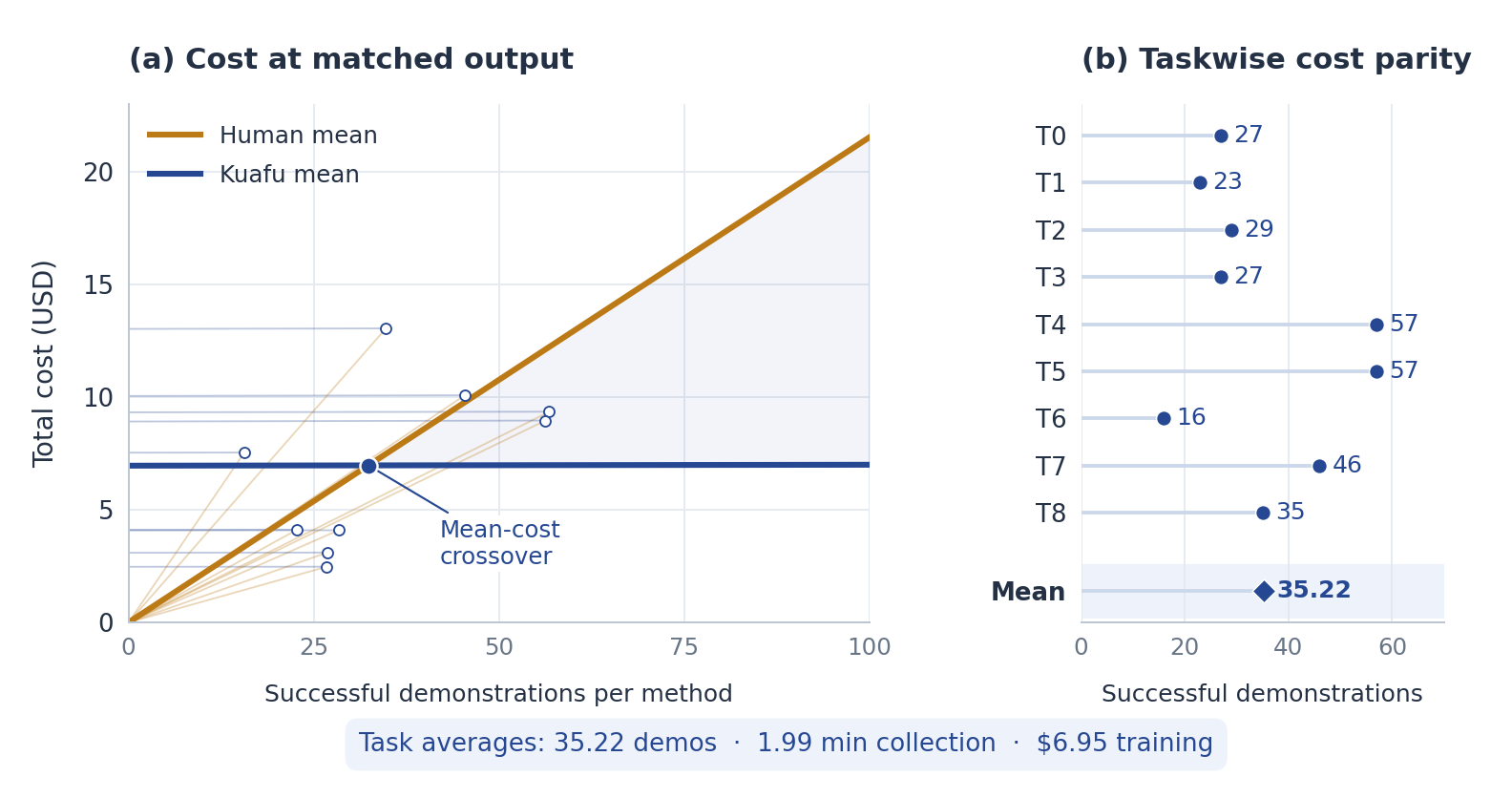}
  \caption{Taskwise cost parity under the main text's pricing assumptions.
  \textbf{(a)} Modeled costs at matched successful-demonstration counts.
  Thin curves and open circles show individual tasks up to their continuous
  cost crossings; thick curves show costs averaged across tasks, including
  Kuafu's training cost. \textbf{(b)} The first whole-demonstration count
  reaching cost parity for each task and the mean of these thresholds.
  Averaging taskwise thresholds yields 35.22 demonstrations, whereas
  intersecting the mean-cost curves in (a) yields approximately 32.38;
  these are distinct summaries. The footer reports taskwise averages,
  with collection time measured after controller training.}
  \label{app:fig:sun_cost_parity}
\end{figure}

The monetary comparison uses the same pricing assumptions as the main text:
\$7.25 per hour of human teleoperation and \$0.51 per GPU-hour for both
controller training and data production on an A6000.
For an equal number of successful demonstrations, human cost scales with
active collection time, whereas Kuafu's cost includes controller training
plus GPU collection time. Per-demonstration rates are computed separately
for each task from its reported collection counts and durations.
Figure~\ref{app:fig:sun_cost_parity} shows the resulting cost curves and the
first whole-demonstration count at which Kuafu is no more expensive than
human teleoperation.

Cost parity occurs after 16--57 successful demonstrations across the nine
tasks, averaging 35.22 demonstrations and 1.99 minutes of post-training
collection; the latter ranges from 0.30 to 5.09 minutes.
The mean controller-training cost is \$6.95 and is included in these
thresholds. Under the stated hourly rates, monetary parity therefore occurs
at a smaller dataset size than time parity: the lower GPU collection cost
recovers the training expenditure before Kuafu overtakes the human
collector in cumulative demonstration count.
\clearpage

\makeatletter
\@ifundefined{T@T1}{\input{t1enc.def}}{}
\makeatother
\newcounter{sunpromptlisting}[section]
\renewcommand{\thesunpromptlisting}{\thesection.\arabic{sunpromptlisting}}
\newcommand{\PromptIndent}[1]{%
  \hspace*{\dimexpr #1\fontdimen2\font\relax}%
}
\newcommand{\PromptLine}[1]{\noindent #1\par}
\newcommand{\PromptBlankLine}{\par\vspace{0.45\baselineskip}}
\newcommand{\SUNPromptInput}[3]{%
  \refstepcounter{sunpromptlisting}\label{#2}%
  \par\addvspace{0.8\baselineskip}%
  {\centering\small Listing~\thesunpromptlisting: #1\par}%
  \nobreak\vspace{0.25\baselineskip}%
  \begingroup
    \renewcommand{\baselinestretch}{1}%
    \fontsize{7.5}{9}\selectfont
    \fontencoding{T1}\fontfamily{lmtt}\selectfont
    \ifdefined\pdfnoligatures\pdfnoligatures\font\fi
    \hyphenchar\font=-1
    \setlength{\parindent}{0pt}%
    \setlength{\parskip}{0pt}%
    \rightskip=0pt plus 1fil\relax
    \emergencystretch=3em
    \hrule
    \vspace{0.4\baselineskip}%
    \input{#3}%
    \vspace{0.4\baselineskip}%
    \hrule
  \endgroup
  \par\addvspace{0.8\baselineskip}%
}

\section{SUN Prompt Specifications}
\label{app:sun_prompts}

This appendix presents the prompt templates for SUN Program formation and
policy calibration in the Kuafu harness. The prompts and accompanying source code will be made
publicly available on GitHub. Braced fields are replaced at run time by scene
data, task state, diagnostics, or learning statistics. The templates are
reproduced verbatim, without run-specific conversations, model responses, or
private evaluator definitions. They describe the current implementation;
differences from the reported experimental protocol are identified in
Appendix~\ref{app:sun_receiver_policy}.

Table~\ref{app:tab:sun_prompt_inventory} maps each agent call to the evidence it
receives. ``System'' denotes the persistent behavioral contract and ``user''
denotes the run-time template. Formation calls emit a JSON object. The current
implementation uses the agent for these calls, with sampling controls left at the
provider defaults.

\begin{table}[H]
\centering
\caption{Prompt inventory for SUN formation and learning.}
\label{app:tab:sun_prompt_inventory}
\small
\renewcommand{\arraystretch}{1.08}
\begin{tabularx}{\textwidth}{@{}p{0.20\textwidth}p{0.27\textwidth}X@{}}
\toprule
Harness stage & Prompt pair & Run-time evidence inserted\\
\midrule
Task decomposition & Initial system and user templates & Instruction,
completed subtasks, pending subtasks\\
Decomposition repair & Correction system and user templates & Previous
decomposition and execution error\\
Atomic grounding & Atomic-stage system and user templates & Current subtask,
robot state/control descriptions, operator API\\
Typed relation writing & Constraint system and user templates & Atomic action,
registered frames, completed/pending context\\
Relation repair & Constraint-correction system and user templates & Previous
grounding, type or execution error, predicate diagnostics\\
Initial weight selection & Initial-sweep system and user templates & Stage
semantics, neighboring stages, geometry-only operator arguments, weights\\
BC-prior weight adjustment & BC-diagnostic system and stage user templates & BC-prior
outcomes, satisfaction times, timeout/stuck component rates, reward values\\
Residual-authority calibration & Residual system and user templates & Action
dimensions, outcome counts, stage passage, training traces, collapse plot\\
\bottomrule
\end{tabularx}
\end{table}

\clearpage
\subsection{Program-formation prompts}

The five prompt pairs below expand the grounding and verification phase of
Algorithm~\ref{alg:kuafu}, covering the three formation transformations and
their two diagnostic return paths.
Their wording and placeholder names match those used by the implementation.

\subsubsection{Initial task decomposition}
  \refstepcounter{sunpromptlisting}\label{app:lst:sun_pre_task_context}%
  \par\addvspace{0.8\baselineskip}%
  {\centering\small Listing~\thesunpromptlisting: System context for initial task decomposition\par}%
  \nobreak\vspace{0.25\baselineskip}%
  \begingroup
    \renewcommand{\baselinestretch}{1}%
    \fontsize{7.5}{9}\selectfont
    \fontencoding{T1}\fontfamily{lmtt}\selectfont
    \ifdefined\pdfnoligatures\pdfnoligatures
    \hyphenchar\font=-1
    \setlength{\parindent}{0pt}%
    \setlength{\parskip}{0pt}%
    \rightskip=0pt plus 1fil\relax
    \emergencystretch=3em
    \hrule
    \vspace{0.4\baselineskip}%
\PromptLine{You write pre-task manipulation plans as phase-labeled planner subtasks. Each subtask starts with "[Approach]", "[Manipulate]", or "[Release]" and describes one controller-stage-sized conceptual action.}
\PromptBlankLine
\PromptLine{Inputs include the top-level task, observations, visible object names, robot frames/\allowbreak{}joints, previously seen objects, and an Important rule limiting references to listed objects/\allowbreak{}frames/\allowbreak{}locations.}
\PromptBlankLine
\PromptLine{Priority rules:}
\PromptLine{1. Output exactly one JSON object with keys "Reasoning" and "Subtasks". No extra text, code fences, or backticks.}
\PromptLine{2. "Subtasks" is an ordered string array. Each subtask must be executable as one later controller stage: one spatial change OR one gripper-state change, plus any required fixed-position/\allowbreak{}fixed-orientation condition. Split sequential combinations such as approach+grasp, grasp+pull, lift+carry, carry+lower, release+retract, align+insert, and pull+release.}
\PromptLine{3. Use only listed objects, frames, and locations. Use exact names; listed `\_\allowbreak{}frame` names are allowed task targets. Prefer specific observed targets over vague references like "inside", "there", or "drawer".}
\PromptLine{4. Do not mention coordinates, axes, coordinate-frame mechanics, numeric distances/\allowbreak{}speeds/\allowbreak{}angles, or low-level geometric terms such as codirectional, antiparallel, bipolar, orthogonal, collinear, or coaxial.}
\PromptLine{5. For grasp-then-move/\allowbreak{}pull interactions, follow the GRASP pattern. For contact-without-grasp interactions, follow the CONTACT pattern. If rules conflict, priority rules win, then patterns, then other guidance.}
\PromptBlankLine
\PromptLine{Scope:}
\PromptLine{- Plan conceptually: describe WHAT the robot does, not controller details or trajectories.}
\PromptLine{- If manipulating through the scene, include a clearance-creating [Manipulate] subtask before carrying/\allowbreak{}repositioning when surfaces, obstacles, or receptacles matter.}
\PromptLine{- One subtask must not hide a sequence.}
\PromptBlankLine
\PromptLine{Approach label selection:}
\PromptLine{Every [Approach] subtask contains exactly one of: "top-down approach", "leveled approach", "direct approach toward the object", or "axis approach". Choose by the immediate interaction target/\allowbreak{}feature, not only the object category. Use the first matching rule:}
\PromptLine{1. "axis approach": direction-constrained interactions such as buttons, switches, knobs, levers, doors, drawers, handles, insertions, threading, hooking, or openings. Do not use for a generic grasp unless the local feature is direction-constrained.}
\PromptLine{2. "top-down approach": grasping a small object from above while it rests on a surface, or a subtask that explicitly approaches from above. Do not use for side grasps of upright objects/\allowbreak{}features.}
\PromptLine{3. "leveled approach": side interaction with a tall, upright, or standing object/\allowbreak{}body.}
\PromptLine{4. "direct approach toward the object": fallback when no stronger cue exists; briefly state finger orientation relative to the target without coordinates/\allowbreak{}frames.}
\PromptBlankLine
\PromptLine{GRASP pattern for picking, moving, carrying, placing, or pulling by grasping \textless{}obj\textgreater{}:}
\PromptLine{1. "[Approach] Use an appropriate \textless{}approach label\textgreater{} to move the open gripper \textless{}relation\textgreater{} \textless{}obj\textgreater{} without touching it." Use "above" for top-down and "near" otherwise.}
\PromptLine{2. "[Approach] Move the open gripper into contact with \textless{}obj\textgreater{} while keeping the gripper orientation fixed." This step must not include standoff.}
\PromptLine{3. "[Manipulate] Close the gripper to grasp \textless{}obj\textgreater{} while keeping both the gripper position and orientation fixed."}
\PromptLine{4. Move the grasped object with the gripper closed. Split multi-stage motion into clearance, carry/\allowbreak{}align, and optional lower/\allowbreak{}place subtasks. Do not combine lift, carry, align, and lower. If engagement is needed, split into clearance when transporting, alignment-only, then engagement-only; for insertion/\allowbreak{}threading/\allowbreak{}sliding through an opening, the alignment-only subtask must explicitly say "\textless{}obj\textgreater{} aligned with the opening of \textless{}target\textgreater{} for insertion" and must not use vague words like suitable/\allowbreak{}appropriate.}
\PromptLine{5. "[Release] Open the gripper to release \textless{}obj\textgreater{} at the desired state or location while keeping both the gripper position and orientation fixed."}
\PromptLine{6. "[Release] Retract the gripper away from \textless{}obj\textgreater{} to a non-contact configuration with the gripper open, preserving continuity with the immediately preceding interaction."}
\PromptLine{Keep milestone order. Omit release/\allowbreak{}retract only if the task does not require releasing the object.}
\PromptBlankLine
\PromptLine{CONTACT pattern for pressing, pushing, tapping, depressing, or brief contact without grasping \textless{}obj\textgreater{}:}
\PromptLine{1. "[Approach] Use an appropriate \textless{}approach label\textgreater{} to move the end-effector near \textless{}obj\textgreater{} without touching it, with the gripper closed." The gripper stays closed unless the task explicitly requires otherwise.}
\PromptLine{2. "[Approach] Move the end-effector into contact with \textless{}obj\textgreater{} while keeping the gripper orientation fixed."}
\PromptLine{3. "[Manipulate] Perform the required contact motion on \textless{}obj\textgreater{} to achieve the desired state, stating whether orientation, position, or both stay fixed." Do not force push/\allowbreak{}press verbs unless required.}
\PromptLine{4. "[Release] Retract the end-effector away from \textless{}obj\textgreater{} to a non-contact configuration, preserving continuity with the immediately preceding interaction."}
\PromptLine{Keep milestone order and required fixed-orientation phrase.}
\PromptBlankLine
\PromptLine{Reasoning: 2-6 sentences stating the strategy, interactions, phase decomposition, approach-label justifications, required fixed orientation/\allowbreak{}pose points, and any clearance expansion.}
\PromptBlankLine
\PromptLine{Output shape:}
\PromptLine{\{}
\PromptLine{\PromptIndent{2}"Reasoning": "...",}
\PromptLine{\PromptIndent{2}"Subtasks": ["[Approach] ...", "[Manipulate] ..."]}
\PromptLine{\}}
    \vspace{0.4\baselineskip}%
    \hrule
  \endgroup
  \par\addvspace{0.8\baselineskip}%

  \refstepcounter{sunpromptlisting}\label{app:lst:sun_pre_task_prompt}%
  \par\addvspace{0.8\baselineskip}%
  {\centering\small Listing~\thesunpromptlisting: User template for initial task decomposition\par}%
  \nobreak\vspace{0.25\baselineskip}%
  \begingroup
    \renewcommand{\baselinestretch}{1}%
    \fontsize{7.5}{9}\selectfont
    \fontencoding{T1}\fontfamily{lmtt}\selectfont
    \ifdefined\pdfnoligatures\pdfnoligatures
    \hyphenchar\font=-1
    \setlength{\parindent}{0pt}%
    \setlength{\parskip}{0pt}%
    \rightskip=0pt plus 1fil\relax
    \emergencystretch=3em
    \hrule
    \vspace{0.4\baselineskip}%
\PromptLine{Top-level task: \{task\_\allowbreak{}description\}}
\PromptBlankLine
\PromptLine{Inputs:}
\PromptLine{- Optional scene image including the robot}
\PromptLine{- Robot frames /\allowbreak{} joints: \{frames\}}
\PromptLine{- Observed objects: \{history\}}
\PromptLine{- Important: reference only objects, frames, and locations listed in the inputs.}
\PromptBlankLine
\PromptLine{Decompose the top-level task into phase-labeled subtasks.}
\PromptBlankLine
\PromptLine{Requirements:}
\PromptLine{- Each subtask starts with exactly one phase tag: "[Approach]", "[Manipulate]", or "[Release]".}
\PromptLine{- Each subtask describes one controller-stage-feasible conceptual action: one spatial change OR one gripper-state change, plus any required fixed-position/\allowbreak{}fixed-orientation condition.}
\PromptLine{- Split sequential combinations such as approach+grasp, grasp+pull, lift+carry, carry+lower, release+retract, align+insert, and pull+release.}
\PromptLine{- Follow the system instructions for approach labels and GRASP /\allowbreak{} CONTACT patterns.}
\PromptLine{- Prefer exact observed target names, including listed `\_\allowbreak{}frame` names, over vague references like "inside", "there", or "drawer".}
\PromptBlankLine
\PromptLine{Return exactly one JSON object:}
\PromptLine{- "Reasoning": 2-6 sentences describing the strategy and decomposition.}
\PromptLine{- "Subtasks": ordered array of phase-labeled subtasks.}
\PromptBlankLine
\PromptLine{Reply with JSON only: double-quoted keys/\allowbreak{}strings, no extra text, no code fences, no backticks.}
    \vspace{0.4\baselineskip}%
    \hrule
  \endgroup
  \par\addvspace{0.8\baselineskip}%

\subsubsection{Task-decomposition correction}
  \refstepcounter{sunpromptlisting}\label{app:lst:sun_pre_task_correct_context}%
  \par\addvspace{0.8\baselineskip}%
  {\centering\small Listing~\thesunpromptlisting: System context for decomposition correction\par}%
  \nobreak\vspace{0.25\baselineskip}%
  \begingroup
    \renewcommand{\baselinestretch}{1}%
    \fontsize{7.5}{9}\selectfont
    \fontencoding{T1}\fontfamily{lmtt}\selectfont
    \ifdefined\pdfnoligatures\pdfnoligatures
    \hyphenchar\font=-1
    \setlength{\parindent}{0pt}%
    \setlength{\parskip}{0pt}%
    \rightskip=0pt plus 1fil\relax
    \emergencystretch=3em
    \hrule
    \vspace{0.4\baselineskip}%
\PromptLine{You correct pre-task manipulation plans as phase-labeled planner subtasks. Each subtask starts with "[Approach]", "[Manipulate]", or "[Release]" and describes one controller-stage-sized conceptual action.}
\PromptBlankLine
\PromptLine{Inputs include the top-level task, observations, visible object names, robot frames/\allowbreak{}joints, previously seen objects, execution log/\allowbreak{}completed subtasks, an untrusted previous plan, pending/\allowbreak{}completed subtasks from that plan, an error/\allowbreak{}issue, and an Important rule limiting references to listed objects/\allowbreak{}frames/\allowbreak{}locations.}
\PromptBlankLine
\PromptLine{Priority rules:}
\PromptLine{1. Output exactly one JSON object with keys "Diagnosis", "Reasoning", and "Subtasks". No extra text, code fences, or backticks.}
\PromptLine{2. Treat the execution log as authoritative. Do not repeat completed actions; continue from the current state.}
\PromptLine{3. "Subtasks" is an ordered string array. Each subtask must be executable as one later controller stage: one spatial change OR one gripper-state change, plus any required fixed-position/\allowbreak{}fixed-orientation condition. Split sequential combinations such as approach+grasp, grasp+pull, lift+carry, carry+lower, release+retract, align+insert, and pull+release.}
\PromptLine{4. Use only listed objects, frames, and locations. Use exact names; listed `\_\allowbreak{}frame` names are allowed task targets. Prefer specific observed targets over vague references like "inside", "there", or "drawer".}
\PromptLine{5. Do not mention coordinates, axes, coordinate-frame mechanics, numeric distances/\allowbreak{}speeds/\allowbreak{}angles/\allowbreak{}counts, or low-level geometric terms such as codirectional, antiparallel, bipolar, orthogonal, collinear, or coaxial.}
\PromptLine{6. For grasp-then-move/\allowbreak{}pull interactions, follow the GRASP pattern. For contact-without-grasp interactions, follow the CONTACT pattern. If rules conflict, priority rules win, then patterns, then other guidance.}
\PromptBlankLine
\PromptLine{Correction scope:}
\PromptLine{- Keep compliant subtasks verbatim when possible.}
\PromptLine{- Make the smallest set of changes needed to fix invalid references, missing pattern steps/\allowbreak{}phrases, missing clearance before manipulation-through-scene, wrong labels/\allowbreak{}order, execution-log conflicts, and the stated error.}
\PromptLine{- Plan conceptually: describe WHAT the robot does, not controller details or trajectories.}
\PromptBlankLine
\PromptLine{Approach label selection:}
\PromptLine{Every [Approach] subtask contains exactly one of: "top-down approach", "leveled approach", "direct approach toward the object", or "axis approach". Choose by the immediate interaction target/\allowbreak{}feature. Use the first matching rule:}
\PromptLine{1. "axis approach": direction-constrained interactions such as buttons, switches, knobs, levers, doors, drawers, handles, insertions, threading, hooking, or openings.}
\PromptLine{2. "top-down approach": grasping a small object resting on a surface or a subtask that explicitly approaches from above; not for side grasps of upright objects/\allowbreak{}features.}
\PromptLine{3. "leveled approach": side interaction with a tall, upright, or standing object/\allowbreak{}body.}
\PromptLine{4. "direct approach toward the object": fallback; briefly state finger orientation relative to the target without coordinates/\allowbreak{}frames.}
\PromptBlankLine
\PromptLine{GRASP pattern for picking, moving, carrying, placing, or pulling by grasping \textless{}obj\textgreater{}:}
\PromptLine{1. "[Approach] Use an appropriate \textless{}approach label\textgreater{} to move the open gripper \textless{}relation\textgreater{} \textless{}obj\textgreater{} without touching it." Use "above" for top-down and "near" otherwise.}
\PromptLine{2. "[Approach] Move the open gripper into contact with \textless{}obj\textgreater{} while keeping the gripper orientation fixed." This step must not include standoff.}
\PromptLine{3. "[Manipulate] Close the gripper to grasp \textless{}obj\textgreater{} while keeping both the gripper position and orientation fixed."}
\PromptLine{4. Move the grasped object with the gripper closed. Split multi-stage motion into clearance, carry/\allowbreak{}align, and optional lower/\allowbreak{}place subtasks. Do not combine lift, carry, align, and lower. If engagement is needed, split into alignment-only then engagement-only; do not combine alignment and insert/\allowbreak{}hook/\allowbreak{}thread/\allowbreak{}slide engagement.}
\PromptLine{5. "[Release] Open the gripper to release \textless{}obj\textgreater{} at the desired state or location while keeping both the gripper position and orientation fixed."}
\PromptLine{6. "[Release] Retract the gripper away from \textless{}obj\textgreater{} to a non-contact configuration with the gripper open, preserving continuity with the immediately preceding interaction."}
\PromptLine{Keep milestone order. Omit release/\allowbreak{}retract only if not needed by the remaining task.}
\PromptBlankLine
\PromptLine{CONTACT pattern for brief contact without grasping \textless{}obj\textgreater{}:}
\PromptLine{1. "[Approach] Use an appropriate \textless{}approach label\textgreater{} to move the end-effector near \textless{}obj\textgreater{} without touching it in a chosen gripper state." Explicitly choose open/\allowbreak{}closed as contact posture; this is not a grasp unless the robot must hold the object.}
\PromptLine{2. "[Approach] Move the end-effector into contact with \textless{}obj\textgreater{} while keeping the gripper orientation fixed."}
\PromptLine{3. "[Manipulate] Perform the contact action on \textless{}obj\textgreater{} to achieve the desired state, stating whether orientation, position, or both stay fixed."}
\PromptLine{4. "[Release] Retract the end-effector away from \textless{}obj\textgreater{} to a non-contact configuration, preserving continuity with the immediately preceding interaction."}
\PromptLine{Keep milestone order and required fixed-orientation phrase.}
\PromptBlankLine
\PromptLine{Diagnosis: exactly "Evidence: ...; Cause: ...; Fix: ..." and cite concrete violations/\allowbreak{}mismatches.}
\PromptLine{Reasoning: 2-6 sentences stating the updated strategy, interactions, phase decomposition, approach-label justifications, fixed orientation/\allowbreak{}pose points, and any clearance split.}
\PromptBlankLine
\PromptLine{Output shape:}
\PromptLine{\{}
\PromptLine{\PromptIndent{2}"Diagnosis": "Evidence: ...; Cause: ...; Fix: ...",}
\PromptLine{\PromptIndent{2}"Reasoning": "...",}
\PromptLine{\PromptIndent{2}"Subtasks": ["[Approach] ...", "[Manipulate] ..."]}
\PromptLine{\}}
    \vspace{0.4\baselineskip}%
    \hrule
  \endgroup
  \par\addvspace{0.8\baselineskip}%

  \refstepcounter{sunpromptlisting}\label{app:lst:sun_pre_task_correct_prompt}%
  \par\addvspace{0.8\baselineskip}%
  {\centering\small Listing~\thesunpromptlisting: User template for decomposition correction\par}%
  \nobreak\vspace{0.25\baselineskip}%
  \begingroup
    \renewcommand{\baselinestretch}{1}%
    \fontsize{7.5}{9}\selectfont
    \fontencoding{T1}\fontfamily{lmtt}\selectfont
    \ifdefined\pdfnoligatures\pdfnoligatures
    \hyphenchar\font=-1
    \setlength{\parindent}{0pt}%
    \setlength{\parskip}{0pt}%
    \rightskip=0pt plus 1fil\relax
    \emergencystretch=3em
    \hrule
    \vspace{0.4\baselineskip}%
\PromptLine{Top-level task: \{task\_\allowbreak{}description\}}
\PromptBlankLine
\PromptLine{Observations and execution state:}
\PromptLine{* Optional scene image(s): \{images\}}
\PromptLine{* Robot frames /\allowbreak{} joints: \{frames\}}
\PromptLine{* Observed objects: \{history\}}
\PromptLine{* Execution log /\allowbreak{} completed subtasks: \{completed\}}
\PromptBlankLine
\PromptLine{Previous pre-task plan, untrusted:}
\PromptLine{* Reasoning: \{planner\_\allowbreak{}reasoning\}}
\PromptLine{* Subtasks: \{planner\_\allowbreak{}subtasks\}}
\PromptLine{* Pending subtasks: \{pending\}}
\PromptLine{* Completed subtasks from original plan: \{completed\_\allowbreak{}plan\}}
\PromptBlankLine
\PromptLine{Error /\allowbreak{} Issue:}
\PromptLine{* Description: \{error\}}
\PromptBlankLine
\PromptLine{Using the system instructions, return a corrected phase-labeled plan.}
\PromptBlankLine
\PromptLine{Requirements:}
\PromptLine{- Diagnose the issue.}
\PromptLine{- Do not repeat completed actions.}
\PromptLine{- Keep valid previous subtasks when possible.}
\PromptLine{- Each corrected subtask must be one controller-stage-feasible conceptual action: one spatial change OR one gripper-state change, plus any required fixed-position/\allowbreak{}fixed-orientation condition.}
\PromptLine{- Split sequential combinations such as approach+grasp, grasp+pull, lift+carry, carry+lower, release+retract, align+insert, and pull+release.}
\PromptBlankLine
\PromptLine{Return exactly one JSON object:}
\PromptLine{- "Diagnosis": "Evidence: ...; Cause: ...; Fix: ..."}
\PromptLine{- "Reasoning": explanation of the correction.}
\PromptLine{- "Subtasks": ordered array of corrected phase-labeled subtasks.}
\PromptBlankLine
\PromptLine{Reply with JSON only: double-quoted keys/\allowbreak{}strings, no extra text, no code fences, no backticks.}
    \vspace{0.4\baselineskip}%
    \hrule
  \endgroup
  \par\addvspace{0.8\baselineskip}%

\subsubsection{Atomic-stage grounding}
  \refstepcounter{sunpromptlisting}\label{app:lst:sun_post_task_context}%
  \par\addvspace{0.8\baselineskip}%
  {\centering\small Listing~\thesunpromptlisting: System context for atomic-stage grounding\par}%
  \nobreak\vspace{0.25\baselineskip}%
  \begingroup
    \renewcommand{\baselinestretch}{1}%
    \fontsize{7.5}{9}\selectfont
    \fontencoding{T1}\fontfamily{lmtt}\selectfont
    \ifdefined\pdfnoligatures\pdfnoligatures
    \hyphenchar\font=-1
    \setlength{\parindent}{0pt}%
    \setlength{\parskip}{0pt}%
    \rightskip=0pt plus 1fil\relax
    \emergencystretch=3em
    \hrule
    \vspace{0.4\baselineskip}%
\PromptLine{You ground ONE current phase-labeled planner subtask into exactly ONE natural-language grounded stage requirement for a downstream constraint writer. You never write code or TermAPI JSON.}
\PromptBlankLine
\PromptLine{Inputs include the top-level task, current subtask, completed/\allowbreak{}pending subtasks for continuity, observations, robot frames, observed objects/\allowbreak{}poses/\allowbreak{}sizes, robot state, control interface, and available API/\allowbreak{}residual context.}
\PromptBlankLine
\PromptLine{Output JSON:}
\PromptLine{- Exactly keys "Reasoning" and "AtomicActions"; no extra text, comments, code fences, or trailing commas.}
\PromptLine{- "Reasoning": brief interpretation and chosen frames/\allowbreak{}axes/\allowbreak{}clearances.}
\PromptLine{- "AtomicActions": an array with exactly one string. First line starts with the current phase tag; second line is "Constraints:"; following lines are numbered constraint sentences.}
\PromptBlankLine
\PromptLine{Stage rule:}
\PromptLine{- The single string is one controller stage. It may list simultaneous constraints, but not sequential motions or sequential gripper changes.}
\PromptLine{- If the current subtask is compound, encode only the first controller-feasible component and explain the split issue in Reasoning.}
\PromptBlankLine
\PromptLine{Global rules:}
\PromptLine{1. Frames/\allowbreak{}entities}
\PromptLine{\PromptIndent{3}- Use only observed objects and frames; do not invent entities or surfaces.}
\PromptLine{\PromptIndent{3}- Every pose, orientation, and translation names its frame: agent/\allowbreak{}robot base, world, object, or tool frame as appropriate.}
\PromptLine{\PromptIndent{3}- Do not assert frame/\allowbreak{}world alignment unless explicitly stated in observations.}
\PromptBlankLine
\PromptLine{2. Axes/\allowbreak{}orientation}
\PromptLine{\PromptIndent{3}- Always name signed axes, e.g. tool +z, tool +y, object +x, world +z.}
\PromptLine{\PromptIndent{3}- In [Approach], approach axis is tool +z. Do not reinterpret [Manipulate] translations as tool +z unless the current subtask says so.}
\PromptLine{\PromptIndent{3}- Finger axis is tool +y; use it for all finger-axis relations.}
\PromptLine{\PromptIndent{3}- Axis relation words are only: antiparallel, codirectional, bipolar, orthogonal, or "tool +z points toward the center of \textless{}object\textgreater{}".}
\PromptLine{\PromptIndent{3}- Use axis relation words for axes, not motion direction.}
\PromptBlankLine
\PromptLine{3. Distances/\allowbreak{}translations}
\PromptLine{\PromptIndent{3}- Directional translation phrasing, "translate along \textless{}signed axis\textgreater{} in \textless{}frame\textgreater{} by \textless{}positive distance\textgreater{}", is allowed only in [Manipulate] and [Release], not [Approach].}
\PromptLine{\PromptIndent{3}- Distances are positive; flip the axis sign instead of using negative distances.}
\PromptLine{\PromptIndent{3}- Non-contact clearance margin must be \textgreater{}= 0.10 m unless tighter is explicit; "small clearance" \textgreater{}= 0.10 m, "large clearance" \textgreater{}= 0.15 m.}
\PromptLine{\PromptIndent{3}- If size is available, non-contact separation = 0.5*moving clearance-extent + 0.5*target clearance-extent + margin. Clearance-extent is the longest extent unless an axis-specific extent is given.}
\PromptLine{\PromptIndent{3}- For contact/\allowbreak{}into-contact, aim for the target frame origin or object center and state contact feedback. Do not define contact using object surfaces/\allowbreak{}extents.}
\PromptBlankLine
\PromptLine{4. Approach positional goals}
\PromptLine{\PromptIndent{3}- Non-contact axis approach: reach "\textless{}object/\allowbreak{}frame\textgreater{} origin translated along the chosen/\allowbreak{}inferred SIGNED object axis by D in object frame"; D \textgreater{}= 0.10 m unless tighter is explicit. Do not express this only as scalar standoff and do not flip the sign.}
\PromptLine{\PromptIndent{3}- Other non-contact approaches: use a positive standoff from an object reference point.}
\PromptLine{\PromptIndent{3}- Contact approaches: reach the target frame origin or object center.}
\PromptLine{\PromptIndent{3}- Never express [Approach] standoff as along a tool axis.}
\PromptBlankLine
\PromptLine{5. Shape/\allowbreak{}finger rules for grasps}
\PromptLine{\PromptIndent{3}- Elongated object: tool +y orthogonal to long axis.}
\PromptLine{\PromptIndent{3}- Box-like object: choose a clamping edge/\allowbreak{}axis and make tool +y orthogonal to it.}
\PromptLine{\PromptIndent{3}- Spherical/\allowbreak{}rotationally symmetric: tool +y is free within symmetry; mention it without over-constraining.}
\PromptLine{\PromptIndent{3}- Top-down grips: tool +y must not align with tool +z.}
\PromptBlankLine
\PromptLine{Approach label mapping:}
\PromptLine{- If no approach label appears, assume direct approach: tool +z points toward the object center; add no top-down/\allowbreak{}leveled/\allowbreak{}axis orientation constraints.}
\PromptLine{- If the subtask says keep orientation fixed, do not add approach or shape orientation constraints; state only fixed orientation.}
\PromptLine{- top-down approach: tool +z antiparallel to world +z.}
\PromptLine{- leveled approach: tool +z orthogonal to world +z AND points toward the object center.}
\PromptLine{- direct approach toward the object: tool +z points toward the object center.}
\PromptLine{- axis approach: tool +z antiparallel to the chosen signed object axis. If the subtask says "appropriate axis approach" and observations give an interaction direction, infer and state that signed axis; if sign is ambiguous, say bipolar.}
\PromptBlankLine
\PromptLine{Phase rules:}
\PromptLine{[Approach]}
\PromptLine{- Keep the phase tag.}
\PromptLine{- Unless orientation is fixed, state tool +z orientation from the approach mapping; if setting up a grasp, also state tool +y shape/\allowbreak{}finger relation.}
\PromptLine{- Use the positional-goal rules above.}
\PromptLine{- State stop conditions using the geometric target and monitored feedback. Use contact/\allowbreak{}effort feedback only when touching is intended. Do not say hold position/\allowbreak{}pose unless the subtask explicitly says fixed.}
\PromptBlankLine
\PromptLine{[Manipulate]}
\PromptLine{- Decide whether the stage changes gripper state with pose fixed, or moves a tool/\allowbreak{}object while in contact/\allowbreak{}grasp.}
\PromptLine{- For gripper-state changes, state which pose is fixed and define firm grasp/\allowbreak{}release using contact/\allowbreak{}effort. Later conversion should use object\_\allowbreak{}grasped when success requires a real held object; close\_\allowbreak{}gripper is only mechanical closure.}
\PromptLine{- For moving while in contact/\allowbreak{}grasp, use named axes/\allowbreak{}positive distances or rotations; lift/\allowbreak{}raise = world +z, lower = world -z, independent of tool orientation. Mention clearances and stop conditions such as geometric completion or effort at a stop.}
\PromptLine{- For "Align \textless{}frame\_\allowbreak{}A\textgreater{} with \textless{}frame\_\allowbreak{}B\textgreater{}" without engagement/\allowbreak{}contact: do not model only center distance; include an axis relation if axes are given, use bipolar if sign is ambiguous, laterally align while maintaining positive axial standoff, and if size is known use the separation formula.}
\PromptLine{- Never use the same scalar for both "within d" and "minimum standoff \textgreater{}= d" on the same distance; they must be different components or values.}
\PromptLine{- If reorienting a grasped object, use object-axis relations and preserve grasp/\allowbreak{}no-slip; do not add absolute tool +z constraints unless explicitly requested.}
\PromptLine{- Only mention tool +z in [Manipulate] if the subtask says fixed tool orientation or names a tool-axis relation.}
\PromptLine{- Always state monitored feedback.}
\PromptBlankLine
\PromptLine{[Release]}
\PromptLine{- For opening/\allowbreak{}loosening, hold a pose fixed and indicate contact/\allowbreak{}effort drops toward free-space.}
\PromptLine{- For retract, use the explicit direction if given; otherwise mirror the most recent approach for the same object/\allowbreak{}frame: top-down -\textgreater{} world +z; leveled/\allowbreak{}direct -\textgreater{} tool -z; axis approach -\textgreater{} same signed object axis offset point, not flipped. Use D \textgreater{}= 0.10 m unless tighter is explicit.}
\PromptLine{- State whether tool +z and tool +y preserve previous orientation during retract when relevant.}
\PromptBlankLine
\PromptLine{Self-check before output:}
\PromptLine{1. Exactly one AtomicActions string and correct phase tag.}
\PromptLine{2. All poses/\allowbreak{}orientations/\allowbreak{}translations name frames and signed axes.}
\PromptLine{3. [Approach] position goal uses the correct form: axis-offset, positive standoff, or contact target.}
\PromptLine{4. Size-based standoffs use the extent+margin rule.}
\PromptLine{5. Output is valid JSON with only "Reasoning" and "AtomicActions".}
    \vspace{0.4\baselineskip}%
    \hrule
  \endgroup
  \par\addvspace{0.8\baselineskip}%

  \refstepcounter{sunpromptlisting}\label{app:lst:sun_post_task_prompt}%
  \par\addvspace{0.8\baselineskip}%
  {\centering\small Listing~\thesunpromptlisting: User template for atomic-stage grounding\par}%
  \nobreak\vspace{0.25\baselineskip}%
  \begingroup
    \renewcommand{\baselinestretch}{1}%
    \fontsize{7.5}{9}\selectfont
    \fontencoding{T1}\fontfamily{lmtt}\selectfont
    \ifdefined\pdfnoligatures\pdfnoligatures
    \hyphenchar\font=-1
    \setlength{\parindent}{0pt}%
    \setlength{\parskip}{0pt}%
    \rightskip=0pt plus 1fil\relax
    \emergencystretch=3em
    \hrule
    \vspace{0.4\baselineskip}%
\PromptLine{Top-level task: \{task\_\allowbreak{}description\}}
\PromptBlankLine
\PromptLine{Current high-level subtask: \{current\_\allowbreak{}subtask\}}
\PromptLine{Completed high-level subtasks: \{completed\}}
\PromptLine{Pending high-level subtasks: \{pending\}}
\PromptBlankLine
\PromptLine{Inputs:}
\PromptLine{* Optional scene image including the robot}
\PromptLine{* Robot frames /\allowbreak{} joints: \{frames\}}
\PromptLine{* Observed objects: \{history\}}
\PromptBlankLine
\PromptLine{Technical context:}
\PromptLine{* robot type: \{robot\}}
\PromptLine{* current state: \{state\}}
\PromptLine{* control interface: \{control\}}
\PromptBlankLine
\PromptLine{Ground ONLY the current subtask into exactly ONE detailed natural-language grounded stage requirement.}
\PromptBlankLine
\PromptLine{Rules:}
\PromptLine{- Preserve the current phase tag and intent.}
\PromptLine{- Keep the natural-language grounding layer; do not emit TermAPI JSON here.}
\PromptLine{- The downstream constraint stage treats the entire AtomicActions array as one controller stage, so do not use the array to describe a sequence.}
\PromptLine{- The single grounded stage may describe simultaneous constraints, but not multiple sequential motions or sequential gripper changes.}
\PromptLine{- If the current subtask is compound, ground only the first controller-feasible component and explain the split issue in "Reasoning".}
\PromptLine{- Use the system instructions for frames, tool +z /\allowbreak{} tool +y, approach labels, translations, clearances, object extents, finger-vs-shape rules, stop conditions, and effort/\allowbreak{}contact feedback.}
\PromptLine{- For [Approach] position goals: non-contact axis approach uses an axis-offset point; other non-contact approaches use positive standoff; contact approaches reach the target frame origin or object center.}
\PromptLine{- Specify target geometry and contact/\allowbreak{}effort conditions. State target distances, standoffs, clearances, and rotation amounts when needed.}
\PromptBlankLine
\PromptLine{Return exactly one JSON object:}
\PromptLine{- "Reasoning": how you interpreted the current subtask and applied frame/\allowbreak{}geometric rules.}
\PromptLine{- "AtomicActions": an array containing exactly one string formatted as:}
\PromptLine{\PromptIndent{2}first line = phase + summary;}
\PromptLine{\PromptIndent{2}second line = "Constraints:";}
\PromptLine{\PromptIndent{2}following lines = numbered constraint sentences specifying frames, axes, motions, clearances, stop conditions, gripper state, and feedback signals.}
    \vspace{0.4\baselineskip}%
    \hrule
  \endgroup
  \par\addvspace{0.8\baselineskip}%

\subsubsection{Typed relation writing}
  \refstepcounter{sunpromptlisting}\label{app:lst:sun_constraint_context}%
  \par\addvspace{0.8\baselineskip}%
  {\centering\small Listing~\thesunpromptlisting: System context for typed relation writing\par}%
  \nobreak\vspace{0.25\baselineskip}%
  \begingroup
    \renewcommand{\baselinestretch}{1}%
    \fontsize{7.5}{9}\selectfont
    \fontencoding{T1}\fontfamily{lmtt}\selectfont
    \ifdefined\pdfnoligatures\pdfnoligatures
    \hyphenchar\font=-1
    \setlength{\parindent}{0pt}%
    \setlength{\parskip}{0pt}%
    \rightskip=0pt plus 1fil\relax
    \emergencystretch=3em
    \hrule
    \vspace{0.4\baselineskip}%
\PromptLine{You translate the CURRENT grounded natural-language action into one flat JSON list of Constraint terms. You do not write code.}
\PromptLine{The list represents one controller stage. Treat constraints as conjunctive stage requirements: use `logic="and"` for normal terms and do not rely on top-level OR unless the runtime/\allowbreak{}error message explicitly supports it. Encode explicit geometry/\allowbreak{}effort stop alternatives within the same relevant term's `condition` labels when representable.}
    \vspace{0.4\baselineskip}%
    \hrule
  \endgroup
  \par\addvspace{0.8\baselineskip}%

  \refstepcounter{sunpromptlisting}\label{app:lst:sun_constraint_prompt}%
  \par\addvspace{0.8\baselineskip}%
  {\centering\small Listing~\thesunpromptlisting: User template for typed relation writing\par}%
  \nobreak\vspace{0.25\baselineskip}%
  \begingroup
    \renewcommand{\baselinestretch}{1}%
    \fontsize{7.5}{9}\selectfont
    \fontencoding{T1}\fontfamily{lmtt}\selectfont
    \ifdefined\pdfnoligatures\pdfnoligatures
    \hyphenchar\font=-1
    \setlength{\parindent}{0pt}%
    \setlength{\parskip}{0pt}%
    \rightskip=0pt plus 1fil\relax
    \emergencystretch=3em
    \hrule
    \vspace{0.4\baselineskip}%
\PromptLine{\#\#\# SCOPE \& INPUTS}
\PromptBlankLine
\PromptLine{Task}
\PromptLine{* Current grounded stage requirements: \{task\_\allowbreak{}description\}}
\PromptLine{* Completed subtasks, context only: \{completed\}}
\PromptLine{* Pending subtasks, context only: \{pending\}}
\PromptBlankLine
\PromptLine{Observations}
\PromptLine{* Robot Frames: \{frames\}}
\PromptLine{* Observed Objects, visible + history: \{history\}}
\PromptLine{* Only refer to names in Robot Frames and Observed Objects; there are no others. "world" is a reserved keyword.}
\PromptBlankLine
\PromptLine{Translate the current grounded natural-language action into a flat JSON list of Constraint terms. Do not write code.}
\PromptLine{Return only the fields in the schema below. Preserve target distances, offsets,}
\PromptLine{rotation angles, frames, and contact conditions.}
\PromptBlankLine
\PromptLine{\#\#\# STAGE SEMANTICS}
\PromptBlankLine
\PromptLine{The Constraints list represents ONE controller stage.}
\PromptLine{* Cover all required components of the stage; do not drop a component because another is easier to encode.}
\PromptLine{* Stage requirements are conjunctive unless a runtime/\allowbreak{}error message explicitly says top-level OR is supported.}
\PromptLine{* For normal terms, set `logic` to `"and"`.}
\PromptLine{* Do not rely on top-level `logic="or"` for alternatives. If the text gives a supported stop alternative such as geometric progress OR manipulator\_\allowbreak{}high\_\allowbreak{}effort, encode it in the same relevant term by including both condition labels, e.g. `["geometry", "manipulator\_\allowbreak{}high\_\allowbreak{}effort"]`. If the alternative cannot be represented, keep the required representable term and note the omitted alternative in Reasoning.}
\PromptBlankLine
\PromptLine{Condition labels describe satisfaction/\allowbreak{}feedback categories:}
\PromptLine{* `geometry`: pose, motion, orientation, or freeze constraints.}
\PromptLine{* `gripper\_\allowbreak{}opened` /\allowbreak{} `gripper\_\allowbreak{}closed`: intended gripper state metadata; not arm contact effort. `gripper\_\allowbreak{}closed` alone does not prove a grasp.}
\PromptLine{* `manipulator\_\allowbreak{}high\_\allowbreak{}effort` /\allowbreak{} `manipulator\_\allowbreak{}low\_\allowbreak{}effort`: arm/\allowbreak{}TCP interaction effort for pushing, pressing, pulling, or sliding; not finger squeeze.}
\PromptBlankLine
\PromptLine{\#\#\# TERM API}
\PromptBlankLine
\PromptLine{Use ONLY these ops and EXACTLY these args keys:}
\PromptBlankLine
\PromptLine{```json}
\PromptLine{\{}
\PromptLine{\PromptIndent{2}"reach\_\allowbreak{}point": \{}
\PromptLine{\PromptIndent{4}"args": \{}
\PromptLine{\PromptIndent{6}"controlled\_\allowbreak{}entity": "robot\_\allowbreak{}frame | gripped\_\allowbreak{}object",}
\PromptLine{\PromptIndent{6}"controlled\_\allowbreak{}target": "NAME",}
\PromptLine{\PromptIndent{6}"reference\_\allowbreak{}basis": "world | object | robot\_\allowbreak{}frame",}
\PromptLine{\PromptIndent{6}"reference\_\allowbreak{}origin": "NAME | world",}
\PromptLine{\PromptIndent{6}"offset\_\allowbreak{}from\_\allowbreak{}origin\_\allowbreak{}xyz": [float, float, float]}
\PromptLine{\PromptIndent{4}\}}
\PromptLine{\PromptIndent{2}\},}
\PromptLine{\PromptIndent{2}"directional\_\allowbreak{}move": \{}
\PromptLine{\PromptIndent{4}"args": \{}
\PromptLine{\PromptIndent{6}"controlled\_\allowbreak{}entity": "robot\_\allowbreak{}frame | gripped\_\allowbreak{}object",}
\PromptLine{\PromptIndent{6}"controlled\_\allowbreak{}target": "NAME",}
\PromptLine{\PromptIndent{6}"reference\_\allowbreak{}basis": "world | object | robot\_\allowbreak{}frame",}
\PromptLine{\PromptIndent{6}"reference\_\allowbreak{}origin": "NAME | world",}
\PromptLine{\PromptIndent{6}"axis": "+x | -x | +y | -y | +z | -z",}
\PromptLine{\PromptIndent{6}"travel\_\allowbreak{}distance\_\allowbreak{}m": float}
\PromptLine{\PromptIndent{4}\}}
\PromptLine{\PromptIndent{2}\},}
\PromptLine{\PromptIndent{2}"axis\_\allowbreak{}align": \{}
\PromptLine{\PromptIndent{4}"args": \{}
\PromptLine{\PromptIndent{6}"controlled\_\allowbreak{}entity": "robot\_\allowbreak{}frame | gripped\_\allowbreak{}object",}
\PromptLine{\PromptIndent{6}"controlled\_\allowbreak{}target": "NAME",}
\PromptLine{\PromptIndent{6}"reference\_\allowbreak{}basis": "world | object | robot\_\allowbreak{}frame",}
\PromptLine{\PromptIndent{6}"reference\_\allowbreak{}origin": "NAME | world",}
\PromptLine{\PromptIndent{6}"controlled\_\allowbreak{}axis": "+x | -x | +y | -y | +z | -z",}
\PromptLine{\PromptIndent{6}"reference\_\allowbreak{}axis": "+x | -x | +y | -y | +z | -z",}
\PromptLine{\PromptIndent{6}"alignment": "codirectional | antiparallel | bipolar | orthogonal"}
\PromptLine{\PromptIndent{4}\}}
\PromptLine{\PromptIndent{2}\},}
\PromptLine{\PromptIndent{2}"point\_\allowbreak{}toward": \{}
\PromptLine{\PromptIndent{4}"args": \{}
\PromptLine{\PromptIndent{6}"controlled\_\allowbreak{}entity": "robot\_\allowbreak{}frame | gripped\_\allowbreak{}object",}
\PromptLine{\PromptIndent{6}"controlled\_\allowbreak{}target": "NAME",}
\PromptLine{\PromptIndent{6}"controlled\_\allowbreak{}axis": "+x | -x | +y | -y | +z | -z",}
\PromptLine{\PromptIndent{6}"reference\_\allowbreak{}basis": "world | object | robot\_\allowbreak{}frame",}
\PromptLine{\PromptIndent{6}"reference\_\allowbreak{}origin": "NAME | world",}
\PromptLine{\PromptIndent{6}"offset\_\allowbreak{}from\_\allowbreak{}origin\_\allowbreak{}xyz": [float, float, float]}
\PromptLine{\PromptIndent{4}\}}
\PromptLine{\PromptIndent{2}\},}
\PromptLine{\PromptIndent{2}"directional\_\allowbreak{}rotate": \{}
\PromptLine{\PromptIndent{4}"args": \{}
\PromptLine{\PromptIndent{6}"controlled\_\allowbreak{}entity": "robot\_\allowbreak{}frame | gripped\_\allowbreak{}object",}
\PromptLine{\PromptIndent{6}"controlled\_\allowbreak{}target": "NAME",}
\PromptLine{\PromptIndent{6}"reference\_\allowbreak{}basis": "world | object | robot\_\allowbreak{}frame",}
\PromptLine{\PromptIndent{6}"reference\_\allowbreak{}origin": "NAME | world",}
\PromptLine{\PromptIndent{6}"rotation\_\allowbreak{}axis": "+x | -x | +y | -y | +z | -z",}
\PromptLine{\PromptIndent{6}"rotation\_\allowbreak{}angle\_\allowbreak{}rad": float}
\PromptLine{\PromptIndent{4}\}}
\PromptLine{\PromptIndent{2}\},}
\PromptLine{\PromptIndent{2}"freeze\_\allowbreak{}position": \{}
\PromptLine{\PromptIndent{4}"args": \{}
\PromptLine{\PromptIndent{6}"controlled\_\allowbreak{}entity": "robot\_\allowbreak{}frame | gripped\_\allowbreak{}object",}
\PromptLine{\PromptIndent{6}"controlled\_\allowbreak{}target": "NAME"}
\PromptLine{\PromptIndent{4}\}}
\PromptLine{\PromptIndent{2}\},}
\PromptLine{\PromptIndent{2}"freeze\_\allowbreak{}orientation": \{}
\PromptLine{\PromptIndent{4}"args": \{}
\PromptLine{\PromptIndent{6}"controlled\_\allowbreak{}entity": "robot\_\allowbreak{}frame | gripped\_\allowbreak{}object",}
\PromptLine{\PromptIndent{6}"controlled\_\allowbreak{}target": "NAME"}
\PromptLine{\PromptIndent{4}\}}
\PromptLine{\PromptIndent{2}\},}
\PromptLine{\PromptIndent{2}"open\_\allowbreak{}gripper": \{"args": \{\}\},}
\PromptLine{\PromptIndent{2}"close\_\allowbreak{}gripper": \{"args": \{\}\},}
\PromptLine{\PromptIndent{2}"object\_\allowbreak{}grasped": \{"args": \{\}\}}
\PromptLine{\}}
\PromptLine{```}
\PromptBlankLine
\PromptLine{NAME must be a valid robot frame or object/\allowbreak{}object-attached frame from Observations. Do not add or omit args keys.}
\PromptBlankLine
\PromptLine{\#\#\# FIELD MEANINGS}
\PromptBlankLine
\PromptLine{* `controlled\_\allowbreak{}entity="robot\_\allowbreak{}frame"` requires `controlled\_\allowbreak{}target` to be a robot frame.}
\PromptLine{* `controlled\_\allowbreak{}entity="gripped\_\allowbreak{}object"` requires `controlled\_\allowbreak{}target` to be the object/\allowbreak{}object-attached frame currently in the gripper.}
\PromptLine{* `reference\_\allowbreak{}basis` is the coordinate system used for vectors/\allowbreak{}axes; `reference\_\allowbreak{}origin` is the pose defining that basis origin/\allowbreak{}axes.}
\PromptLine{* `offset\_\allowbreak{}from\_\allowbreak{}origin\_\allowbreak{}xyz` is applied at `reference\_\allowbreak{}origin` in `reference\_\allowbreak{}basis`.}
\PromptLine{* `\_\allowbreak{}m` values are meters; `\_\allowbreak{}rad` values are radians.}
\PromptBlankLine
\PromptLine{\#\#\# OPERATOR RULES}
\PromptBlankLine
\PromptLine{* `reach\_\allowbreak{}point`: reach the specified target point; path is unconstrained. Use for concrete targets, pre-grasps, standoffs, and approach-side/\allowbreak{}axis offsets.}
\PromptLine{* `directional\_\allowbreak{}move`: move strictly along one signed axis by positive `travel\_\allowbreak{}distance\_\allowbreak{}m`. Use only when the text explicitly specifies axis/\allowbreak{}direction plus distance; otherwise use `reach\_\allowbreak{}point`.}
\PromptLine{* `axis\_\allowbreak{}align`: align axes with codirectional, antiparallel, bipolar, or orthogonal relation. Use `bipolar` when either direction is acceptable.}
\PromptLine{* `point\_\allowbreak{}toward`: make a controlled signed axis aim at a target point.}
\PromptLine{* `directional\_\allowbreak{}rotate`: rotate about a signed axis by an explicit angle.}
\PromptLine{* Do not combine `directional\_\allowbreak{}rotate` with `axis\_\allowbreak{}align` in the same stage when the `axis\_\allowbreak{}align` constrains or references the frame being rotated. Put approach/\allowbreak{}pre-grasp axis alignment in an earlier stage; the rotation stage should keep the finite `directional\_\allowbreak{}rotate` goal.}
\PromptLine{* `freeze\_\allowbreak{}position`: keep current position fixed; do not use as a target-reaching goal.}
\PromptLine{* `freeze\_\allowbreak{}orientation`: keep current orientation fixed; do not use as a goal-alignment operator. If the stage establishes a new axis relation, emit `axis\_\allowbreak{}align` or `directional\_\allowbreak{}rotate`; `freeze\_\allowbreak{}orientation` alone is invalid for that.}
\PromptLine{* `open\_\allowbreak{}gripper`: gripper open/\allowbreak{}released at success.}
\PromptLine{* `close\_\allowbreak{}gripper`: mechanical finger closure only when a real held object is not required.}
\PromptLine{* `object\_\allowbreak{}grasped`: object must be physically held; use for grasp, hold, keep holding, and any later moving/\allowbreak{}pulling/\allowbreak{}carrying/\allowbreak{}aligning stage that assumes the object remains in the gripper. Do not emit both `close\_\allowbreak{}gripper` and `object\_\allowbreak{}grasped` in one list.}
\PromptBlankLine
\PromptLine{General constraints:}
\PromptLine{1. At most one position op in the whole list: `reach\_\allowbreak{}point` OR `directional\_\allowbreak{}move` OR `freeze\_\allowbreak{}position`.}
\PromptLine{2. Every constraint must be justified by the current stage requirements.}
\PromptLine{3. Describe the intended contact/\allowbreak{}fit precisely through geometry and conditions.}
\PromptLine{4. If a required clause cannot be represented, omit that clause from Constraints and state it in Reasoning.}
\PromptBlankLine
\PromptLine{\#\#\# REQUIRED OUTPUT}
\PromptBlankLine
\PromptLine{Return exactly one JSON object:}
\PromptBlankLine
\PromptLine{```json}
\PromptLine{\{}
\PromptLine{\PromptIndent{2}"Reasoning": "string",}
\PromptLine{\PromptIndent{2}"Constraints": [}
\PromptLine{\PromptIndent{4}\{}
\PromptLine{\PromptIndent{6}"logic": "and | or",}
\PromptLine{\PromptIndent{6}"condition": ["geometry", "gripper\_\allowbreak{}opened", "gripper\_\allowbreak{}closed", "manipulator\_\allowbreak{}high\_\allowbreak{}effort", "manipulator\_\allowbreak{}low\_\allowbreak{}effort"],}
\PromptLine{\PromptIndent{6}"op": "reach\_\allowbreak{}point | directional\_\allowbreak{}move | axis\_\allowbreak{}align | point\_\allowbreak{}toward | directional\_\allowbreak{}rotate | freeze\_\allowbreak{}position | freeze\_\allowbreak{}orientation | open\_\allowbreak{}gripper | close\_\allowbreak{}gripper | object\_\allowbreak{}grasped",}
\PromptLine{\PromptIndent{6}"args": \{\}}
\PromptLine{\PromptIndent{4}\}}
\PromptLine{\PromptIndent{2}]}
\PromptLine{\}}
\PromptLine{```}
\PromptBlankLine
\PromptLine{Reasoning briefly explains success, each constraint, any condition-label alternative, and any unrepresented clause.}
    \vspace{0.4\baselineskip}%
    \hrule
  \endgroup
  \par\addvspace{0.8\baselineskip}%

\subsubsection{Typed-relation correction}
  \refstepcounter{sunpromptlisting}\label{app:lst:sun_constraint_correct_context}%
  \par\addvspace{0.8\baselineskip}%
  {\centering\small Listing~\thesunpromptlisting: System context for typed-relation correction\par}%
  \nobreak\vspace{0.25\baselineskip}%
  \begingroup
    \renewcommand{\baselinestretch}{1}%
    \fontsize{7.5}{9}\selectfont
    \fontencoding{T1}\fontfamily{lmtt}\selectfont
    \ifdefined\pdfnoligatures\pdfnoligatures
    \hyphenchar\font=-1
    \setlength{\parindent}{0pt}%
    \setlength{\parskip}{0pt}%
    \rightskip=0pt plus 1fil\relax
    \emergencystretch=3em
    \hrule
    \vspace{0.4\baselineskip}%
\PromptLine{You correct the PREVIOUS grounded output for the CURRENT stage after a failure. You do not write code.}
\PromptLine{Use the error message as the primary diagnostic signal. Make the smallest change needed, keeping correct constraints verbatim.}
\PromptLine{The corrected Constraints list is one controller stage: prefer `logic="and"`; do not rely on top-level OR unless the runtime/\allowbreak{}error message explicitly supports it. Encode representable stop alternatives inside the same term's `condition` labels.}
    \vspace{0.4\baselineskip}%
    \hrule
  \endgroup
  \par\addvspace{0.8\baselineskip}%

  \refstepcounter{sunpromptlisting}\label{app:lst:sun_constraint_correct_prompt}%
  \par\addvspace{0.8\baselineskip}%
  {\centering\small Listing~\thesunpromptlisting: User template for typed-relation correction\par}%
  \nobreak\vspace{0.25\baselineskip}%
  \begingroup
    \renewcommand{\baselinestretch}{1}%
    \fontsize{7.5}{9}\selectfont
    \fontencoding{T1}\fontfamily{lmtt}\selectfont
    \ifdefined\pdfnoligatures\pdfnoligatures
    \hyphenchar\font=-1
    \setlength{\parindent}{0pt}%
    \setlength{\parskip}{0pt}%
    \rightskip=0pt plus 1fil\relax
    \emergencystretch=3em
    \hrule
    \vspace{0.4\baselineskip}%
\PromptLine{\#\#\# SCOPE \& INPUTS}
\PromptBlankLine
\PromptLine{Task}
\PromptLine{* Current grounded stage requirements: \{task\_\allowbreak{}description\}}
\PromptLine{* Completed subtasks, context only: \{completed\}}
\PromptLine{* Pending subtasks, context only: \{pending\}}
\PromptBlankLine
\PromptLine{Observations}
\PromptLine{* Robot Frames: \{frames\}}
\PromptLine{* Observed Objects, visible + history: \{history\}}
\PromptLine{* Only refer to names in Robot Frames and Observed Objects; there are no others. "world" is a reserved keyword.}
\PromptBlankLine
\PromptLine{Previous grounded output to correct:}
\PromptLine{* \{previous\_\allowbreak{}grounding\}}
\PromptBlankLine
\PromptLine{Current error message, primary signal:}
\PromptLine{* \{error\_\allowbreak{}message\}}
\PromptBlankLine
\PromptLine{\#\#\# YOUR JOB}
\PromptBlankLine
\PromptLine{Return only the fields in the schema below. Correct geometry/\allowbreak{}contact conditions}
\PromptLine{while preserving the intended target and required physical result.}
\PromptBlankLine
\PromptLine{Diagnose the failure, then return corrected Constraint JSON that:}
\PromptLine{1. Uses only TermAPI ops and exactly their allowed args keys.}
\PromptLine{2. References only valid observed names.}
\PromptLine{3. Preserves all intended success criteria for the current stage.}
\PromptLine{4. Changes as little as possible: edit only fields implicated by the error, schema/\allowbreak{}API violations, missing coverage, unsupported top-level OR, or the one-position-op rule. Keep correct fields verbatim.}
\PromptBlankLine
\PromptLine{\#\#\# STAGE SEMANTICS AND REPAIR RULES}
\PromptBlankLine
\PromptLine{The Constraints list represents ONE controller stage.}
\PromptLine{* Cover all required components; do not drop a component because another is easier to encode.}
\PromptLine{* Stage requirements are conjunctive unless a runtime/\allowbreak{}error message explicitly says top-level OR is supported.}
\PromptLine{* For normal terms, set `logic` to `"and"`.}
\PromptLine{* Do not rely on top-level `logic="or"` for alternatives. If the text gives a supported stop alternative such as geometric progress OR manipulator\_\allowbreak{}high\_\allowbreak{}effort, encode it in the same relevant term by including both condition labels, e.g. `["geometry", "manipulator\_\allowbreak{}high\_\allowbreak{}effort"]`. If the alternative cannot be represented, keep the required representable term and note the omitted alternative in Diagnosis or Reasoning.}
\PromptLine{* Across the whole list, keep at most one position op: `reach\_\allowbreak{}point` OR `directional\_\allowbreak{}move` OR `freeze\_\allowbreak{}position`. If violated, keep the one that best matches the explicit positional requirement and mention the dropped op in Diagnosis.}
\PromptLine{* Remove only exact duplicates: same `op` and identical `args`; merge `condition` labels by unique union. Do not collapse merely similar constraints.}
\PromptLine{* `gripper\_\allowbreak{}opened` /\allowbreak{} `gripper\_\allowbreak{}closed` are gripper-state metadata, not arm contact effort. `manipulator\_\allowbreak{}high\_\allowbreak{}effort` /\allowbreak{} `manipulator\_\allowbreak{}low\_\allowbreak{}effort` are arm/\allowbreak{}TCP contact effort, not finger squeeze. Use `object\_\allowbreak{}grasped` when a real held object is required; do not represent contact/\allowbreak{}effort with gripper ops.}
\PromptBlankLine
\PromptLine{\#\#\# TERM API}
\PromptBlankLine
\PromptLine{Use ONLY these ops and EXACTLY these args keys:}
\PromptBlankLine
\PromptLine{```json}
\PromptLine{\{}
\PromptLine{\PromptIndent{2}"reach\_\allowbreak{}point": \{}
\PromptLine{\PromptIndent{4}"args": \{}
\PromptLine{\PromptIndent{6}"controlled\_\allowbreak{}entity": "robot\_\allowbreak{}frame | gripped\_\allowbreak{}object",}
\PromptLine{\PromptIndent{6}"controlled\_\allowbreak{}target": "NAME",}
\PromptLine{\PromptIndent{6}"reference\_\allowbreak{}basis": "world | object | robot\_\allowbreak{}frame",}
\PromptLine{\PromptIndent{6}"reference\_\allowbreak{}origin": "NAME | world",}
\PromptLine{\PromptIndent{6}"offset\_\allowbreak{}from\_\allowbreak{}origin\_\allowbreak{}xyz": [float, float, float]}
\PromptLine{\PromptIndent{4}\}}
\PromptLine{\PromptIndent{2}\},}
\PromptLine{\PromptIndent{2}"directional\_\allowbreak{}move": \{}
\PromptLine{\PromptIndent{4}"args": \{}
\PromptLine{\PromptIndent{6}"controlled\_\allowbreak{}entity": "robot\_\allowbreak{}frame | gripped\_\allowbreak{}object",}
\PromptLine{\PromptIndent{6}"controlled\_\allowbreak{}target": "NAME",}
\PromptLine{\PromptIndent{6}"reference\_\allowbreak{}basis": "world | object | robot\_\allowbreak{}frame",}
\PromptLine{\PromptIndent{6}"reference\_\allowbreak{}origin": "NAME | world",}
\PromptLine{\PromptIndent{6}"axis": "+x | -x | +y | -y | +z | -z",}
\PromptLine{\PromptIndent{6}"travel\_\allowbreak{}distance\_\allowbreak{}m": float}
\PromptLine{\PromptIndent{4}\}}
\PromptLine{\PromptIndent{2}\},}
\PromptLine{\PromptIndent{2}"axis\_\allowbreak{}align": \{}
\PromptLine{\PromptIndent{4}"args": \{}
\PromptLine{\PromptIndent{6}"controlled\_\allowbreak{}entity": "robot\_\allowbreak{}frame | gripped\_\allowbreak{}object",}
\PromptLine{\PromptIndent{6}"controlled\_\allowbreak{}target": "NAME",}
\PromptLine{\PromptIndent{6}"reference\_\allowbreak{}basis": "world | object | robot\_\allowbreak{}frame",}
\PromptLine{\PromptIndent{6}"reference\_\allowbreak{}origin": "NAME | world",}
\PromptLine{\PromptIndent{6}"controlled\_\allowbreak{}axis": "+x | -x | +y | -y | +z | -z",}
\PromptLine{\PromptIndent{6}"reference\_\allowbreak{}axis": "+x | -x | +y | -y | +z | -z",}
\PromptLine{\PromptIndent{6}"alignment": "codirectional | antiparallel | bipolar | orthogonal"}
\PromptLine{\PromptIndent{4}\}}
\PromptLine{\PromptIndent{2}\},}
\PromptLine{\PromptIndent{2}"point\_\allowbreak{}toward": \{}
\PromptLine{\PromptIndent{4}"args": \{}
\PromptLine{\PromptIndent{6}"controlled\_\allowbreak{}entity": "robot\_\allowbreak{}frame | gripped\_\allowbreak{}object",}
\PromptLine{\PromptIndent{6}"controlled\_\allowbreak{}target": "NAME",}
\PromptLine{\PromptIndent{6}"controlled\_\allowbreak{}axis": "+x | -x | +y | -y | +z | -z",}
\PromptLine{\PromptIndent{6}"reference\_\allowbreak{}basis": "world | object | robot\_\allowbreak{}frame",}
\PromptLine{\PromptIndent{6}"reference\_\allowbreak{}origin": "NAME | world",}
\PromptLine{\PromptIndent{6}"offset\_\allowbreak{}from\_\allowbreak{}origin\_\allowbreak{}xyz": [float, float, float]}
\PromptLine{\PromptIndent{4}\}}
\PromptLine{\PromptIndent{2}\},}
\PromptLine{\PromptIndent{2}"directional\_\allowbreak{}rotate": \{}
\PromptLine{\PromptIndent{4}"args": \{}
\PromptLine{\PromptIndent{6}"controlled\_\allowbreak{}entity": "robot\_\allowbreak{}frame | gripped\_\allowbreak{}object",}
\PromptLine{\PromptIndent{6}"controlled\_\allowbreak{}target": "NAME",}
\PromptLine{\PromptIndent{6}"reference\_\allowbreak{}basis": "world | object | robot\_\allowbreak{}frame",}
\PromptLine{\PromptIndent{6}"reference\_\allowbreak{}origin": "NAME | world",}
\PromptLine{\PromptIndent{6}"rotation\_\allowbreak{}axis": "+x | -x | +y | -y | +z | -z",}
\PromptLine{\PromptIndent{6}"rotation\_\allowbreak{}angle\_\allowbreak{}rad": float}
\PromptLine{\PromptIndent{4}\}}
\PromptLine{\PromptIndent{2}\},}
\PromptLine{\PromptIndent{2}"freeze\_\allowbreak{}position": \{}
\PromptLine{\PromptIndent{4}"args": \{}
\PromptLine{\PromptIndent{6}"controlled\_\allowbreak{}entity": "robot\_\allowbreak{}frame | gripped\_\allowbreak{}object",}
\PromptLine{\PromptIndent{6}"controlled\_\allowbreak{}target": "NAME"}
\PromptLine{\PromptIndent{4}\}}
\PromptLine{\PromptIndent{2}\},}
\PromptLine{\PromptIndent{2}"freeze\_\allowbreak{}orientation": \{}
\PromptLine{\PromptIndent{4}"args": \{}
\PromptLine{\PromptIndent{6}"controlled\_\allowbreak{}entity": "robot\_\allowbreak{}frame | gripped\_\allowbreak{}object",}
\PromptLine{\PromptIndent{6}"controlled\_\allowbreak{}target": "NAME"}
\PromptLine{\PromptIndent{4}\}}
\PromptLine{\PromptIndent{2}\},}
\PromptLine{\PromptIndent{2}"open\_\allowbreak{}gripper": \{"args": \{\}\},}
\PromptLine{\PromptIndent{2}"close\_\allowbreak{}gripper": \{"args": \{\}\},}
\PromptLine{\PromptIndent{2}"object\_\allowbreak{}grasped": \{"args": \{\}\}}
\PromptLine{\}}
\PromptLine{```}
\PromptBlankLine
\PromptLine{NAME must be a valid robot frame or object/\allowbreak{}object-attached frame from Observations. Do not add or omit args keys.}
\PromptBlankLine
\PromptLine{\#\#\# FIELD AND OPERATOR RULES}
\PromptBlankLine
\PromptLine{* `controlled\_\allowbreak{}entity="robot\_\allowbreak{}frame"` requires a robot-frame target; `controlled\_\allowbreak{}entity="gripped\_\allowbreak{}object"` requires an object/\allowbreak{}object-attached frame currently in the gripper.}
\PromptLine{* `reference\_\allowbreak{}basis` is the coordinate system for vectors/\allowbreak{}axes; `reference\_\allowbreak{}origin` defines the pose. `\_\allowbreak{}m` = meters; `\_\allowbreak{}rad` = radians.}
\PromptLine{* `reach\_\allowbreak{}point`: concrete target point, pre-grasp, standoff, or approach offset; path unconstrained.}
\PromptLine{* `directional\_\allowbreak{}move`: strictly along one signed axis by positive distance; only when axis/\allowbreak{}direction plus distance is explicit. Otherwise use `reach\_\allowbreak{}point`.}
\PromptLine{* `axis\_\allowbreak{}align`: axis relation; use `bipolar` when either direction is acceptable.}
\PromptLine{* `point\_\allowbreak{}toward`: make a controlled axis aim at a target point.}
\PromptLine{* `directional\_\allowbreak{}rotate`: explicit rotation about an axis by an angle.}
\PromptLine{* Do not combine `directional\_\allowbreak{}rotate` with `axis\_\allowbreak{}align` in the same stage when the `axis\_\allowbreak{}align` constrains or references the frame being rotated. Put approach/\allowbreak{}pre-grasp axis alignment in an earlier stage; the rotation stage should keep the finite `directional\_\allowbreak{}rotate` goal.}
\PromptLine{* `freeze\_\allowbreak{}position`: keep current position fixed; not a target-reaching goal.}
\PromptLine{* `freeze\_\allowbreak{}orientation`: keep current orientation fixed; not goal alignment. If the stage establishes a new axis relation, corrected output must contain `axis\_\allowbreak{}align` or `directional\_\allowbreak{}rotate`.}
\PromptLine{* `open\_\allowbreak{}gripper`: open/\allowbreak{}released. `close\_\allowbreak{}gripper`: mechanical closure only. `object\_\allowbreak{}grasped`: real held object for grasp/\allowbreak{}hold/\allowbreak{}keep holding and later carrying/\allowbreak{}pulling/\allowbreak{}manipulating. Do not emit both `close\_\allowbreak{}gripper` and `object\_\allowbreak{}grasped`.}
\PromptBlankLine
\PromptLine{\#\#\# REQUIRED OUTPUT}
\PromptBlankLine
\PromptLine{Return exactly one JSON object:}
\PromptBlankLine
\PromptLine{```json}
\PromptLine{\{}
\PromptLine{\PromptIndent{2}"Diagnosis": "string",}
\PromptLine{\PromptIndent{2}"Reasoning": "string",}
\PromptLine{\PromptIndent{2}"Constraints": [}
\PromptLine{\PromptIndent{4}\{}
\PromptLine{\PromptIndent{6}"logic": "and | or",}
\PromptLine{\PromptIndent{6}"condition": ["geometry", "gripper\_\allowbreak{}opened", "gripper\_\allowbreak{}closed", "manipulator\_\allowbreak{}high\_\allowbreak{}effort", "manipulator\_\allowbreak{}low\_\allowbreak{}effort"],}
\PromptLine{\PromptIndent{6}"op": "reach\_\allowbreak{}point | directional\_\allowbreak{}move | axis\_\allowbreak{}align | point\_\allowbreak{}toward | directional\_\allowbreak{}rotate | freeze\_\allowbreak{}position | freeze\_\allowbreak{}orientation | open\_\allowbreak{}gripper | close\_\allowbreak{}gripper | object\_\allowbreak{}grasped",}
\PromptLine{\PromptIndent{6}"args": \{\}}
\PromptLine{\PromptIndent{4}\}}
\PromptLine{\PromptIndent{2}]}
\PromptLine{\}}
\PromptLine{```}
\PromptBlankLine
\PromptLine{Diagnosis ties the failure to the error message. Reasoning explains success, each constraint, any condition-label alternative, and any unrepresented clause. Constraints are the corrected ordered list.}
    \vspace{0.4\baselineskip}%
    \hrule
  \endgroup
  \par\addvspace{0.8\baselineskip}%

\subsection{Learning and calibration prompts}

These templates expose scalar choices and measured evidence without permitting
the model to invent a new operator, stage, reward implementation, or task
evaluator. Initial selection adjusts weights without rollout evidence. The
BC-prior diagnostic variant then uses a deterministic prior rollout, not
residual-RL training statistics, to identify components blocking progress.
Its system prompt concatenates the shared supervision contract, the source's
BC-specific note, and the stagewise weight-adjustment instructions. The separate
residual-authority pair receives learning traces. None of these templates
receives or edits the code-owned tolerance fields described in
Appendix~\ref{app:sun_receiver_policy}.

An optional video summarizer can produce evidence, but it has no write
authority over the program. Its prose suggestions are constrained by the
system prompts below, which define the editable fields, numeric ranges, and
output schemas. Accordingly, this appendix includes the formation and active
initialization/calibration templates, while excluding the deprecated
stage/global-runner variants, video-captioning instructions, and run-specific
transcripts.

\subsubsection{Initial reward-weight selection}
  \refstepcounter{sunpromptlisting}\label{app:lst:sun_initial_sweep_system}%
  \par\addvspace{0.8\baselineskip}%
  {\centering\small Listing~\thesunpromptlisting: System prompt for initial reward-weight selection\par}%
  \nobreak\vspace{0.25\baselineskip}%
  \begingroup
    \renewcommand{\baselinestretch}{1}%
    \fontsize{7.5}{9}\selectfont
    \fontencoding{T1}\fontfamily{lmtt}\selectfont
    \ifdefined\pdfnoligatures\pdfnoligatures
    \hyphenchar\font=-1
    \setlength{\parindent}{0pt}%
    \setlength{\parskip}{0pt}%
    \rightskip=0pt plus 1fil\relax
    \emergencystretch=3em
    \hrule
    \vspace{0.4\baselineskip}%
\PromptBlankLine
\PromptLine{You are a supervisor for a reinforcement learning agent. The agent is training on a multi-stage task where each stage has its own set of reward components.}
\PromptBlankLine
\PromptLine{These reward components are based on cost terms that have been effective in prior optimal control settings. Do not add, remove, or redefine reward components.}
\PromptBlankLine
\PromptLine{Each reward component *i* defines:}
\PromptBlankLine
\PromptLine{1. a goal requirement,}
\PromptLine{2. a shaping reward function that increases as the agent gets closer to meeting that requirement, and}
\PromptLine{3. a constraint check that measures whether the goal requirement has been met.}
\PromptBlankLine
\PromptLine{A stage is considered complete when all component constraints are satisfied.}
\PromptBlankLine
\PromptLine{Task names, object labels, stage numbers, and component indices identify the}
\PromptLine{provided data; they do not determine which component deserves more weight.}
\PromptLine{Base each decision on the current physical requirement and, when provided,}
\PromptLine{measured evidence. Do not infer a stored solution from a familiar name.}
\PromptBlankLine
\PromptLine{Short rule examples, not task solutions:}
\PromptLine{- Weights [1, 2] and [2, 4] express the same relative preference after normalization.}
\PromptLine{- A pose hold during release and a pose requirement that validates the final}
\PromptLine{\PromptIndent{2}object state can have different roles even when both use the same operator.}
\PromptBlankLine
\PromptLine{Physical roles used by the weight rules:}
\PromptLine{- Broad body: roughly solid/\allowbreak{}boxlike/\allowbreak{}round grasp cross-section \textgreater{}= 0.05 m,}
\PromptLine{\PromptIndent{2}without a narrow axis or receiver-fit requirement.}
\PromptLine{- Small solid body: cross-section above 0.03 m and below 0.05 m; size alone}
\PromptLine{\PromptIndent{2}does not make orientation mission-critical.}
\PromptLine{- Small/\allowbreak{}slender: cross-section \textless{}= 0.03 m, or a long/\allowbreak{}thin/\allowbreak{}axis-critical shape.}
\PromptLine{- Release guard: robot/\allowbreak{}tool pose held while opening after a supported result;}
\PromptLine{\PromptIndent{2}distinguish this from a component checking the object's final validity.}
\PromptLine{- Precision progress: the traveled amount itself validates current contact,}
\PromptLine{\PromptIndent{2}fit, or actuation; free-space waypoints are not precision progress.}
\PromptBlankLine
\PromptLine{You initialize reward-component Weights for ONE stage before RL.}
\PromptLine{Adjust ONLY reward-component Weights.}
\PromptBlankLine
\PromptLine{Do NOT change reward/\allowbreak{}constraint definitions, indices, ordering, logic}
\PromptLine{structure, targets, distances, or any other configuration.}
\PromptBlankLine
\PromptLine{Use only the provided task/\allowbreak{}stage semantics, observed objects, adjacent-stage}
\PromptLine{context, component explanations, and current weights.}
\PromptLine{Do NOT assume rollout statistics, success rates, reward traces, or hidden task}
\PromptLine{facts.}
\PromptBlankLine
\PromptLine{Weights:}
\PromptLine{- Set weights by semantic role, not performance. Reward is normalized by sum(weights), so tune relative ratios.}
\PromptLine{- Default init-sweep weight is 1.0. Use simple ratios.}
\PromptLine{- Broad free-space progress, broad-body ordinary pose holds, broad-body top-down/\allowbreak{}side orientation maintenance, and non-critical maintenance terms usually stay in [1.0, 1.5].}
\PromptLine{- For broad\_\allowbreak{}body tasks with no small/\allowbreak{}slender, receiver/\allowbreak{}opening, constrained-feature, or final-object-validity requirement, avoid weights above 2.0. Grasp/\allowbreak{}object-state may be 2.0 if losing the state invalidates all progress, but do not inflate it to 2.5-3.0 on init sweep.}
\PromptLine{- Direct fit/\allowbreak{}contact, small-cross-section grasp-lock, constrained-feature manipulation, precision directional progress, mission-critical angular constraints, and final object validity may use [1.5, 3.0] depending on whether losing that component invalidates the stage.}
\PromptLine{- Open-gripper setup/\allowbreak{}contact orientation terms and broad sliding-pull orientation terms are usually [1.0, 1.5] and should not be weighted as mission-critical angular constraints unless the current stage truly requires angular fit, insertion, final object validity, or a force/\allowbreak{}torque axis.}
\PromptLine{- Gripper/\allowbreak{}object state components are important when they define contact, release, or grasp validity; usually use 1.5, use 2.0 for broad-body state validity when losing state invalidates progress, and reserve 2.5-3.0 for small/\allowbreak{}slender, receiver/\allowbreak{}opening, constrained-feature, or final-object-validity cases.}
\PromptLine{- Release\_\allowbreak{}guard robot/\allowbreak{}tool pose holds usually stay in [1.0, 1.5]; do not give them narrow-feature/\allowbreak{}manipulation-level weights unless the stage explicitly says the held tool pose itself is the contact-validity condition.}
\PromptLine{- Avoid weight inflation: do not set every core component above 2.0 on the init sweep, and do not make broad-body orientation/\allowbreak{}maintenance terms higher than the primary progress/\allowbreak{}contact term.}
\PromptLine{- weight must be finite and in [0.1, 10.0]. Correct any out-of-bound value.}
\PromptBlankLine
\PromptLine{Return ONLY one valid JSON object with exactly:}
\PromptLine{"Diagnosis", "Reasoning", "Weights"}
\PromptBlankLine
\PromptLine{"Diagnosis": briefly list the semantic role of each tunable component.}
\PromptLine{"Reasoning": briefly justify the labels and relative weight ratios.}
\PromptLine{"Weights" MUST contain EXACTLY the same indices as current weights.}
    \vspace{0.4\baselineskip}%
    \hrule
  \endgroup
  \par\addvspace{0.8\baselineskip}%

  \refstepcounter{sunpromptlisting}\label{app:lst:sun_initial_sweep_user}%
  \par\addvspace{0.8\baselineskip}%
  {\centering\small Listing~\thesunpromptlisting: User template for initial reward-weight selection\par}%
  \nobreak\vspace{0.25\baselineskip}%
  \begingroup
    \renewcommand{\baselinestretch}{1}%
    \fontsize{7.5}{9}\selectfont
    \fontencoding{T1}\fontfamily{lmtt}\selectfont
    \ifdefined\pdfnoligatures\pdfnoligatures
    \hyphenchar\font=-1
    \setlength{\parindent}{0pt}%
    \setlength{\parskip}{0pt}%
    \rightskip=0pt plus 1fil\relax
    \emergencystretch=3em
    \hrule
    \vspace{0.4\baselineskip}%
\PromptBlankLine
\PromptLine{Overall task description: \{task\_\allowbreak{}description\}}
\PromptBlankLine
\PromptLine{Detectable objects: \{detectable\_\allowbreak{}objects\}}
\PromptBlankLine
\PromptLine{Stage index/\allowbreak{}total number of stages: \{stage\_\allowbreak{}id\}/\allowbreak{}\{stage\_\allowbreak{}count\}}
\PromptBlankLine
\PromptLine{Stage description: \{atomic\_\allowbreak{}action\}}
\PromptBlankLine
\PromptLine{Current subtask constraint ops/\allowbreak{}args:}
\PromptLine{\{current\_\allowbreak{}constraint\_\allowbreak{}ops\}}
\PromptBlankLine
\PromptLine{Adjacent subtask context (for transition semantics only; tune ONLY the current stage):}
\PromptLine{Previous subtask:}
\PromptLine{\{previous\_\allowbreak{}stage\_\allowbreak{}context\}}
\PromptBlankLine
\PromptLine{Next subtask:}
\PromptLine{\{next\_\allowbreak{}stage\_\allowbreak{}context\}}
\PromptBlankLine
\PromptLine{Index-ordered reward component explanations:}
\PromptLine{\{constraints\_\allowbreak{}explanations\}}
\PromptBlankLine
\PromptLine{Current weights:}
\PromptLine{\{current\_\allowbreak{}weights\}}
    \vspace{0.4\baselineskip}%
    \hrule
  \endgroup
  \par\addvspace{0.8\baselineskip}%

\subsubsection{BC-prior-informed reward-weight adjustment}
  \refstepcounter{sunpromptlisting}\label{app:lst:sun_stage_weight_system}%
  \par\addvspace{0.8\baselineskip}%
  {\centering\small Listing~\thesunpromptlisting: System prompt for BC-prior-informed weight adjustment\par}%
  \nobreak\vspace{0.25\baselineskip}%
  \begingroup
    \renewcommand{\baselinestretch}{1}%
    \fontsize{7.5}{9}\selectfont
    \fontencoding{T1}\fontfamily{lmtt}\selectfont
    \ifdefined\pdfnoligatures\pdfnoligatures
    \hyphenchar\font=-1
    \setlength{\parindent}{0pt}%
    \setlength{\parskip}{0pt}%
    \rightskip=0pt plus 1fil\relax
    \emergencystretch=3em
    \hrule
    \vspace{0.4\baselineskip}%
\PromptBlankLine
\PromptLine{You are a supervisor for a reinforcement learning agent. The agent is training on a multi-stage task where each stage has its own set of reward components.}
\PromptBlankLine
\PromptLine{These reward components are based on cost terms that have been effective in prior optimal control settings. Do not add, remove, or redefine reward components.}
\PromptBlankLine
\PromptLine{Each reward component *i* defines:}
\PromptBlankLine
\PromptLine{1. a goal requirement,}
\PromptLine{2. a shaping reward function that increases as the agent gets closer to meeting that requirement, and}
\PromptLine{3. a constraint check that measures whether the goal requirement has been met.}
\PromptBlankLine
\PromptLine{A stage is considered complete when all component constraints are satisfied.}
\PromptBlankLine
\PromptLine{Task names, object labels, stage numbers, and component indices identify the}
\PromptLine{provided data; they do not determine which component deserves more weight.}
\PromptLine{Base each decision on the current physical requirement and, when provided,}
\PromptLine{measured evidence. Do not infer a stored solution from a familiar name.}
\PromptBlankLine
\PromptLine{Short rule examples, not task solutions:}
\PromptLine{- Weights [1, 2] and [2, 4] express the same relative preference after normalization.}
\PromptLine{- A pose hold during release and a pose requirement that validates the final}
\PromptLine{\PromptIndent{2}object state can have different roles even when both use the same operator.}
\PromptBlankLine
\PromptLine{Physical roles used by the weight rules:}
\PromptLine{- Broad body: roughly solid/\allowbreak{}boxlike/\allowbreak{}round grasp cross-section \textgreater{}= 0.05 m,}
\PromptLine{\PromptIndent{2}without a narrow axis or receiver-fit requirement.}
\PromptLine{- Small solid body: cross-section above 0.03 m and below 0.05 m; size alone}
\PromptLine{\PromptIndent{2}does not make orientation mission-critical.}
\PromptLine{- Small/\allowbreak{}slender: cross-section \textless{}= 0.03 m, or a long/\allowbreak{}thin/\allowbreak{}axis-critical shape.}
\PromptLine{- Release guard: robot/\allowbreak{}tool pose held while opening after a supported result;}
\PromptLine{\PromptIndent{2}distinguish this from a component checking the object's final validity.}
\PromptLine{- Precision progress: the traveled amount itself validates current contact,}
\PromptLine{\PromptIndent{2}fit, or actuation; free-space waypoints are not precision progress.}
\PromptBlankLine
\PromptLine{You are doing an initial weight acquisition pass from a deterministic BC-prior rollout.}
\PromptLine{The rollout statistics are not RL-training results; they show where the BC prior times out.}
\PromptLine{The timeout/\allowbreak{}stuck component rates are the key evidence: high not\_\allowbreak{}satisfied means that}
\PromptLine{component blocked the BC prior for envs that reached and timed out in that stage.}
\PromptBlankLine
\PromptLine{Adjust ONLY reward-component Weights for the current stage.}
\PromptBlankLine
\PromptLine{You tune ONE stage by adjusting ONLY reward-component Weights.}
\PromptBlankLine
\PromptLine{Do NOT change reward/\allowbreak{}constraint definitions, indices, ordering, logic structure, or any other configuration.}
\PromptBlankLine
\PromptLine{Use stage semantics and population stats.}
\PromptLine{Reward is normalized by sum(weights), so tune relative ratios.}
\PromptBlankLine
\PromptLine{For each component, identify its semantic role and main blocker:}
\PromptLine{- poor progress /\allowbreak{} control}
\PromptLine{- weight tradeoff}
\PromptLine{- required state not maintained}
\PromptBlankLine
\PromptLine{Judge meaningful completion by whether the resulting state is actually suitable for the intended interaction, not merely whether predicates became true.}
\PromptLine{Do NOT use weighted reward magnitude alone as evidence of geometric closeness.}
\PromptBlankLine
\PromptLine{Return ONLY one valid JSON object (no markdown, no code fences, no extra text).}
\PromptLine{Top-level keys MUST be EXACTLY:}
\PromptLine{\PromptIndent{2}"Diagnosis", "Reasoning", "Weights"}
\PromptBlankLine
\PromptLine{- "Weights" MUST contain EXACTLY the same indices as the provided current weights.}
\PromptLine{- Do NOT add/\allowbreak{}remove/\allowbreak{}rename indices.}
\PromptBlankLine
\PromptLine{Weights:}
\PromptLine{- finite, final value in [0.1, 10.0]}
\PromptLine{- per-index step limit: new\_\allowbreak{}w[i] in [0.5*old\_\allowbreak{}w[i], 2.0*old\_\allowbreak{}w[i]]}
\PromptLine{- if a current weight lies outside the hard bounds, you MUST correct it into the hard-bound interval on this pass}
\PromptBlankLine
\PromptLine{Optimization priority:}
\PromptLine{1) Increase meaningful stage completion.}
\PromptLine{2) Preserve semantic validity of the interaction.}
\PromptLine{3) Improve precision only if (1) and (2) do not worsen.}
\PromptLine{4) Improve efficiency only if (1), (2), and (3) do not worsen.}
    \vspace{0.4\baselineskip}%
    \hrule
  \endgroup
  \par\addvspace{0.8\baselineskip}%

  \refstepcounter{sunpromptlisting}\label{app:lst:sun_stage_weight_user}%
  \par\addvspace{0.8\baselineskip}%
  {\centering\small Listing~\thesunpromptlisting: User template for BC-prior-informed weight adjustment\par}%
  \nobreak\vspace{0.25\baselineskip}%
  \begingroup
    \renewcommand{\baselinestretch}{1}%
    \fontsize{7.5}{9}\selectfont
    \fontencoding{T1}\fontfamily{lmtt}\selectfont
    \ifdefined\pdfnoligatures\pdfnoligatures
    \hyphenchar\font=-1
    \setlength{\parindent}{0pt}%
    \setlength{\parskip}{0pt}%
    \rightskip=0pt plus 1fil\relax
    \emergencystretch=3em
    \hrule
    \vspace{0.4\baselineskip}%
\PromptBlankLine
\PromptLine{Overall task description: \{task\_\allowbreak{}description\}}
\PromptBlankLine
\PromptLine{Detectable objects: \{detectable\_\allowbreak{}objects\}}
\PromptBlankLine
\PromptLine{Stage index/\allowbreak{}total number of stages: \{stage\_\allowbreak{}id\}/\allowbreak{}\{stage\_\allowbreak{}count\}}
\PromptBlankLine
\PromptLine{Stage description: \{atomic\_\allowbreak{}action\}}
\PromptBlankLine
\PromptLine{Index-ordered reward component explanations:}
\PromptLine{\{constraints\_\allowbreak{}explanations\}}
\PromptBlankLine
\PromptLine{Current weights:}
\PromptLine{\{current\_\allowbreak{}weights\}}
\PromptBlankLine
\PromptLine{========================}
\PromptLine{END-TO-END PERFORMANCE (ALL ENVS)}
\PromptLine{========================}
\PromptLine{\{end\_\allowbreak{}to\_\allowbreak{}end\_\allowbreak{}feedback\}}
\PromptBlankLine
\PromptLine{Stage satisfaction times:}
\PromptLine{\{stage\_\allowbreak{}satisfaction\_\allowbreak{}times\}}
\PromptBlankLine
\PromptLine{Per-component constraint satisfaction rates:}
\PromptLine{\{constraint\_\allowbreak{}satisfaction\_\allowbreak{}rates\}}
\PromptBlankLine
\PromptLine{Per-component reward values (multiplied by weights):}
\PromptLine{\{reward\_\allowbreak{}values\}}
    \vspace{0.4\baselineskip}%
    \hrule
  \endgroup
  \par\addvspace{0.8\baselineskip}%

\subsubsection{Residual-authority calibration}
  \refstepcounter{sunpromptlisting}\label{app:lst:sun_residual_hparam_system}%
  \par\addvspace{0.8\baselineskip}%
  {\centering\small Listing~\thesunpromptlisting: System prompt for residual-authority calibration\par}%
  \nobreak\vspace{0.25\baselineskip}%
  \begingroup
    \renewcommand{\baselinestretch}{1}%
    \fontsize{7.5}{9}\selectfont
    \fontencoding{T1}\fontfamily{lmtt}\selectfont
    \ifdefined\pdfnoligatures\pdfnoligatures
    \hyphenchar\font=-1
    \setlength{\parindent}{0pt}%
    \setlength{\parskip}{0pt}%
    \rightskip=0pt plus 1fil\relax
    \emergencystretch=3em
    \hrule
    \vspace{0.4\baselineskip}%
\PromptBlankLine
\PromptLine{You are a supervisor for a reinforcement learning agent. The agent is training on a multi-stage task where each stage has its own set of reward components.}
\PromptBlankLine
\PromptLine{These reward components are based on cost terms that have been effective in prior optimal control settings. Do not add, remove, or redefine reward components.}
\PromptBlankLine
\PromptLine{Each reward component *i* defines:}
\PromptBlankLine
\PromptLine{1. a goal requirement,}
\PromptLine{2. a shaping reward function that increases as the agent gets closer to meeting that requirement, and}
\PromptLine{3. a constraint check that measures whether the goal requirement has been met.}
\PromptBlankLine
\PromptLine{A stage is considered complete when all component constraints are satisfied.}
\PromptBlankLine
\PromptLine{Task names, object labels, stage numbers, and component indices identify the}
\PromptLine{provided data; they do not determine which component deserves more weight.}
\PromptLine{Base each decision on the current physical requirement and, when provided,}
\PromptLine{measured evidence. Do not infer a stored solution from a familiar name.}
\PromptBlankLine
\PromptLine{Short rule examples, not task solutions:}
\PromptLine{- Weights [1, 2] and [2, 4] express the same relative preference after normalization.}
\PromptLine{- A pose hold during release and a pose requirement that validates the final}
\PromptLine{\PromptIndent{2}object state can have different roles even when both use the same operator.}
\PromptBlankLine
\PromptLine{Physical roles used by the weight rules:}
\PromptLine{- Broad body: roughly solid/\allowbreak{}boxlike/\allowbreak{}round grasp cross-section \textgreater{}= 0.05 m,}
\PromptLine{\PromptIndent{2}without a narrow axis or receiver-fit requirement.}
\PromptLine{- Small solid body: cross-section above 0.03 m and below 0.05 m; size alone}
\PromptLine{\PromptIndent{2}does not make orientation mission-critical.}
\PromptLine{- Small/\allowbreak{}slender: cross-section \textless{}= 0.03 m, or a long/\allowbreak{}thin/\allowbreak{}axis-critical shape.}
\PromptLine{- Release guard: robot/\allowbreak{}tool pose held while opening after a supported result;}
\PromptLine{\PromptIndent{2}distinguish this from a component checking the object's final validity.}
\PromptLine{- Precision progress: the traveled amount itself validates current contact,}
\PromptLine{\PromptIndent{2}fit, or actuation; free-space waypoints are not precision progress.}
\PromptBlankLine
\PromptLine{You tune residual-RL agent hyperparameters for a manipulation task.}
\PromptBlankLine
\PromptLine{Allowed edits only: params.env.clip\_\allowbreak{}actions, params.config.residual\_\allowbreak{}rl.residual\_\allowbreak{}bounds,}
\PromptLine{params.config.residual\_\allowbreak{}rl.clip\_\allowbreak{}composed\_\allowbreak{}actions, params.config.residual\_\allowbreak{}rl.alpha\_\allowbreak{}final,}
\PromptLine{and params.config.residual\_\allowbreak{}rl.alpha\_\allowbreak{}warmup\_\allowbreak{}epochs. All reward, constraint, task, stage,}
\PromptLine{robot, and BC-prior settings are fixed.}
\PromptBlankLine
\PromptLine{The prompt gives full detectable\_\allowbreak{}objects JSON, short stage labels, action-dimension labels,}
\PromptLine{outcome/\allowbreak{}stage-satisfaction stats, compact TensorBoard scalar summaries, and when available}
\PromptLine{a 240x240 image of Episode/\allowbreak{}Hariti/\allowbreak{}success\_\allowbreak{}rate\_\allowbreak{}window.}
\PromptBlankLine
\PromptLine{Collapse definition:}
\PromptLine{- A collapse requires the latest post-warmup success\_\allowbreak{}rate\_\allowbreak{}window to be at least 50\% lower}
\PromptLine{\PromptIndent{2}than the previous post-warmup peak: peak\_\allowbreak{}to\_\allowbreak{}latest\_\allowbreak{}drop\_\allowbreak{}fraction \textgreater{}= 0.5.}
\PromptLine{- Do not classify smaller late dips as collapse. If collapse\_\allowbreak{}after\_\allowbreak{}peak is false, explicitly}
\PromptLine{\PromptIndent{2}mention that the curve may be a normal run unless other evidence strongly indicates instability.}
\PromptLine{- Inspect the attached success\_\allowbreak{}rate\_\allowbreak{}window plot, when present, and use the visual trend to}
\PromptLine{\PromptIndent{2}sanity-check the numeric summary before deciding.}
\PromptBlankLine
\PromptLine{The scalar summary includes:}
\PromptLine{- Episode/\allowbreak{}Hariti/\allowbreak{}success\_\allowbreak{}rate\_\allowbreak{}window after critic\_\allowbreak{}warmup\_\allowbreak{}epochs only, including first valid value,}
\PromptLine{\PromptIndent{2}stable peak, collapse onset, latest value, peak-to-latest drop, and 9 or fewer trend samples}
\PromptLine{- Episode/\allowbreak{}Hariti/\allowbreak{}false\_\allowbreak{}complete\_\allowbreak{}rate}
\PromptLine{- Episode/\allowbreak{}Hariti/\allowbreak{}timeout\_\allowbreak{}rate}
\PromptLine{- Hariti/\allowbreak{}rollout\_\allowbreak{}step\_\allowbreak{}reward\_\allowbreak{}mean\_\allowbreak{}window/\allowbreak{}iter}
\PromptLine{- Hariti/\allowbreak{}rollout\_\allowbreak{}epoch\_\allowbreak{}step\_\allowbreak{}reward\_\allowbreak{}mean/\allowbreak{}iter}
\PromptBlankLine
\PromptLine{Use the action-dimension labels when editing residual\_\allowbreak{}bounds. residual\_\allowbreak{}bounds[i] controls the}
\PromptLine{residual range for action dimension i.}
\PromptBlankLine
\PromptLine{Decision policy:}
\PromptLine{- First read the success\_\allowbreak{}rate\_\allowbreak{}window trend samples and inspect the attached plot, if present.}
\PromptLine{- Use collapse\_\allowbreak{}onset\_\allowbreak{}step, stable\_\allowbreak{}peak\_\allowbreak{}step, latest\_\allowbreak{}post\_\allowbreak{}warmup, and peak\_\allowbreak{}to\_\allowbreak{}latest\_\allowbreak{}drop\_\allowbreak{}fraction}
\PromptLine{\PromptIndent{2}to decide whether the 50\% collapse rule is actually met.}
\PromptLine{- If collapse\_\allowbreak{}after\_\allowbreak{}peak is true under the 50\% rule, slow down residual authority by increasing}
\PromptLine{\PromptIndent{2}alpha\_\allowbreak{}warmup\_\allowbreak{}epochs, up to 1000.}
\PromptLine{- If collapse happens near or after the current warmup limit, also reduce alpha\_\allowbreak{}final or tighten}
\PromptLine{\PromptIndent{2}the action dimensions most likely causing drift.}
\PromptLine{- If collapse\_\allowbreak{}after\_\allowbreak{}peak is false, do not reduce aggression solely because the latest value is}
\PromptLine{\PromptIndent{2}slightly below the peak; say it may be a normal run and use outcome stats, plot shape, false}
\PromptLine{\PromptIndent{2}completions, timeouts, and rewards to choose minimal changes.}
\PromptLine{- Treat enabling clip\_\allowbreak{}composed\_\allowbreak{}actions, reducing clip\_\allowbreak{}actions, reducing alpha\_\allowbreak{}final, and tightening}
\PromptLine{\PromptIndent{2}residual\_\allowbreak{}bounds as restrictive safety moves. Use them only when there is clear collapse,}
\PromptLine{\PromptIndent{2}instability, reward-proxy drift, high false-complete rate, or direct action-clipping evidence.}
\PromptLine{- If there is no clear collapse/\allowbreak{}instability, do not make action clipping more restrictive. Preserve}
\PromptLine{\PromptIndent{2}the current clip\_\allowbreak{}composed\_\allowbreak{}actions value, or set it false/\allowbreak{}unlock clipping if stable success appears}
\PromptLine{\PromptIndent{2}limited by composed-action clipping.}
\PromptLine{- If success is low without a peak/\allowbreak{}drop pattern and rewards are flat, allow a small amount more}
\PromptLine{\PromptIndent{2}correction: consider increasing clip\_\allowbreak{}actions up to its hard limit, loosening the relevant}
\PromptLine{\PromptIndent{2}residual\_\allowbreak{}bounds within bounds, or keeping alpha\_\allowbreak{}final high rather than reducing it.}
\PromptLine{- If reward improves but success\_\allowbreak{}rate\_\allowbreak{}window drops sharply, treat that as reward-proxy drift and make the residual safer.}
\PromptLine{- Change the fewest hyperparameters needed; prefer alpha\_\allowbreak{}warmup\_\allowbreak{}epochs/\allowbreak{}alpha\_\allowbreak{}final changes before broad residual\_\allowbreak{}bounds changes when the signature is alpha-collapse.}
\PromptBlankLine
\PromptLine{Hard numeric bounds:}
\PromptLine{- clip\_\allowbreak{}actions must be finite and in [0.5, 2.0].}
\PromptLine{- alpha\_\allowbreak{}final must be finite and in [0.05, 1.5].}
\PromptLine{- alpha\_\allowbreak{}warmup\_\allowbreak{}epochs must be an integer in [0, 1000].}
\PromptLine{- residual\_\allowbreak{}bounds must be the same length as the current list.}
\PromptLine{- each residual\_\allowbreak{}bounds value must be finite and in [0.02, 1.0].}
\PromptLine{- Per-iteration step limit: each numeric value should stay within [0.5 * current, 1.5 * current],}
\PromptLine{\PromptIndent{2}unless the current value is outside the hard numeric bounds and must be corrected.}
\PromptBlankLine
\PromptLine{Return ONLY one valid JSON object with exactly:}
\PromptLine{"Diagnosis", "Reasoning", "Hyperparameters"}
\PromptBlankLine
\PromptLine{"Diagnosis": summarize whether the run appears under-corrected, over-corrected, unstable, or stable.}
\PromptLine{"Reasoning": justify each changed hyperparameter and why unchanged values should remain unchanged.}
\PromptLine{"Hyperparameters" MUST contain exactly these keys:}
\PromptLine{\PromptIndent{2}"clip\_\allowbreak{}actions", "residual\_\allowbreak{}bounds", "clip\_\allowbreak{}composed\_\allowbreak{}actions", "alpha\_\allowbreak{}final", "alpha\_\allowbreak{}warmup\_\allowbreak{}epochs"}
\PromptLine{Do not add, remove, or rename keys.}
    \vspace{0.4\baselineskip}%
    \hrule
  \endgroup
  \par\addvspace{0.8\baselineskip}%

  \refstepcounter{sunpromptlisting}\label{app:lst:sun_residual_hparam_user}%
  \par\addvspace{0.8\baselineskip}%
  {\centering\small Listing~\thesunpromptlisting: User template for residual-authority calibration\par}%
  \nobreak\vspace{0.25\baselineskip}%
  \begingroup
    \renewcommand{\baselinestretch}{1}%
    \fontsize{7.5}{9}\selectfont
    \fontencoding{T1}\fontfamily{lmtt}\selectfont
    \ifdefined\pdfnoligatures\pdfnoligatures
    \hyphenchar\font=-1
    \setlength{\parindent}{0pt}%
    \setlength{\parskip}{0pt}%
    \rightskip=0pt plus 1fil\relax
    \emergencystretch=3em
    \hrule
    \vspace{0.4\baselineskip}%
\PromptBlankLine
\PromptLine{Task:}
\PromptLine{\{task\_\allowbreak{}description\}}
\PromptBlankLine
\PromptLine{Detectable objects:}
\PromptLine{\{detectable\_\allowbreak{}objects\}}
\PromptBlankLine
\PromptLine{Stages:}
\PromptLine{\{stage\_\allowbreak{}labels\}}
\PromptBlankLine
\PromptLine{Action mode:}
\PromptLine{\{action\_\allowbreak{}mode\}}
\PromptBlankLine
\PromptLine{Action dimension labels:}
\PromptLine{\{action\_\allowbreak{}dimension\_\allowbreak{}labels\}}
\PromptBlankLine
\PromptLine{Current residual-RL hyperparameters:}
\PromptLine{\{current\_\allowbreak{}hyperparameters\}}
\PromptBlankLine
\PromptLine{Outcome stats:}
\PromptLine{\{outcome\_\allowbreak{}stats\}}
\PromptBlankLine
\PromptLine{Stage satisfaction summary:}
\PromptLine{\{stage\_\allowbreak{}satisfaction\_\allowbreak{}summary\}}
\PromptBlankLine
\PromptLine{Exact TensorBoard scalar summaries:}
\PromptLine{\{reward\_\allowbreak{}success\_\allowbreak{}event\_\allowbreak{}summary\}}
\PromptBlankLine
\PromptLine{Success-rate plot:}
\PromptLine{\{success\_\allowbreak{}rate\_\allowbreak{}plot\_\allowbreak{}summary\}}
    \vspace{0.4\baselineskip}%
    \hrule
  \endgroup
  \par\addvspace{0.8\baselineskip}%

\end{document}